%% file: main.tex
\documentclass[10pt,journal,compsoc]{IEEEtran}

\input{packages.tex}

\SetKwInput{KwData}{Inputs}
\usepackage{adjustbox}
\def\argmin{\mathop{\arg\min}\limits}

\newcommand{\prl}[1]{\left(#1\right)}
\newcommand{\brl}[1]{\left[#1\right]}
\newcommand{\crl}[1]{\left\{#1\right\}}

\begin{document}
\title{Distributed Bayesian Estimation in Sensor Networks: Consensus on Marginal Densities}



\author{Parth~Paritosh,~\IEEEmembership{Member,~IEEE,}
        Nikolay~Atanasov,~\IEEEmembership{Senior Member,~IEEE,}
        and~Sonia~Mart{\'\i}nez,~\IEEEmembership{Fellow,~IEEE}
\IEEEcompsocitemizethanks{\IEEEcompsocthanksitem The authors are with the Contextual Robotics Institute, University of California San Diego, 9500 Gilman Dr, La Jolla, CA 92093 (e-mails: \{pparitos,natanasov,soniamd\}@ucsd.edu). We gratefully acknowledge support from ONR N00014-19-1-2471, ARL DCIST CRA W911NF17-2-0181, and NSF FRR CAREER 2045945.
}
\thanks{Manuscript received ...}}

\markboth{Transactions on Network Science and Engineering}
{Authors ...}
%

\IEEEtitleabstractindextext{%
\begin{abstract}
  In this paper, we design and analyze distributed Bayesian
  estimation algorithms for sensor networks. \revisionAdd{We consider estimation problems, such as cooperative localization and federated learning, where the data collected at any agent depends on a subset of all variables of interest. We provide a unified formulation of centralized, distributed and marginal probabilistic estimation as a Bayesian density estimation problem using data from non-linear likelihoods at agent. We develop distributed estimation algorithms based on stochastic mirror descent with appropriate regularization to enforce distributed or marginal density constraints. We prove almost-sure convergence to the optimal set of probabilities at each agent in both the distributed and marginal settings. Finally, we present Gaussian density versions of these algorithms and compare them to belief propagation variants in a node localization problem with relative position measurements. We also demonstrate our algorithms in a multi-agent mapping problem using LiDAR data.}
\end{abstract}

}

\maketitle

\IEEEdisplaynontitleabstractindextext

%
\IEEEpeerreviewmaketitle


\section{Introduction}
\IEEEPARstart{T}{he} advent of low-cost computing, storage and
communication devices has made large sensor networks integral to
urban, transportion and power-grid infrastructure. 
\revisionAdd{Efficient inference algorithms are needed for automated monitoring of the underlying processes.}
Any centralized solution to this inference problem necessitates data aggregation
\revisionAdd{which, while potentially more accurate, incurs prohibitive processing and communication costs, } 
especially in real-time settings.
Real-time inference is crucial for tasks such as indoor positioning~\cite{FZ-AG-KKL:19}, urban monitoring~\cite{SK-AD-SSH-JSK-SES:16}, and path planning for robotic networks~\cite{FB-JC-SM:09}. Thus, modern sensor networks parallelize inference across nodes improving communication efficiency and robustness to node failures.


However, most distributed algorithms do not account for the relevance of
the information shared among the nodes. Motivated by this, we
design algorithms to simultaneously address the inherent
commmunication network constraints while accounting for variable
relevance at each node. 


\textit{Literature review}: To achieve online estimation in connected sensor 
networks, researchers have studied schemes \revisionAdd{to combine distributed 
estimates \cite{AJ-PM-AS-ATS:12}, notably classified as} opinion 
pooling \cite{RTC-RLW:99} and 
graph-based message-passing algorithms \cite{TM:05}. 
Message-passing algorithms, \revisionAdd{such as Gaussian, sigma-point and 
non-linear belief propagation (BP),} are appropriate when the causal 
relationships between variables are known. 
For further insights, see \cite{FM-OH-FH:13} and references therein. 
In contrast, linear and geometric averages of \revisionAdd{probabilistic estimates} 
are commonly used to \revisionAdd{pool opinions \cite{MK-YI-ET-AS:23}} in a network 
with communication across one-hop neighbors. 
The seminal work in \cite{AJ-PM-AS-ATS:12} presents a local and computationally tractable consensus estimation algorithm as a two step process, consisting of a
non-Bayesian pooling step followed by a Bayesian update with locally
available data.

Distributed estimation algorithms can be analyzed as steps of
\revisionAdd{gradient-based optimization methods \cite{MEK-HR:23}} that minimize 
the divergence between the data generating process and the estimated model. 
This approach establishes consistency of the estimation task, 
with estimation quality as the objective.  
For the consensus step, this approach generates algorithms
beyond linear and logarithmic \revisionAdd{pooling choices, see~\cite{AG-TSJ-SV-HZ:04, GK-YE-PD-FH:22}}.  
Mirror descent methods~\cite{AN:12, ZZ-PM-NB-SPB-PWG:20} 
generalize the first-order gradient methods via metric-space projections 
to exploit the inherent problem geometry. 
Past research on distributed estimation using partially informative
observation models has relied on fusing observation likelihoods with
individual agent's network sized estimates~\cite{KRR-ATS:10,
  AN-AO-CAA:17, AL-AS-TJ:14}. Doan et al.~\cite{TTD-SB-HDN-CLB:18}
apply mirror descent to the linear average of neighbor estimates for
consistent estimation in discrete space. 
Another algorithm in \cite{AN-AO-CAA:17} incorporates geometric averaging 
with stochastic mirror descent (SMD) to achieve consensus over the network. 
\revisionAdd{As centralized objective, one can select the divergence between true and estimated densities
to derive linear regression updates, Kalman filter and particle filters as special cases. } 
The work in \cite{ZZ-PM-NB-SPB-PWG:20} further extends the SMD
algorithm for finding optimal continuous-space probability density
functions (pdfs), although in a centralized setting with a variationally
coherent objective. More recently, \cite{CAA-AO-AN:22} studied 
convergence of variational estimates on compact subsets of hypotheses.
However, all of these papers assume that agents estimate a common set of 
variables and neither one includes distributional convergence guarantees.

In this work, in addition to distributing the estimation process, we focus on distributing the storage by estimating only a subset of variables relevant to the local data generating process at each node. This significantly reduces the storage and communication requirements for distributed inference.
One example of estimating relevant variable subsets at different nodes 
is a sensor network using relative measurements for node localization 
\cite{NA-RT-VP:14}. 
In this problem, the measurement likelihoods are determined by the position of node $i$ 
making the measurement and the positions of the measured neighbors $\mathcal{V}_i$. 
A practical example of relative-measurement localization is a beacon network deployed 
in underwater or indoor settings using range or acoustic measurements to estimate 
the node positions \cite{NA-RT-VP:14, GP-IS-BF-FB-BDA:13}.
Since we estimate marginal densities over the relevant variables at different nodes, 
we design and analyze algorithms to enforce consistent marginals of the network-sized joint pdf.

\revisionAdd{
BP \cite{JP:82,JSY-WTF-YW:02} is a widely used algorithm for probabilistic estimation of marginal densities in a network with applications in error-correcting codes \cite{YL-PD:19}, computer vision, and robotics \cite{DK-SL-AO-OCJ:19}.
This method employs node-specific observation models and pairwise interaction models between agents, 
utilizing messages exchanged between neighboring nodes to compute the marginal probabilities 
of individual variables at each node.
The convergence of BP in generally not guaranteed in graphs with loops \cite{TH:04}. 
Recently proposed variants, such as $\alpha$-BP \cite{LD:20} 
and circular BP \cite{VB:21}, obtain consistent estimates
in arbitrary graphs but the convergence guarantees are limited to binary probabilities. Instead of learning marginals over local node variables only, 
we design an algorithm estimating the marginal probability density over a set of relevant variables at each node.}

The key challenge \revisionAdd{to guarantee consistency in estimating different variables at different agents is the incompatibility of the variable domains due to the different number of neighbors at each agent.} 
This sub-problem of combining partial estimates has been framed 
in terms of statistical matching \cite{BV:08} \revisionAdd{and minimum entropy coupling \cite{YYS-AKY:23}}, 
aiming to find a joint pdf minimizing divergence to the relevant marginal densities.
\revisionAdd{A recursive optimization approach is proposed in \cite{JK:11, FC-LG-UV:19} 
but it is computationally expensive for real-time inference.
In the presence of streaming measurements, our prior work \cite{PP-NA-SM:19} addressed a discrete version of this problem. However, in various applications it is necessary to consider probability densities in continuous space.}



\textit{Statement of Contributions:} 

\revisionAdd{This work proposes a distributed Bayesian estimation algorithm to obtain marginal densities  over relevant variable subsets at each node.}  
\revisionAdd{The contributions of this paper are summarized as follows.}
(i) We formulate the estimation problem as a stochastic optimization 
over the functional space of probability density functions, 
\revisionAdd{presenting a unified framework to express centralized, distributed 
and marginal estimation in a network.
This formulation relates their solutions using gradient descent variants to 
Bayesian estimation algorithms.}
\revisionAdd{
(ii) We develop two distributed estimation algorithms relying on one-hop neighbor communication, one estimating densities over all unknown variables and the other 
estimating marginal densities only over a relevant set of variables at each agent. Our distributed marginal density estimation algorithm reduces the storage, communication, and computation requirements compared to consensus-based distributed estimation algorithms \cite{AN-AO-CAA:17, RP-MF-BT:22,CAA-AO-AN:22,MK-YI-ET-AS:23}
(iii) We prove novel almost-sure convergence result for our distributed and marginal algorithms. Our results apply to continuous probability densities and hold in any connected network, in contrasts with message-passing and belief propagation methods  \cite{DB:08,JP:22} that generally cannot provide convergence guarantees in graphs with cycles.
(iv) We demonstrate that our algorithms achieve higher estimation accuracy than belief propagation in a distributed node localization problem using relative position measurements and significantly reduce the storage and communication load compared to full state estimation algorithms in a distributed mapping problem using LiDAR data.}


This paper extends our prior work \cite{PP-NA-SM:20} on estimating marginal densities 
over the states of an agent and its neighbors to an arbitrary set of variables by introducing a marginal consensus constraint. \revisionAdd{Additionally, we analyze the convergence of the distributed and marginal algorithms and provide a new application to distributed mapping}.
\revisionAdd{We also extend a relative localization example from \cite{PP-NA-SM:20} by comparing the performance of our algorithms to new variants of the BP algorithm \cite{ATI-ASW:05, DB:08, VB-RJ-SD:24} in networks with different connectivity and observation noise.}

In Section~\ref{sec:probform}, we pose the distributed estimation problem as minimizing 
divergence between the data-generating density and an estimated likelihood, 
and recall relevant mathematical preliminaries in Section~\ref{sec:functional}.
An SMD-based solution to this problem is presented in Section~\ref{sec:centralized}.
Next, we solve the distributed estimation problem in Section~\ref{sec:distSMD} 
where agents maintain equal network-scale estimates.
Section~\ref{sec:partcon} extends the estimation problem to a marginal density setting where agents maintain estimates on variables co-estimated with one-hop neighbors.
Finally, Section~\ref{sec:dmgvi} presents \revisionAdd{a distributed relative localization 
example comparing the proposed algorithms with BP variants} and a distributed mapping application 
using the marginal estimation in conjunction with variational inference.

\section{Problem Formulation: Distributed partial parameter estimation}
\label{sec:probform}
We consider an estimation problem with cooperative agents 
in the set $\nodes = \until{n}$ communicating over a static connected network. 
The agents aim to infer $m$~vector values collectively given as the
$d$-dimensional vector~$\hypspace^\star = \brl{\vect{x}_{1}^\star, \dots, \vect{x}_{m}^\star }^{\top}$
with $\vect{x}_{v}^\star \in \real^{d_v}$ and $d = \sum_v d_v$.
With a abuse of notation, we overload~$\hypspace^{\star}$ to also
denote the set of $m$-vectors~$\crl{\vect{x}_{v}^\star}_{v=1}^m$.
The terms~$\vect{x}_v^\star$ may represent the value of
model parameters in a mapping problem, 
or the agents' pose in a relative localization problem.
Each agent receives measurements from a local probability density 
function dependent on a subset~$\hypspace_i^\star 
\subseteq \hypspace^\star$ and shares its estimates with one-hop neighbors.
The variables in the local subset~$\hypspace_i^\star$ could represent 
model parameters relevant to the agent's trajectory in a mapping problem,
or the agent neighbors' poses in a localization problem.
Relying on the subsets $\hypspace_i^\star$ instead of $\hypspace^\star$ reduces the storage and communication costs of distributed estimation at individual agents.


To set up the estimation problem formally, 
we define a vector~$\hypspace = \brl{\vect{x}_{1}, \dots, \vect{x}_{m}}$
with $\vect{x}_v \in \real^{d_v}$ corresponding to the variables of interest $\vect{x}_v^\star$. 
At time step~$t$, the known likelihood of receiving measurement $z_{i,t} \in \real^{\ell_i}$ by agent~$i$ is given as $\qz{i}(z_{i,t}|\hypspace_i)$, where $\hypspace_i \subseteq \hypspace$. Thus, the measurement generation at each agent $i$ is determined by the unknown variables $\hypspace_i^\star$ via the density model $\qz{i}^\star (z_{i,t}) = \qz{i}(z_{i,t}|\hypspace_i = \hypspace_i^\star) \in \calF_{\ell_i}$,
where the space $\calF_{\ell}$ of pdfs is defined as:
\begin{equation}
\label{eqn:functionalspace}
\resizebox{0.9\linewidth}{!}{$\displaystyle{\calF_{\ell} = \left\{g \in \Lp(\real^{\ell})\, | \int
  g(\vect{x})d\vect{x} = 1, g(\vect{x}) \geq 0, \forall
  \vect{x} \in \real^{\ell} \right\}.}$}
\end{equation}
We assume that $\cup_i \hypspace_i^\star = \hypspace^\star$ to ensure that the combined agent network can jointly observe all variables of interest.
Let~$z_t$ represent all observations~$z_{i,t}$ collected by 
the multi-agent system at time~$t$ with combined likelihood model $\qz{}(z_{t}|\hypspace) \in \calF_{\ell}$, where~$\ell = \sum_{i=1}^n \ell_i$.

\begin{assumption}[Independence]
  \label{assume:independence}
  Agent~$i$ samples observation~$z_{i,t}$ at time $t$ independently across time and agents as, 
  \begin{align}
    \qz{}(z_1, \dots, z_T|\hypspace^\star) = \prod_{t=1}^{T}\qz{}(z_t|\hypspace^\star) 
    = \prod_{t=1}^{T} \prod_{i \in \nodes }\qz{i}(z_{i,t}|\hypspace_i^\star)
  \end{align}
\end{assumption}

Since the agents need to reach consistent estimates, 
any two agents observing the same variable communicate their estimates 
over a connected digraph~$\graph$ \cite{FB-JC-SM:09}, 
with node set~$\nodes$ and edge set~$\edges$. 
The neighbors of agent $i$, including itself, are denoted as~$\neighbor{i}$. 
The communication graph has an associated
non-negative adjacency matrix~$A \in \real^{n \times n}$ with entries~$A_{ij}>0$ 
iff $(i, j) \in \edges$, including self-loops. 
Any such matrix~$A$ representing a connected network can be made symmetric 
and doubly stochastic, e.g., via the Sinkhorn's algorithm~\cite{RS-PK:67}. 

\begin{assumption}[Graph adjacency]
\label{assume:dsmat}
The connected digraph $\graph$ is represented by a symmetric, doubly stochastic
adjacency matrix~$A$ with $A \ones[n] = \ones[n], A =
A^{\top}, $ and diagonal entries $A_{ii} > 0 \, , \forall i\in \{1,
\dots, n\}$, where~$\ones[n]\in \real^n$ is a vector of ones. 
\end{assumption} 

Next, we express the estimation problem using a pdf $p(\hypspace)\in \calF_d$ 
instead of a point estimate in $\real^d$ to capture the associated epistemic uncertainty. We aim to find the pdf $p \in \calF_{d}$ minimizing the objective:
  \begin{align}
  \label{eqn:c_objective}
  \min_{p \in \calF_{d}} & \left\lbrace \underset{\hypspace \sim
      p}{\expect} [\KL [\qz{}(\cdot |
      \hypspace^\star), \qz{}(\cdot |\hypspace)] \right\rbrace,
  \end{align}
  where the expectation is defined over the KL-divergence term 
  $\KL[\qz{}^\star, \qz{}(\cdot |\hypspace)] = \int_{\real^{\ell}}
  \qz{}(z | \hypspace^\star) \log(\frac{\qz{}(z | \hypspace^\star)}{\qz{}(z |\hypspace)} ) dz $ quantifying the discrepancy between 
  the true likelihood pdf~$\qz{}^\star \triangleq \qz{}(\cdot|\hypspace^\star)$ and the agent likelihood models.  
Since the divergence is zero iff~$\qz{}^\star = \qz{}(\cdot| \hypspace)$ almost everywhere (a.e.) w.r.t. the Lebesgue measure, 
the Dirac-delta function at~$\hypspace = \hypspace^\star$ 
lies in this objective's minimizer set.
Please note that the equality of measures is understood 
in this sense throughout the manuscript. 
Additional minimizers would satisfy the property of
\textbf{observational equivalence}; i.e.,~any two values~$\hypspace_a,
\hypspace_b \in \real^d$ are observationally equivalent, if the
corresponding likelihoods satisfy~$\qz{}(\cdot| \hypspace_a) =
\qz{}(\cdot|\hypspace_b)$. Observational equivalence relates the
solutions in pdf space to the vector space of $\hypspace$. Every point
$\hypspace_a$ observationally equivalent to $\hypspace^\star$ is
included in the set of minimizers. 


As we sequentially sample the true likelihood pdf~$\qz{}^\star$, we aim 
to find the minimizing argument~$p$ of the sample average approximation
w.r.t.~$z_t$ as shown next. The optimization presented here
follows stochastic programming~\cite{AS-DD-AR:14}, and we make use of
the inner product notation~$\langle p_1, p_2 \rangle = \int p_1 p_2 dz, 
\mathrm{for }\, p_1, p_2 \in \calF_{\ell}$. From~\eqref{eqn:c_objective},
\begin{align}
\label{eqn:c_objective0}
p^\star \in & \argmin_{p \in \calF_d} \left\lbrace \underset{\hypspace
    \sim p}{\expect} [\KL [\qz{}^\star ,
  \qz{}(\cdot|\hypspace) ]] \right\rbrace \nonumber \\
= & \argmin_{p \in \calF_{d}} \left\lbrace \underset{\hypspace \sim
    p}{\expect} [- \langle \qz{}^\star,
  \log(\qz{}(\cdot|\hypspace)) \rangle] \right\rbrace  \\
= & \argmin_{p \in \calF_{d}} \left\lbrace \underset{z_t \sim
    \qz{}^\star}{\expect} \, F_t[p] \right\rbrace \equiv
\calF^\star, \nonumber
\end{align}
\begin{align}
\label{eqn:c_objective1}
f[p]  = \underset{z_t \sim \qz{}^\star}{\expect} F_t[p], \;\;\; 
  & F_t[p] = \underset{\hypspace \sim p}{\expect} [-
\log(\qz{}(z_t|\hypspace) )],
\end{align}
where the first equality in~\eqref{eqn:c_objective0}, follows from the independence 
of the entropy term~$\int \qz{}^\star \log(\qz{}^\star)$ w.r.t.~$\hypspace$. 
The set~$\calF^\star$ contains pdfs minimizing the objective function in~\eqref{eqn:c_objective0}.
Using Fubini-Tonelli's theorem, we switch the data and state variable integrals
to obtain the last equality of~\eqref{eqn:c_objective0}, defined using~\eqref{eqn:c_objective1}.
Since $\qz{}^\star$ is unknown, we approximate the expectation operator 
in the final equality of~\eqref{eqn:c_objective0} in terms of sampled data in~$\{z_t\}$,
and state the estimation problem as follows.

\begin{problem}[Centralized estimation]
  Given observations $\crl{z_{i,t}}_{i=1}^n$ and known agent likelihoods 
  $\prod_{i=1}^n \qz{i}(z_{i,t}| \hypspace_i)$
  defined over the subsets of $\hypspace$,
  find the pdf~$p \in \calF_{d}$ minimizing the approximation to the objective in~\eqref{eqn:c_objective}:
  \begin{align}
    \label{eqn:c_objective_sum}
    \min_{p \in \calF_d} \left\lbrace \frac{1}{T} \sum_{t=1}^T F_t[p] \right\rbrace ,
  \end{align}
  where the functional $F_t$ is defined in \eqref{eqn:c_objective1}.
\end{problem}


Assuming that the estimate pdf~$p \in \calF_d$ lies in $\Lp$, 
the inner product objective defined in~\eqref{eqn:c_objective1} exists
if the gradient of the objective is defined in the 
dual space~$\mathrm{L}^{\infty}$. 
Given the gradient definition~$\frac{\delta}{\delta p} F[p] =
-\log(\qz{}(z|\hypspace))$, the dual-space norm
is~$\Vert\frac{\delta}{\delta p} F[p]\Vert_{\infty} =
\mathrm{sup}_{z} [- \log(\qz{}(z|\hypspace))]$. Therefore, the
gradient exists if the~$[- \log(\qz{}(z|\hypspace))] < \infty$ for
all choices of~$z$.  We highlight this requirement in the
next assumption\footnote{The assumption makes use of a functional 
derivative defined in the following section.}.

\begin{assumption}[Bounded gradient]
\label{assume:gradbound}
The gradient of the objective functional~$\left\vert\frac{\delta F}{\delta
    \pi}(\pi, z)\right\vert= |-\log(\qz{i}(z|\hypspace) )|
\leq L$ is uniformly bounded for all~$\pi \in \calF_{m},$
$z \in \real^{d_z}$. This implies
that~$|\log(\qz{i}(\cdot|\hypspace) )|$ (resp.~$
\qz{i}(\cdot|\hypspace)$) are uniformly upper (resp.~lower)
bounded. 
\end{assumption}
The uniform lower bound on the likelihood~$0< \alpha<
\qz{i}(\cdot|\hypspace)$ has an `expected data'
interpretation, i.e., a strictly positive likelihood of receiving data~$z_{i,t}$ at agent~$i$.

The linearity of the objective function with respect to $p$
and the independence assumptions on the data model are necessary to
derive the algorithms in this work. The independence across time
enables writing the sampling average, whereas the independence across
agents allows us to obtain a distributed formulation in the following
sections so that each agent~$i$ can estimate a copy or a marginal of a true
pdf~$p^\star$.


\section{Convex functionals and sequences}
\label{sec:functional}
This section reviews the stochastic mirror descent (SMD) algorithm,
and relevant functional analysis and stochastic sequence results
needed to apply it to functional spaces. 

\subsection{The stochastic mirror descent algorithm}
The SMD algorithm \cite{SB:15, ASN-DBY:83} generalizes stochastic
gradient descent (SGD) to non-Euclidean spaces for convex optimization
problems via a divergence operator. 
Consider an arbitrary real-valued function $f(\vect{w},\vect{v})$ that is convex 
in its first argument $\vect{w} \in \real^n$ for $\vect{v} \in \real^m$ in its second argument.
We define an associated stochastic optimization problem as:
\begin{equation*}
  \min_{\vect{w}} 
  \mathbb{E}[f(\vect{w},\vect{v})] \approx \frac{1}{T} \sum_{t=1}^T f(\vect{w},\vect{v}_t),
\end{equation*}
where $\crl{\vect{v}_t}$ is a series of independent samples from a random variable whose distribution defines the expectation $\expect$. 
Precisely computing gradient with extensive sampling is computationally expensive. 
Instead, the \emph{SMD algorithm} optimizes iteratively 
using gradient samples $\nabla f(\vect{w}, \vect{v}_t)$ as,
\begin{equation}
\label{eqn:SMD}
\scaleMathLine[0.9]{\vect{w}_{t+1} \in \argmin_{\vect{w}} \left\{ \langle \nabla
    f(\vect{w}_t,\vect{v}_t), \vect{w} \rangle + \frac{1}{\alpha_t}
    D_\phi(\vect{w}, \vect{w}_t) \right\}.}
\end{equation}
Here, $\langle \cdot, \cdot \rangle$ is the inner product on $\real^n$
and $D_\phi(\vect{w}, \vect{w}_t)$ is the \emph{Bregman divergence} \cite{BAF-SS-MRG:08}
between $\vect{w}$ and $\vect{w}_t$.
\begin{definition}[Bregman divergence] 
  Consider a continuously differentiable and strictly convex
  function~$\phi:\calW \subseteq \real^n \rightarrow \real$.
  The \emph{Bregman divergence} associated with $\phi$ for
  points $\vect{w},\bar{\vect{w}} \in \calW$ is $D_\phi(\vect{w},
  \bar{\vect{w}}) := \phi(\vect{w}) - \phi(\bar{\vect{w}}) - \langle
  \nabla \phi(\bar{\vect{w}}), \vect{w}- \bar{\vect{w}} \rangle$.
\end{definition}


The choice $\phi(\vect{w}) = \|\vect{w}\|_2^2$ makes $D_\phi$ the
squared Euclidean distance and \eqref{eqn:SMD} the standard SGD algorithm. 
The convergence rate for the minimization of convex functions is $O(\frac{1}{\sqrt{T}})$, 
independently of the problem dimension~\cite{ASN-DBY:83}.

\subsection{Functional Bregman divergence and derivatives}
The stochastic optimization in~\eqref{eqn:c_objective_sum} is defined over the
functional space of pdfs $\calF_{d}$. Therefore, we generalize the terms in~\eqref{eqn:SMD}
to the pdf space $\calF_{d}$ to apply the SMD from~\eqref{eqn:c_objective_sum}.

Consider functions $p,g \in \Lp(\mathbb{R}^d)$. As before, the inner
product notation on $\Lp(\mathbb{R}^d)$ is defined as $\langle p,g \rangle := \int p g dx$, assuming the existence of this integral.
A subset $\cal{A}$ of $\Lp(\mathbb{R}^d)$ is convex if and only if 
$\alpha p + (1-\alpha)g \in \cal{A}$ for any $p,g \in \cal{A}$ 
and $\alpha \in [0,1]$. 
Therefore, the set of pdfs $\calF_{d}$ defined in~\eqref{eqn:functionalspace}
 is a closed convex subset of $\Lp(\mathbb{R}^d)$. To define a divergence operator
over $\calF_{d}$, we consider the entropy functional $\Psi[p] = \int p
\log(p) d\mu$ for $p \in \calF_{d}$. Entropy is continuously
differentiable and strictly convex as (i) $\calF_{d}$ is convex, (ii)
$x \log (x)$ is strictly convex over the positive real domain,
and (iii) the integration operator is linear, so it holds that
$\Psi[\alpha p + (1-\alpha)g] < \alpha \Psi [p] + (1-\alpha) \Psi[g]$
for all $p, g \in \calF_{d}$, $p \neq g$ a.e.. The Bregman divergence
associated with $\Psi$ is the \emph{Kullback-Leibler divergence}
 $\KL[p, g] := \int p \log(p /g) d\mu$. The KL-divergence inherits following properties
from the Bregman divergence \cite{BAF-SS-MRG:08}:
\begin{itemize}
\item (Convexity) The functional $\KL[p, g]$ is convex w.r.t. the
  first argument $p \in \calF_d$.
\item (Generalized Pythagorean inequality) For pdf's $p_0, p_1, p_2
  \in \calF_d$, the divergence terms are related to the directional
  gradients of~$\Psi$ as,
\begin{align}
\label{eqn:gpi}
& \left\langle \frac{\delta \Psi}{\delta p} [p_{2}], p_0-p_{2}
\right\rangle - \left\langle \frac{\delta \Psi}{\delta p}[p_{1}],
  p_0-p_{2} \right\rangle
\nonumber \\
& \quad = \KL [p_0, p_{1}] - \KL [p_0, p_{2}] - \KL [p_{2}, p_{1}].
\end{align}
\end{itemize}


The extension of SMD to pdfs in $\calF_{d}$ requires a definition of
the functional derivative. To evaluate how a functional $F$ changes in the
vicinity of $g \in \Lp(\mathbb{R}^d)$, we consider variations of $g$
defined as $g + \epsilon \eta$, where $\eta \in \Lp(\mathbb{R}^d)$ and
$\epsilon \ge 0$ is a small scalar. For fixed $g,\eta$, $F[g +
\epsilon \eta]$ is a function of $\epsilon$ and limits can be
evaluated in the usual sense.

\begin{definition} (\cite[p.~16]{DL:11})\label{def:var1} Consider a
  functional $F:\Lp(\real^d) \rightarrow \real$ and an arbitrary
  function $g \in \Lp(\real^d)$. A linear functional~$\frac{\delta
    F}{\delta g} [\eta]$ is called the \emph{first variation} of $F$
  at $g$ if for all $\eta \in \Lp(\real^d)$ and $\epsilon>0$ we have
\begin{align*}
F[g + \epsilon \eta] = F[g] + \epsilon \frac{\delta F}{\delta g}[\eta] + o(\epsilon),
\end{align*}
where $o(\epsilon)$ satisfies $\lim_{\epsilon \rightarrow 0}
o(\epsilon)/ \epsilon = 0$.
\end{definition}
The first variation of a functional is related to the Gateaux
derivative defined below.
\begin{definition} (\cite[p.~49]{WC:01})\label{def:gateaux}
  A functional $F:\Lp(\real^d) \rightarrow \real$ is
  \emph{Gateaux differentiable} at $g\in \Lp(\real^d)$, if the limit
  \begin{equation}\label{eq:directional-derivative}
    F'[g,\eta] := \lim_{\epsilon \rightarrow 0^+} 
    \frac{F[g + \epsilon    \eta] - F[g]}{\epsilon}
  \end{equation}
  exists for any $\eta \in \Lp(\real^d)$ and there is an element
  $\frac{\delta F}{\delta g} \in \Lp(\real^d)$ such that
  $\int\frac{\delta F}{\delta g} \eta d\mu = F'[g;\eta]$. 
  The element $\frac{\delta F}{\delta g}$ is the Gateaux derivative
  of functional $F$.
\end{definition}



\begin{proposition} \label{prop:variations}
  For $p, g \in \calF_{d}$, we have the following:
\begin{enumerate}
\item If $\Lambda[p] = \langle p, g \rangle$, then $\frac{\delta \Lambda}{\delta p}=
  g$,  
\item if $\Psi[p] = \langle p, \log(p) \rangle$, then $\frac{\delta
    \Psi}{\delta p}= 1 + \log p$, 
\item if  $\KL [p, g] = \langle p, \log(p/g) \rangle
    $, then $\frac{\delta \KL}{\delta p}= 1+ \log (p/g)$. 
\end{enumerate}
Each of the above first variations allow the computation of the
corresponding Gateaux derivatives following
Definition~\ref{def:gateaux}.
\end{proposition}
\begin{proof}
  See Appendix~\ref{app:Centralized}.
\end{proof}

\begin{definition} (\cite[Definition~$2.4$]{AT:08})\label{def:tvdist} 
  Let set $\mathcal{B}(\real^d)$ be the $\sigma$-algebra of the set $\real^d$.
  The total variation distance (TV) between two pdfs $p_0, p_1$ defined on $(\real^d, \mathcal{B}(\real^d))$ is,
  \begin{align*}
    \| p_0 - p_1 \|_{TV} = \underset{A \in \mathcal{B}(\real^d)}{\sup} |p_0(A) - p_1(A)|.
  \end{align*}
\end{definition}

\begin{lemma}[Pinsker's Lemma~\cite{MSP:60}] 
\label{lemma:pinsker}
The KL-divergence between pdfs $p, g \in \calF_d$ satisfies $\KL[p,g] \geq 2 \| p - g\|_{TV}^2$ .
\end{lemma}

\begin{lemma} 
\label{lemma:tvholder}
Given functions $\Psi_0 \in \mathrm{L}^{\infty}$ and $p,g \in
\Lp$,  it holds that $\langle \Psi_0, p-g\rangle \leq 2 \| \Psi_0
\|_{\infty} \| p-g \|_{TV}$.
\end{lemma}
\begin{proof}
  See Appendix~\ref{app:Centralized}.
\end{proof}

\subsection{Convergent stochastic sequences}
To aid with the convergence analysis of the proposed algorithms, 
we next introduce known sufficient conditions for convergence of sequences.

\begin{definition}
  A filtration is an increasing nested sequence of $\sigma$-algebras, 
  $\calZ_1 \subseteq \calZ_2 \subseteq \dots $,
  where $\calZ_t = \sigma(X_{1}, \dots, X_{t})$.
  If $S_{t}$ is $\calZ_t$-measurable, then $\crl{S_{t}}$ is $\crl{\calZ_t}$-adapted.
\end{definition} 

\begin{definition}
  A $\crl{\calZ_t}$-adapted sequence $\{X_t\}$ on the probability space
  $(\Omega, \crl{\calZ_t}, \mathbb{P})$ is a \emph{martingale difference
    sequence} if $\expect\brl{|X_t|} < \infty$ and
  $\expect\brl{X_t|\calZ_{t-1}} = 0$, a.s..
\end{definition}

\begin{lemma}[Gladyshev's Lemma~{\cite[Lemma~2.2.9]{BTP:87}}]
\label{lemma:gladyshev}
  Let $\{X_t\}_{t=1}^\infty$ be a sequence of non-negative random
  variables such that $\expect\brl{X_1} < \infty$ and $\expect
  \brl{X_{t+1} | X_1, \ldots, X_t} \leq (1+ \delta_t) X_t +
  \epsilon_t$, where $\delta_t$, $\epsilon_t$ are non-negative
  deterministic sequences with $\sum_{t=1}^{\infty} \delta_t <
  \infty$, $\sum_{t=1}^{\infty} \epsilon_t < \infty$. Then, $X_t$
  converges almost surely to some random variable $X_{\infty}\geq 0$.
\end{lemma}


\begin{lemma}[{\cite[Thm.~2.18]{PH-CHH:14}}]
\label{lemma:stronglaw}
Let $S_t := \sum_{\tau=1}^t X_\tau$ be a martingale with respect to
the filtration $\calZ_t$ on a probability space $(\Omega,
\crl{\calZ_{t}}, \mathbb{P})$. Let $\crl{\beta_t}_{t=1}^\infty$ be a
non-decreasing sequence of positive numbers with $\lim_{t \to \infty}
\beta_t = \infty$. If $\sum_{t=1}^\infty \beta_t^{-p} \expect\brl{
  |X_t|^p | \calZ_{t-1}} < \infty$ a.s. for some $p \in [1,2]$, then
$\lim_{t\to\infty} \beta_t^{-1} S_t = 0$ almost surely.  
\end{lemma}

\section{Centralized Estimation} 
\label{sec:centralized}
We begin our discussion with designing and analyzing an estimation
algorithm in the centralized setting, as this provides the necessary
components for upcoming sections.  To obtain an iterative update in
$\calF_d$, we apply the SMD algorithm to minimize the objective
in~\eqref{eqn:c_objective_sum}.  Then, we prove the convergence of
this algorithm to the set $\calF^\star$ composed of pdfs minimizing
the objective defined in~\eqref{eqn:c_objective}.

\subsection{Centralized SMD algorithm}
We define $\KL [p,p_t]$ as the KL-divergence between
$p,p_t \in\calF_{d}$ (c.f.~Sec.~\ref{sec:functional}).  The
generalized SMD algorithm iteratively minimizes the objective
in~\eqref{eqn:c_objective_sum} to generate pdf~$p_{t+1}$ as,
\begin{equation}
\label{eqn:cSMD}
\scaleMathLine[0.89]{p_{t+1} \in \arg\min_{p \in \calF_d} 
  \left\{ \alpha_t \left\langle \frac{\delta F_t}{\delta p}[p_t], 
      p \right\rangle + \KL [p,p_t] \right\}.}
\end{equation}
Let us define the
term~$J_t[p,p_t] = \alpha_t \langle \frac{\delta F_t}{\delta p}[p_t],
p \rangle + \KL[p, p_t]$ as the shorthand for the minimization
objective at each iteration.
The functional~$J_t[p, p_t]$ is convex in pdf $p$ as it is a linear
combination of a convex entropy and linear functionals.  The SMD
algorithm is guaranteed to optimize any convex functional $F$ using
noisy gradients if the steps $\alpha_t$ satisfy the following
condition:
\begin{assumption}[Robbins-Monro condition]
  \label{assume:summable}
  The positive step-size sequence $\{\alpha_t\}$ is square-summable
  but not summable i.e. $\sum_{t=0}^{\infty} \alpha_t = \infty$ and
  $ \sum_{t=0}^{\infty} \alpha_t^2 < \infty$.
\end{assumption}  

\begin{proposition}
\label{prop:cSMD}
The closed-form solution to~\eqref{eqn:cSMD} is,
\begin{align}
\label{eqn:csmd_updates}
  p_{t+1} = \frac{1}{Z_t}
  \exp\prl{- \alpha_t \frac{\delta F_t}{\delta p}[p_t]}p_t,
\end{align}
where
$Z_t = \int \exp\prl{- \alpha_t \frac{\delta F_t}{\delta p}[p_t]}p_t$.
\end{proposition}
\begin{proof}
  See Appendix~\ref{app:Centralized}.
\end{proof}
For our specific choice of
$F_t[p] = -\langle \log \qz{}(z_t | \hypspace), p \rangle$,
\begin{align*}
  \frac{\delta F_t}{\delta p}[p_t]
                                    & = -\log \qz{}(z_t | \,\hypspace).
\end{align*}
Applying Proposition~\ref{prop:cSMD} leads to the following pdf
update,
\begin{align}
\label{eqn:cSMD_update}
p_{t+1} = \qz{}(z_t | \hypspace)^{\alpha_t} p_t/
\left(\int \qz{}(z_t | \hypspace)^{\alpha_t} p_t\right).
\end{align} 

\begin{assumption}[Positive initial probability]
\label{assume:positive}
The prior pdf at initial time step is strictly positive, i.e.,
$p_{0} > 0, \forall \hypspace$.
\end{assumption}
Assuming a positive initial pdf is sufficient to estimate any possible pdfs. 
A weaker assumption would require that the positive domain of pdf~$p^\star$ is contained within the positive domain of the prior $p_{0}>0$.

\subsection{Almost sure convergence with centralized SMD}
In this subsection, we study the convergence properties of the
estimated pdf $p_t$ to the optimal set~$\calF^\star$ under the centralized SMD algorithm. 
The first theorem proves that the KL divergence between any optimal pdf $p^\star \in \calF^\star$ 
and  $p_t$ converges to a constant, while the second result shows that this constant is zero. 
To begin, we introduce the divergence neighborhood of a set of pdfs as,
\begin{definition}[$\epsilon$-Divergence neighborhood]
\label{def:div_neighbor}
The $\epsilon$-neighborhood $\Be(\calF^\star, \epsilon)$ of the pdf set $\calF^\star$
is given as,
\begin{align*}
  \Be(\calF^\star, \epsilon) = \left\{p \in \calF_d| \min_{p^\star \in
      \calF^\star} \KL[p^\star, p] \leq \epsilon \right\} .
\end{align*}
\end{definition}
Here, we choose the order of the pdf arguments in the divergence term to match the 
unknowns in the objective function. This definition aids the upcoming analysis.
The proofs to the following claims are in \textbf{Appendix~\ref{app:Centralized}}.
\begin{proposition}
\label{prop:optim_KL}
Let pdf $p_{t+1}$ in \eqref{eqn:csmd_updates} minimize the
optimization argument $J_t[p, p_t]$ with arbitrary pdf $p \in \calF_d$
in \eqref{eqn:cSMD}, then the change in divergence in each update is
upper bounded as,
\begin{align*}
\KL [p, p_{t+1}] - \KL [p, p_{t}] 
\leq \alpha_t & \langle \frac{\delta F_t[p_t]}{\delta p},
  p - p_t \rangle + 2 \alpha_t^2 L^2.
\end{align*}
\end{proposition}


This previous result relies on the sampled gradient of the objective $\frac{\delta}{\delta p}F_t[p]$, that we next relate to its
expected value.
\begin{lemma}
\label{lemma:grad_expect}
Under Assumption \ref{assume:gradbound}, the gradient of the expected
value of objective functional defined in \eqref{eqn:c_objective1} is equal to the expectation of its
gradient, i.e. 
$\frac{\delta f}{\delta p}[p_t] = \underset{z_t \sim
  \qz{}^\star}{\expect} \frac{\delta F_t}{\delta p}[p_t]$.
\end{lemma}

Next, we will employ Proposition~\ref{prop:optim_KL} to upper bound
the divergence from the estimate to the optimal set $\calF^\star$ to
show convergence of this divergence term.
\begin{theorem}
\label{thm:centralized_distance}
Under Assumptions \ref{assume:independence}-\ref{assume:positive}, the
KL-divergence~$\KL[p^\star, p_t]$ between any minimizer $p^\star \in
\calF^\star$ and the estimate $p_t$ generated by
the SMD algorithm in~\eqref{eqn:cSMD_update} converges almost
surely to some finite value.
\end{theorem}


Next, we use Theorem~\ref{thm:centralized_distance} to prove almost
sure convergence of the divergence terms arbitrarily close to zero.
\begin{theorem} 
\label{thm:centralized_pdf}
Under Assumptions~\ref{assume:independence}-\ref{assume:positive}, the
pdf sequence $\{p_t\}$ generated by the SMD algorithm
in~\eqref{eqn:cSMD_update} converges almost surely to an
$\epsilon$-divergence neighborhood
$\mathcal{B}(\calF^\star, \epsilon)$ around the set of minimizers in
$\calF^\star$ for any $\epsilon > 0$.
\end{theorem}


Theorem~\ref{thm:centralized_pdf} establishes the convergence of the
pdf iterates in centralized SMD algorithm to $\epsilon$-divergence
neighborhood of the optimal set $\calF^\star$.  We have shown this
result for adaptive learning rate $\alpha_t$ satisfying Robbins-Monro
condition.  While this is sufficient to prove almost sure convergence
of the centralized update in \eqref{eqn:cSMD_update}, we can leverage
the existence of an adaptive learning rate to prove that the objective
function converges at the rate
$O(1/\sqrt{T})$. 

\begin{theorem}
  \label{thm:centralized_rate}
  For a natural filtration of observations
  $\calZ_{t-1} = \sigma_t(z_1, \dots, z_{t-1})$, and the adaptive step
  sizes $\alpha_t < (f[p_t]- f[p^\star])/2L^2$, the expected objective
  function satisfies,
  \begin{align}
    f[\bar{p}_t] - f[p^\star] \leq \sqrt{\frac{8L^2 \KL[p^\star, p_0]}{t}},
  \end{align}
  where $\bar{p}_t = \frac{1}{t} \sum_{k=1}^t p_k$ and $p^\star$ minimizes $f[p]$.
\end{theorem}


In this section, we have established the weak convergence of pdf estimates in a centralized setting 
for the proposed SMD algorithm with square summable step sizes. Additionally, we have shown existence of a decaying 
step size that achieves a $\mathcal{O}(1/\sqrt{t})$ convergence rate.

\section{Distributed Estimation}
\label{sec:distSMD}

In this section, we present and analyze a distributed estimation
algorithm in which each agent updates a pdf for all variables and
shares it with one-hop neighbors.  While our proposed algorithm is
similar to \cite{CAA-AO-AN:22, PP-NA-SM:22}, our novel analysis
demonstrates almost sure convergence to a common pdf in a functional
space.  This analysis is integral for the subsequent analysis of the
marginal distributed algorithm in Section~\ref{sec:partcon}.

\subsection{Distributed estimation problem}
We start by setting up a distributed estimation problem, noting the
separability of the objective function $F$
in~\eqref{eqn:c_objective_sum} across agents.  Since agents sample
$z_i$ independently, the likelihood and the data-generating density
are separable across agents as,
\begin{align}
  \label{eqn:independence}
  \qz{}(z|\hypspace) = \prod_{i=1}^n \qz{i}(z_i|\hypspace_i), \quad
  \qz{}^\star(z_t) = \prod_{i \in \nodes} \qz{i}^\star(z_{i,t}).
\end{align}
Thus, each component of $F$ can be expressed in terms of the
likelihood of the agents' private observations. That is, the
centralized objective in~\eqref{eqn:c_objective1} separates across
agents as $F_t[p] = \sum_{i=1}^n F_{i,t}[p_i]$, where,
\begin{align}
\label{eqn:d_objective_a}
  F_{i, t}[p_i] = \underset{\hypspace \sim p_i}{\expect} [- \log(\qz{i}(z_{i,t}|\hypspace_i) )].
\end{align}
Here, the expectation is computed using the variables in $\hypspace_i$ 
even though the samples from $p_i$ contain all variables in $\hypspace$.
\begin{problem}[Distributed Estimation]
  \label{prob:dist_est}
  Given observations $z_{i,t}$ and agent likelihoods
  $\qz{i}(z_{i,t}| \hypspace_i)$, for each $i \in \nodes$, find the
  pdf $p_i \in \calF_{d}$ minimizing the sample average approximation
  to the agent objective defined using $F_i$ in
  \eqref{eqn:d_objective_a} as:
  \begin{align}
    \label{eqn:d_objective_sum}
    \min_{p_i \in \calF_d} \left\lbrace \frac{1}{T} \sum_{t=1}^T F_{i,t}[p_i] \right\rbrace , \,\text{s.t. } p_i = p_j, \forall i, j \in \nodes,
  \end{align}
  under the consensus constraint enforcing equal estimates.
\end{problem}

\subsection{Distributed SMD algorithm}
For Problem~\ref{prob:dist_est}, each agent~$i$ learns a copy $p_i$ of
the pdf solution $p \in \calF^\star$.  Taking inspiration from the
centralized setting, we deploy the SMD algorithm at any time $t$ to
compute pdf~$p_{i,t+1}$ based on agent~$i$'s local log-likelihood
samples and a prior mixed with neighbor estimates as,
\begin{align}
  \label{eqn:dSMD}
  \min_{p \in \calF_{d}} & J_{i,t}[p, v_{i,t}] , \; v_{i,t} = \prod_{j \in \nodes_i} \prl{ p_{j,t}}^{A_{ij}}, \\
  & J_{i,t}[p, v_{i,t}] = -\langle \log \qz{i}(z_{i,t} | \hypspace),
p\rangle + \frac{1}{\alpha_{t}} \KL [p, v_{i,t} ]. \nonumber
\end{align}
To achieve consensus, we substitute the prior $p_{i,t}$ with the mixed
pdf~$v_{i,t}$, a geometric average of neighbor estimates
$p_{j,t}$ weighted by terms $A_{ij}$ satisfying Assumption~\ref{assume:dsmat}.  Thus, the
distributed update at agent~$i$ is,
\begin{equation}
\label{eqn:dSMD_update}
p_{i,t+1} = \qz{i}(z_{i,t} | \hypspace_i)^{\alpha_{t}} v_{i,t}/\left(\int \qz{i}(z_{i,t} | \hypspace_i)^{\alpha_{t}} v_{i,t}\right).
\end{equation}

The work in~\cite{CAA-AO-AN:22} makes use of geometrically averaged
neighbor estimates to achieve consensus. They analyze the convergence of probabilities
estimated by this algorithm over compact sets in the domain of
variables $\hypspace$.  With this consensus update, \cite{PP-NA-SM:22}
shows the convergence of the modes of estimated pdfs to the same
optimizer as the centralized case.  Instead of these probability
concentration results to the optimal parameter, we prove almost sure
convergence of the KL-divergence between the estimated and an optimal
pdf in $\calF^\star$ defined over the continuous domain.

Our \textbf{analysis strategy} first studies the relative change of
the algorithm mixing-step with respect to the previous algorithm
iterate with respect to a reference pdf
(cf.~Section~\ref{sec:mixing-anal}), then provides summable
upper-bounds for various sequential differences
(cf.~Section~\ref{sec:mixed-update}), then uses these to eventually
prove convergence to the optimal probability density $p^\star$
(cf.~Section~\ref{sec:asDSMD}). In what follows, the expected
value of centralized and agent-specific objectives are,
\begin{align*}
  f[p] = \underset{z_t \sim \qz{}^\star (z_t)}{\expect}
  F_t[p], \,
  f_i[p_i] = \underset{z_{i,t} \sim \qz{i}^\star (z_{i,t})}{\expect} 
\, F_{i,t}[p_i],
\end{align*}
and their derivatives as $\frac{\delta f}{\delta p}$ and
$\frac{\delta f_i}{\delta p_i}$.  By the linearity of the
expectation operator, it follows that $f[p] = \sum_{i=1}^n f_i[p]$.
The proofs to our claims are 
presented in \textbf{Appendix~\ref{app:distributed}}.
\subsection{Analysis of probability-mixing steps}\label{sec:mixing-anal}
We first analyze the convergence characteristics of the mixing step;
that is the behavior of $v_{i,t}$ relative to $p_{i,t}$ for all $t$
and $i$.  This analysis entails the definition of a consensus manifold
for the estimated pdfs.
\begin{definition}
  \label{def:cm}
  The \textbf{consensus manifold} for a connected graph $\calG$
  satisfying Assumption~\ref{assume:dsmat} is a set $\calM$ of pdfs
  that are a.e. equal to some pdf $\bar{p} \in \calF_d$,
  \begin{align*}
    \calM = \crl{\crl{p_{i,t}}_{i=1}^n | \sum_{i=1}^n \KL [\bar{p}, p_{i, t}] = 0, p_{i,t} \in \calF_{d}, \bar{p} \in \calF_d}.
  \end{align*}
\end{definition}
Note that the estimated pdfs lying on the consensus manifold are equal
a.e.  Now, we show that the divergence between any pdf $p\in \calF_d$
to the estimated pdfs $\crl{p_{i,t}}$ decreases under the mixing step
in \eqref{eqn:dSMD}, unless the pdfs lie on the consensus manifold.
This result is critical to work with $\epsilon$-divergence
neighborhoods around optimal pdfs.
\begin{proposition}
\label{prop:d_dist_dec}
The sum of divergences between an arbitrary pdf $p \in \calF_d$
to the estimates $p_{i,t} \in \calF_d$ upper bounds 
the divergence sum to the agent geometric averages $v_{i,t} = \frac{1}{Z^v_{i,t}} \prod_{j=1}^n p_{j,t}^{A_{ij}}$ 
with normalization factor $Z^v_{i,t} = \int \left(\prod_{j=1}^n p_{j,t}^{A_{ij}}\right) d\hypspace $ as, 
\begin{align*}
  \sum_{i=1}^n \KL [p, v_{i, t}] & \leq \sum_{i=1}^n \KL [p, p_{i, t}],
\end{align*}
with equality holding iff pdfs $\crl{p_{i,t}}$ lie on the consensus manifold.
\end{proposition}

The previous proposition establishes that the sum of divergences from 
an arbitrary pdf to agent estimates decreases with the mixing step.
The next proposition establishes a geometric contraction rate for the
consensus step of the algorithm to the network wide average
$p_t \propto \prod_{i=1}^n p_{i,t}^{1/n}$.
\begin{proposition}(See~\cite[Theorem~$5$]{SB-SJC:14})
\label{prop:attract}
Under Assumption~\ref{assume:dsmat}, we
have~$\| v_{i,t}(\vect{x}) - p_t(\vect{x})\|_{TV} \leq \sigma(A)
\| p_{i,t}(\vect{x}) - p_t(\vect{x})\|_{TV}$ with~$\sigma(A)<1$.
\end{proposition}
This allows us to later prove distributed estimation guarantees 
similar to Theorem~\ref{thm:centralized_distance}.
Based on the consensus results, we continue to analyzing objective functional evaluated 
at probability estimates and their geometric average.

\subsection{Probability-mixing and algorithm iterate gaps} \label{sec:mixed-update}
In this subsection, we prove the sequence of 
total variation (TV) distance between terms after likelihood updates are summable. 
Summability of positive sequences \cite{BF-SG:22} implies vanishing terms, 
and this property aids our convergence results in the next Subsection~\ref{sec:asDSMD}.
More specifically, we upper bound TV distances between the mixed 
pdf $v_{i,t}$, agents' next estimate $p_{i,t+1}$,
and network wide-averages $p_t, p_{t+1}$.
  Next, we upper bound the TV distance between
the mixed prior~$v_{i,t}$ and estimate $p_{i,t+1}$.
\begin{proposition}
\label{prop:dist_bound}
Under Assumption \ref{assume:gradbound}, the pdf~$p_{i,t+1}$ minimizing the distributed objective
$J_{i,t}[p, v_{i,t}]$ in~\eqref{eqn:dSMD} satisfies,
\begin{align*}
\alpha_t L \| v_{i,t} - p_{i,t+1} \|_{TV} \leq  \frac{\alpha_t^2 L^2}{2}.
\end{align*}
\end{proposition}

Note that the upper bound in Proposition~\ref{prop:dist_bound} relies on the boundedness of log-likelihood from the Assumption~\ref{assume:gradbound}.
We show that a similar bound exists for the geometric average $p_{t} \propto \prod_{i=1}^n p_{i,t}^{1/n}$, a proxy for centralized estimate.
\begin{proposition}
\label{prop:dist_bound_f}
Let Assumptions~\ref{assume:dsmat}-\ref{assume:gradbound} hold.
Following the distributed SMD algorithm in~\eqref{eqn:dSMD}, the update to the geometric 
average $p_{t}= \prod_{i=1}^n p_{i,t}^{1/n}/Z_t$ for normalization factor $Z_t = \left(\int \prod_{i=1}^n p_{i,t}^{1/n} d\hypspace \right)$ 
satisfies~$ \| p_{t} - p_{t+1} \|_{TV} \leq \alpha_t L/2$.
\end{proposition}

The presence of $\alpha_t$ in the upper bound limits the
relative error between network estimates at each time step.  
Now, we study the convergence of the TV distances between the agent estimates
$p_{i,t}$ to the geometric average $p_t$ and the true pdf $p^\star$.
To establish vanishing distances, we bypass the need for a geometric
rate of contraction like Proposition~\ref{prop:attract} by showing the
summability of this sequence with distance terms. The following
technical result relates the difference between objective functions at
these pdfs to the TV distance.  

\begin{proposition}
  \label{prop:summable_c}
  For the pdf estimates in~\eqref{eqn:dSMD_update}, 
  the sum of objectives is upper bounded as
  $\alpha_t \sum_{i=1}^n (f_i[p^\star] -  f_i[v_{i,t}]) \leq 2 \sigma \alpha_t L \sum_{i=1}^n \| p_t - p_{i,t} \|_{TV} $ for $\sigma < 1$.
\end{proposition}


Now, we show that the upper bounding distance between the average
$p_t$ and estimate $p_{i,t}$ in Proposition~\ref{prop:summable_c} is
summable.  With decaying step-size $\alpha_t$, this implies that the
individual estimates would converge to their geometric average.  In
comparison to the last subsection, here the averages include the
likelihood updates across time.
\begin{proposition}
\label{prop:summable_cTV}
Under Assumptions~\ref{assume:dsmat}-\ref{assume:gradbound}, the updates in~\eqref{eqn:dSMD} 
lead to a summable sequence of distance terms 
$\alpha_t L \sum_{i=1}^n \| p_t - p_{i,t} \|_{TV}$
between the geometric average $p_t$ and agent estimates.
\end{proposition}



\subsection{Almost sure convergence with distributed SMD}
\label{sec:asDSMD}
Aided by the preliminary results, we prove the convergence of the distributed estimation algorithm with the next two theorems.
The first theorem shows almost sure convergence of the KL-divergence between the estimated and true pdf to a finite positive value, 
and the next one proves existence of a subsequence of pdf estimates to the optimal set.

\begin{theorem}
\label{thm:distributed_distance}
Under Assumptions~\ref{assume:independence}-\ref{assume:positive}, the divergence functional~$\sum_{i=1}^n \KL[p^\star, v_{i,t}]$ of the mixed pdf sequence~$\{v_{i,t}\}_{i \in \nodes}$ generated 
via distributed SMD algorithm in~\eqref{eqn:dSMD} almost surely converges to some non-negative value.
\end{theorem}


Next, we show that the divergence sum in Theorem~\ref{thm:distributed_distance} converges arbitrarily close to zero.
\begin{theorem} 
\label{thm:distributed_pdf}
Under Assumptions~\ref{assume:independence}-\ref{assume:positive}, the sequence~$\{v_{i,t}\}$  generated by applying distributed SMD algorithm in~\eqref{eqn:dSMD} converges almost surely to $\epsilon$-divergence neighborhood~$\mathcal{B}(\calF^\star, \epsilon)$ around optimal pdf set~$\calF^\star$ for any~$\epsilon > 0$.
\end{theorem}


This proves that the pdf estimates generated by the proposed algorithm in a connected network 
almost surely converge to the set of optimal pdfs.
Based on the proposed distributed estimation algorithm and its analysis,
we will extend our discussion to estimating marginal pdfs over subset of variables $\hypspace$ in connected networks. 

\section{Distributed Marginal Estimation}
\label{sec:partcon}

In several inference problems over networks, the data likelihood at a node depends on the state of that node and its one-hop neighbors, rather than the entire network. 
Motivated by this, this section extends the distributed SMD algorithm to find marginal densities defined over a relevant subset of variables at each node. 
First, we derive a distributed estimation objective, then modify the algorithm to store and update pdf over node-specific variable sets, and finally discuss the convergence properties.

\subsection{Distributed Marginal Estimation Problem}
We aim to estimate the marginal density of local subsets of variables $\hypspace_i$ at each agent $i$. 
This is enabled by Assumption~\ref{assume:independence} that establishes the independence among the observations $z_{i,t}$ generated using likelihoods $\qz{i}(z_{i,t} | \hypspace_i)$ at agents $i \in \nodes$. 
Let us denote the set of variables common to agents $i,j$ as $\calX_{ij} = \hypspace_i \cap \hypspace_j$.
For a well-posed estimation problem,  we assume the existence of a communication pathway between agents $i,j$ estimating any common variables in $\hypspace_{ij}$.

\begin{assumption}[Marginal consensus]\label{assume:marginal_network}
The set of agents $\nodes(\vect{x}_i) \subseteq \nodes$ estimating the same variable $\vect{x}_i \in \real^{d_i}$ induces a connected subgraph $\graph(\vect{x}_i)$ of $\graph$ with edge set $\edges(\vect{x}_i) = \crl{(j,k) \in \edges| \forall j, k \in \nodes(\vect{x}_i)}$.
\end{assumption}

For a given communication network, the problem of assigning connected subgraphs to estimate particular variables is NP-hard, with a feasible solution presented in \cite{PP-NA-SM:19}. We will leverage this assumption to design our marginal estimation algorithm, and show that it achieves consistent estimates on the relevant subspaces.


We follow the distributed SMD derivation in Section~\ref{sec:distSMD} to distribute the centralized estimation objective in \eqref{eqn:c_objective} along the agents' independent observations. 
We first drop the entropy term unrelated to the optimization argument of the objective in \eqref{eqn:c_objective}. 
Then, the observational independence in \eqref{eqn:independence} allows us to define objective functionals of marginal pdfs 
$p_i(\hypspace_i)$ integrated along individual observations as,
\begin{align*}
  & \min_{p} \underset{\hypspace \sim p}{\expect}[\KL
  [\qz{}^\star(z_{1:n}), \qz{}(z_{1:n}|\hypspace)] ] \nonumber \\
  & = \min_{p} \underset{\hypspace \sim p}{\expect} \sum_{i \in \nodes} 
  \left[ \int_{ z_{1:n}} - \qz{}^\star(z_{1:n}) \log(\qz{i}(z_i|\hypspace_i)) \right] \nonumber \\
  & = \min_{p} \sum_{i \in \nodes} \underset{\hypspace_i \sim
  p_i}{\expect} \left[ \int_{ z_{i}} - \qz{i}^\star(z_{i}) \log(\qz{i}(z_i|\hypspace_i))
  \right]
  \nonumber \\
& = \sum_{i \in \nodes} \min_{p_i} \underset{\hypspace_i \sim p_i}{\expect} \, \underset{z_{i} \sim q_i^\star}{\expect} [-
\log(\qz{i}(z_{i}|\hypspace_i)) ] = \min_{p} f[p],
\end{align*}
%
where each pdf $p_i(\hypspace_i) \in \calF_{\mathfrak{d}_i}$ is a marginal of the joint pdf $p(\hypspace)\in \calF_d$ and $\mathfrak{d}_i$ is the dimension of $\hypspace_i$.
Making the objective $f[p]$ distributed along marginals $p_i(\hypspace_i)$ is possible with additional equality constraints on the shared states $\hypspace_{ij}$. These constraints are represented as agreement on marginal pdfs $p_i, \forall i \in \nodes$ over shared variables as,
\begin{align*}
  \int p_i(\hypspace_i) d\vect{x}|_{\vect{x} \in \hypspace_i \backslash
    \hypspace_{ij}} = \int p_j(\hypspace_j) d\vect{x}|_{\vect{x} \in \hypspace_j
    \backslash \hypspace_{ij}} , \, \forall (i, j) \in \edges,
\end{align*}
where $\int p_j d\vect{x}|_{\vect{x} \in \hypspace_j \backslash \hypspace_{ij}}$
defines an integral over all variables in the set $\hypspace_j \backslash \hypspace_{ij}$.
As before, a finite objective allows using Fubini-Tonelli's theorem to switch the order of expectations. Along with a sample-average approximation of the integral over data in~$\{z_{i,t}\}$, the online objective is expressed as,
\begin{align*}
  \min_{p} f[p] & = \sum_{i \in \nodes}
  \min_{p_i} f_i[p_i], \\
  f_i[p_i] & = \underset{z_{i} \sim q_i^\star}{\expect} \,
  \underset{\hypspace_i \sim p_i}{\expect} [- \log(\qz{i}(z_{i}|\hypspace_i)) ]
  \nonumber \\
  & \approx \sum_{i \in \nodes} \min_{p_i} \sum_{t=1}^T \underset{\hypspace_i \sim p_i}{\expect} \, [- \log(\qz{i}(z_{i, t}|\hypspace_i)) ]. \nonumber
\end{align*}
Thus, the distributed objective at time $t$ becomes,
\begin{align}
\label{eqn:d_objective1}
F_{i,t}[p_i] = \underset{\hypspace_i \sim
  p_i}{\expect} \, [- \log(\qz{i}(z_{i, t}|\hypspace_i))].
\end{align}

\begin{problem}[Distributed marginal estimation]
  Given observations $z_{i,t}$ and agent likelihoods $\qz{i}(z_{i,t}| \hypspace_i)$
  at any agent $i \in \nodes$,
  find pdf~$p_i \in \calF_{\mathfrak{d}_i}$ minimizing: 
  \begin{align}
    \label{eqn:d_objective_sum_m}
    \min_{p_i \in \calF_{\mathfrak{d}_i}} \left\lbrace \frac{1}{T} \sum_{t=1}^T F_{i,t}[p_i] \right\rbrace ,
    \text{s.t. } p_i(\hypspace_{ij}) = p_j(\hypspace_{ij}),
  \end{align}
  for all agents $ i, j \in \nodes$ over the marginal pdfs $p_i(\hypspace_{ij}) = \int p_i(\hypspace_{i})d\vect{x}|_{\vect{x} \in \hypspace_i \backslash
  \hypspace_{ij}}$.
\end{problem}

\subsection{Distributed Marginal SMD Algorithm (DMSMD)}

Similar to Sec.~V, each agent $i$ applies the SMD algorithm to its local objective in \eqref{eqn:d_objective_sum_m},
with two exceptions.
Firstly, the agents locally estimate a pdf over relevant variables~$p_{i,t}(\hypspace_i)$,
and secondly, they enforce marginal consensus constraint equating agent~$i$'s 
marginal~$p_{ij} = \int_{\hypspace_i \backslash \hypspace_{ij}} p_i$ to agent~$j$'s marginal~$p_{ji}$.  
As before, the likelihood update follows from the Gateaux derivative~$\frac{\delta}{\delta p_i} F_{i,t}[p_i] 
= - \log(\qz{i}(z_{i, t}|\hypspace_i))$ as computed for linear functional in Proposition~\ref{prop:variations}. 

Each agent $i$ co-estimates some variables with its one-hop neighbors. 
Therefore, it merges neighbor~$j$'s information over shared variables $\hypspace_{ij}$ to own estimate on distinct variables $\hypspace_i \backslash \hypspace_{ij}$.
The incoming density over the shared variables is $p_{ji,t}(\hypspace_{ij})$ and the self-conditional density at agent $i$ over distinct variables w.r.t. neighbor $j$ is given by $p_{i,t}(\hypspace_i \backslash \hypspace_{ij}| \hypspace_{ij})$.
The marginal agreement is enforced with geometric averaging on 
self-conditional and neighbor-marginals product~$\tilde{p}_{ji,t}$ as,
\begin{align}
  \label{eqn:m_mixed}
  v_{i,t} & = \frac{1}{Z_{i,t}^v} \prod_{j \in \nodes_i} \left(\tilde{p}_{ji,t} \right)^{A_{ij}}, \, Z_{i,t}^v = \int \prod_{j \in \nodes_i} \left(\tilde{p}_{ji,t} \right)^{A_{ij}}, \\
  & \tilde{p}_{ji,t} = p_{i,t}(\hypspace_i \backslash \hypspace_{ij}| \hypspace_{ij}) p_{ji,t}(\hypspace_{ij}), \\
  & p_{ji,t}(\hypspace_{ij}) = \int p_{j,t} (\hypspace_{j}) d\vect{x}|_{\vect{x} \in \hypspace_j
    \backslash \hypspace_{ij}} .\nonumber
\end{align}
Now, applying the SMD algorithm with the gradient defined as negative log-likelihood sample in Section~\ref{sec:distSMD},
and the mixed pdf $v_{i,t}$ in \eqref{eqn:m_mixed}, the marginal consensus estimation is performed as follows,
\begin{align}
\label{eqn:dmsmd}
& p_{i,t+1}(\hypspace_i) \in \argmin_{p \in \calF_{\mathfrak{d}_i}} J_{i,t}[p, v_{i,t}], \\
& J_{i,t}[p, v_{i,t}] =  \left\{ \alpha_t\left\langle \frac{\delta F_{i,t}}{\delta p}[p_{i,t}], p \right\rangle + \KL [p, v_{i,t}] \right\}. \nonumber 
\end{align}


We summarize the updates for agent $i$ at time $t$ in \textbf{Algorithm~\ref{algo:nonlin_algo}}. 
The algorithm consists of \textbf{edge merging}, \textbf{geometric pooling}, \textbf{likelihood update} and \textbf{message generation}.
At each agent, these steps correspond to self-conditional and neighbor-marginal products, their weighted average, 
Bayesian likelihood update, and generation of marginal densities for its neighbors.


In comparison to the distributed algorithm in Section~\ref{sec:distSMD}, 
estimating the marginals reduces the set of stored variables at agent $i$ to $\hypspace_i$ with dimensions $\mathfrak{d}_i < d$. 
The size of the communicated messages reduces from a pdf in $\calF_d$ over all network variables to a partial set $\hypspace_{ij}$ 
shared between sensors $i, j$. 
Although, each node additionally computes the conditional density. The trade-off between memory and computation depends on the average degree in the network.

Following the previous section on distributed algorithm, our \textbf{analysis strategy} 
first discusses the monotonic convergence of estimates under marginal mixing step to 
an invariant consensus manifold defined later (cf.~Section~\ref{sec:marginal-mixing-anal}).
, and then presents a specific independent variable setting for 
similar results in terms of total variation distances (cf.~Section~\ref{sec:marginal-TV-anal}).  
We use them to establish summable upper-bounds for sequential differences between marginal estimates, 
and eventually prove convergence to the marginals of the optimal probability density $p^\star$ (cf.~Section~\ref{sec:asDMSMD}).
All proofs to the claims in this section are in \textbf{Appendix~\ref{app:marginal}}.

\subsection{Marginal Consensus Analysis}
\label{sec:marginal-mixing-anal}

\begin{algorithm}[t]
  \caption{Marginal density averaging at agent $i$}
  \label{algo:nonlin_algo}
  \small \KwData{estimate $p_{i,t}(\hypspace_i)$, 
      weights $\{A_{ij}\}_{j \in \nodes_i}$, 
      neighbor messages $p_{ji,t}(\hypspace_{ij})$, 
      measurement $z_{i,t}$, 
      measurement model $\qz{i}(z_{i,t}| \hypspace_i)$}
    \tcp{Receive neighbor messages.}  \For{$j
    \in \neighbor{i}$} { Common marginals at neighbors
      $p_{ji,t}(\hypspace_{ij}) = 
      \int_{\hypspace_{j} \backslash \hypspace_{ij}} p_{j,t}(\hypspace_j)$\, }
  
    \tcp{Combine neighbor estimates.}
    \For{$j \in \neighbor{i}$} { Product of $j$'s marginal and $i$'s conditional:\\
    $\tilde{p}_{ji,t} =
    p_{i,t}(\hypspace_i \backslash \hypspace_{ij}| \hypspace_{ij}) p_{ji,t}(\hypspace_{ij})$\,}
  
    Weighted average:~$v_{i,t}(\hypspace_i) := \prod_{j \in
    \neighbor{i}}\tilde{p}_{ji,t}(\hypspace_i)^{A_{ij}}$\,
    
    \tcp{Bayesian update.}
    $p_{i,t+1}(\hypspace_i) = \qz{i}(z_{i,t+1}| \hypspace_i) v_{i,t}(\hypspace_i)$\,
    
\end{algorithm}

In this subsection, we establish the invariance and convergence properties of the marginal consensus steps defined in~\eqref{eqn:m_mixed}. We define a marginal consensus manifold and analyze convergence of the consensus steps to this manifold.

\begin{definition}\label{def:mcm}
The marginal consensus manifold for a graph $\calG$ that satisfies Assumption~\ref{assume:marginal_network} is a set $\calM = \crl{\crl{p_{i,t}}_{i=1}^n | \sum_{i=1}^{n} \KL[\bar{p}_i, p_{i,t}] = 0, p_{i,t} \in \calF_{\mathfrak{d}_i}, \bar{p} \in \calF}$ of marginal pdfs consistent with some joint pdf $\bar{p} \in \calF$.
\end{definition}

The manifold consists of coherent marginals of some joint pdf $\bar{p}$ 
with $p_{i,t} = \bar{p}_i \in \calF_{\mathfrak{d}_i}$ for all agents. 
The following technical result shows that the product of normalization factors of 
mixed pdfs obtained after applying $\eqref{eqn:m_mixed}$ to pdfs in the marginal consensus manifold $\calM$ is $1$.

\begin{proposition}
    \label{prop:marg_equality}
    The product of normalization factors of mixed marginals satisfies $\prod_{i=1}^n Z^v_{i,t} = 1$, where $Z^v_{i,t} = \int \prod_{j=1}^n \left( \tilde{p}_{ji,t} \right)^{A_{ij}} d\hypspace_i$, if and only if the original pdfs $\{p_{i,t}\}$ lie on the marginal consensus manifold $\calM$. 
\end{proposition}


Next, we establish that the sum of KL divergences decreases strictly due to marginal mixing step if the agent pdfs are not on the marginal consensus manifold.

\begin{proposition}
\label{prop:m_dist_dec}
For any pdf $p \in \calF$,  the mixed and original pdfs $\{v_{i,t}\}, \{p_{i,t}\}$, defined in the mixing step \eqref{eqn:dmsmd}, satisfy
\[
\sum_{i=1}^n \KL [p_i, v_{i, t}] \leq \sum_{i=1}^n \KL [p_i, p_{i, t}],
\]
with equality if and only if the original pdfs $\{p_{i,t}\}$ lie on the marginal consensus manifold $\calM$ in Definition~\ref{def:mcm}.
\end{proposition}

To study convergence properties of marginal consensus manifold, denote $p_{i,t}^{(k)}$ as the pdf computed 
at agent~$i$ after the $k$-step marginal mixing from~\eqref{eqn:m_mixed} on estimated pdfs $\crl{p_{i,t}}$. 
For instance, mixed pdf $v_{i,t} = p_{i,t}^{(1)}$.
Based on the consensus properties established in Propositions~\ref{prop:marg_equality}-\ref{prop:m_dist_dec}, 
we show that the pdfs $p_{i,t}^{(k)}$ converge to the marginal pdfs~$\bar{p}_{i,t}$ 
in the marginal consensus manifold $\calM$ of Definition~\ref{def:mcm}.

\begin{proposition}\label{prop:m_mix_eqm}
Repeated application of the marginal consensus steps in \eqref{eqn:m_mixed} to pdfs $\crl{p_{i,t}}$ leads to a limit pdf $\lim_{k \rightarrow \infty} p_{i,t}^{(k)}$ that lies in the marginal consensus manifold in Definition~\ref{def:mcm}.
\end{proposition}

As a consequence of Proposition~\ref{prop:m_mix_eqm}, the estimates after marginal mixing converge to marginals $\bar{p}_{i,t}$ on the manifold $\calM$ consistent with some joint pdf $\bar{p}_t$,
\begin{align}
  \label{eqn:marginal_def}
  \bar{p}_{i,t}(\hypspace_{i}) =  \int_{\hypspace \backslash \hypspace_{i}} \bar{p}_t(\hypspace), \forall i \in \nodes.
\end{align}
Since we do not have an explicit form for the pdf $\bar{p}_t$, we study its properties in
a specific case, where the pdf is independent w.r.t. the variables in $\calX$.


\subsection{Marginal Consensus with Independent Variables}
\label{sec:marginal-TV-anal}
We begin by recalling the mixing properties established for the 
distributed setting in Propositions~\ref{prop:attract}-\ref{prop:dist_bound_f}.
We list the desired properties for $\bar{p}_t$ in the following conjecture and 
prove them for a special case with independence over the variables in $\calX$. 

\begin{conjecture}
  \label{conj:attract}
  For~$v_{i,t}$ defined in~\eqref{eqn:dmsmd} and arbitrary joint pdf $\bar{p}_t$, $\| v_{i,t} - \bar{p}_{i,t} \|_{TV} \leq \sigma(A)  \| p_{i,t} - \bar{p}_{i,t} \|_{TV}$ for $\sigma(A)\in (0,1)$ and $ \| \bar{p}_{t} - \bar{p}_{t+1} \|_{TV} \leq (c-1) \alpha_t L/2$ for some $c>1$.
\end{conjecture}

We consider the following special case where the estimated probabilities $p_{i,t}$ 
are independent w.r.t. each variable $\vect{x} \in \hypspace_i$, the set of 
variables estimated by agent $i$ as,
\begin{align}
  \label{eqn:independent_average}
  & p_{i,t}(\hypspace_i) = \prod_{\vect{x} \in \hypspace_i} p_{i,t}(\vect{x}).
\end{align}
Since Assumption~\ref{assume:marginal_network} assigns a connected subgraph $\graph(\vect{x})$ to any variable $\vect{x}$, the resulting mixed pdf is expressed in terms of independent pdf components at $\vect{x}$ as, 
$$v_{i,t}(\vect{x}) \propto \prod_{j \in \nodes \backslash \nodes(\vect{x})} p_{i,t}(\vect{x})^{ A_{ij} }
  \prod_{j \in \nodes(\vect{x})} p_{j,t}(\vect{x})^{A_{ij}}. $$
Next, we will use this form to show that computing an independent component of agent estimates $p_{i,t+1}(\vect{x})$ involves multiplying the mixed pdf component with a bounded likelihood similar to the Assumption~\ref{assume:gradbound}. 
\begin{lemma}
  \label{lemma:marginal_likelihood}
  Assuming that the mixed pdfs $v_{i,t}$ are independent w.r.t. variable $\vect{x} \in \hypspace_i$, we can represent agent~$i$'s update w.r.t. any variable at time $t$ as,
  \begin{align*}
    p_{i,t+1}(\vect{x}) \propto \qz{i}(z_{i,t}|\vect{x})^{\alpha_t} v_{i,t}(\vect{x}),
  \end{align*}
  with the agent-variable likelihood,
  \begin{align*}
    \qz{i,t}(z_{i,t}|\vect{x})^{\alpha_t} = \int \qz{i}(z_{i,t}|\hypspace_i)^{\alpha_t} \prod_{\vect{y} \in \hypspace_i \backslash \vect{x}} v_{i,t}(\vect{y}) d\hypspace_i \backslash \vect{x}
  \end{align*}
  satisfying $\qz{i,t}(z_{i,t}|\vect{x})^{\alpha_t} \in [e^{-\alpha_t L}, e^{\alpha_t L}]$.
\end{lemma}


Since the estimates $p_{i,t}$ converge to consensus manifold $\calM$,
we now prove a geometric convergence bound for the independent form of $\bar{p}_t$ specified as follows,
\begin{align} 
  & \bar{p}_t(\hypspace) = \prod_{\vect{x} \in \hypspace} \bar{p}_t(\vect{x}),
  \bar{p}_t(\vect{x}) \propto \prod_{j \in \nodes(\vect{x})} p_{j,t}(\vect{x})^{\frac{1}{|\nodes(\vect{x})|}}. \nonumber
\end{align}
\begin{lemma}
\label{lemma:attract}
For~$v_{i,t}$ defined in~\eqref{eqn:dmsmd} with connectivity requirements in Assumption~\ref{assume:dsmat}, additional variable independence assumption, and geometric average $\bar{p}_t$ in \eqref{eqn:independent_average}, we have the TV distance $\| v_{i,t}(\vect{x}) - \bar{p}_{t}(\vect{x})\|_{TV} 
  \leq \sigma \| p_{i,t}(\vect{x}) - \bar{p}_t(\vect{x})\|_{TV}$ with~$\sigma<1$ and $ \| \bar{p}_{t} - \bar{p}_{t+1} \|_{TV} \leq (c-1) \alpha_t L/2$ with $c = 1 + 2 m $.
\end{lemma}
\subsection{Almost Sure Convergence of DMSMD}
\label{sec:asDMSMD}

Using the upper bounds computed for independent densities, we guarantee almost-sure convergence of the iterates to the marginal pdfs.
The presentation here borrows from the distributed SMD algorithm analysis, with the 
following propositions establishing bounded iterate gaps similar to Section~\ref{sec:mixed-update} 
and the final two theorems proving almost sure convergence as Section~\ref{sec:asDSMD}.

As discussed in Section~\ref{sec:mixed-update}, summability of positive upper bounds on the iterate gaps implies their asymptotic convergence to zero.
To this end, the next proposition upper bounds the TV distance between estimates across the likelihood update.
\begin{proposition}
  \label{prop:m_dist_bound}
  The pdf~$p_{i,t+1}$ minimizing~$J_{i,t}[p, v_{i,t}]$ defined in~\eqref{eqn:dmsmd}
  satisfies,~$\| v_{i,t} - p_{i,t+1} \|_{TV} \leq \alpha_t L/2$.
\end{proposition}

For the following analysis, we consider the marginals of the optimal pdf~$p^\star(\hypspace)$ defined as,
\begin{align}
  p^\star_i(\hypspace_{i}) & =  \int_{\hypspace_{i} \backslash \hypspace_{i}} p^\star(\hypspace). 
\end{align}

We next produce an upper bound similar to Proposition~\ref{prop:m_dist_bound}, but for the gap between the objective function evaluated at mixed estimate to true marginal. 
\begin{proposition}
  \label{prop:m_summable_c}
  The term~$\sum_{i=1}^n (f_i[p^\star_i] -  f_i[v_{i,t}])$ is upper bounded by the distances $\sigma(A) \sum_{i=1}^n L \| \bar{p}_{i,t} - p_{i,t} \|_{TV}$.
\end{proposition}
Now, we show summability of the upper bound in Proposition~\ref{prop:m_summable_c} containing the TV distance 
between marginal average $\bar{p}_{i,t}$ to the agent estimate $p_{i,t}$. With square summable $\alpha_t$ \cite{BF-SG:22},
this implies asymptotic convergence of the two pdfs.
\begin{proposition}
\label{prop:m_summable_cTV}
With Proposition~\ref{prop:m_dist_bound} and Conjecture~\ref{conj:attract}, 
the sequence with terms $a_t = \sigma(A) \alpha_t L \sum_{i=1}^n \| \bar{p}_{i,t} - p_{i,t} \|_{TV}$ is summable.
\end{proposition}

\begin{proposition}
\label{prop:m_summable_cl}
Assuming Conjecture~\ref{conj:attract} holds, the sequence~$\alpha_t L \| \bar{p}_t - \bar{p}_t(\hypspace|\hypspace_i)v_{i,t} \|_{TV} $ is summable for any~$i \in \nodes$.
\end{proposition}

Since the estimated pdfs are defined over distinct spaces, we define a neighborhood-based divergence metric relating marginal densities at any agent to the complete pdf.

\begin{definition}
Define the $\epsilon$-neighborhood of a marginal $p_i^\star$ of $p^\star \in \calF^\star$ as:
\[
\Be_i(\calF^\star, \epsilon) = \left\{p_i \in \calF_{\mathfrak{d}_i}| \min_{p^\star \in \calF^\star} 
\KL[p^\star_i, p_i] \leq \epsilon, p_i^\star = \int_{\hypspace \backslash \hypspace_i} p^\star\right\}.
\]
\end{definition}

As seen in prior sections, we employ the preliminary results to 
prove the convergence of the DMSMD algorithm with the next two
theorems. The first theorem shows almost sure convergence
of the KL-divergence between the estimated and marginals of the true pdf to
a finite positive value, and the next one proves that 
the finite value is arbitrarily close to zero.

\begin{theorem}
\label{thm:m_distributed_distance}
Under Assumptions~\ref{assume:independence}-\ref{assume:marginal_network} and Conjecture~\ref{conj:attract}, the divergence functional~$\sum_{i=1}^n \KL[p^\star_i, v_{i,t}]$ of pdf sequences~$\{v_{i,t}\}_{i \in \nodes}$ generated by applying the distributed SMD algorithm in~\eqref{eqn:dmsmd} almost surely converges to some finite value.
\end{theorem}

\begin{theorem} 
\label{thm:m_distributed_pdf}
Under Assumptions~\ref{assume:independence}-\ref{assume:marginal_network} and Conjecture~\ref{conj:attract}, the marginal pdfs~$v_{i,t}$ generated by the distributed marginal algorithm in~\eqref{eqn:dmsmd} for any agent $i  \in \nodes$ converge almost surely to the partial neighborhood~$\mathcal{B}_i(\calF^\star, \epsilon)$ around optimal set~$\calF^\star$ for any~$\epsilon > 0$.
\end{theorem}

\section{Distributed Marginal Gaussian Variational Inference}
\label{sec:dmgvi}

In this section, we specialize the distributed algorithms in
Sections~\ref{sec:distSMD} and~\ref{sec:partcon} for Gaussian
estimates.  At each agent, implementing the proposed algorithms is a
two-step process: mixing the neighbor priors, and updating the
likelihood.  

Marginal mixing requires computing the Gaussian conditionals 
and marginals, and their product and geometric averages.
Algorithm~\ref{algo:marginal_gaussian} computes 
this mixed Gaussian pdf $v_{i,t}(\hypspace_i)$ using the derivations in our 
prior work \cite{PP-NA-SM:20}. 
This algorithm trivially holds for the standard distributed setting 
with conditional-marginal product equal to the neighbor estimate, i.e. $\tilde{p}_{ji,t}=p_{j,t}$.
Here, we represent a Gaussian random variable with mean~$\mu$ and
information matrix $\Omega$ as $\calN(\mu, \Omega^{-1})$, and its
density function as~$\phi(\cdot|\mu, \Omega^{-1})$. 
\begin{algorithm}
  \caption{Marginal density mixing at agent $i$}
  \label{algo:marginal_gaussian}
  \small \KwData{estimate $p_{i,t}=\phi(\hypspace_i|\mu, \Omega^{-1})$, 
    weights $\{A_{ij}\}_{j \in \nodes_i}$, 
    neighbor estimates $p_{j,t}(\hypspace_j)$}
    \tcp{Receive marginals from neighbors.}  
    \For{$j \in \neighbor{i}$} 
      { Compute marginal $p_{ji,t}$
      using \cite[Lemma~1]{PP-NA-SM:20} over $\nodes_{ij}$
        }
    \tcp{Combine neighbor estimates.}
    \For{$j \in \neighbor{i}$} 
      { 
      Use \cite[Lemma~2]{PP-NA-SM:20} to compute conditional pdf $p_{i,t}(X_1|X_2)$ with separate variables $X_1=\hypspace_i \backslash \hypspace_{ij}$ and shared variables $X_2=\hypspace_{ij}$ \\
      Compute $\tilde{p}_{ji,t}(\hypspace_i)$ by multiplying $i$'s conditional with marginal $p_{ji,t}$ using \cite[Proposition~3]{PP-NA-SM:20}
      }
    
    Compute mixed pdf~$v_{i,t}(\hypspace_i)$ using \cite[Lemma~3]{PP-NA-SM:20}
      over $\tilde{p}_{ji,t}(\hypspace_i)$ \,
      
\end{algorithm}

Next, we express an analytic form of the likelihood update step in
Algorithm~\ref{algo:nonlin_algo} assuming that the prior mixed pdf
$v_{i,t}$ and posterior $p_{i,t+1}$ are Gaussian. 
The analytic updates associated with the linear log-likelihood setting 
was presented in \cite{PP-NA-SM:20} is given as,
\revisionAdd{
\begin{lemma}[Likelihood update]
  \label{lemma:gaussian_likelihood_update}
  Let the likelihood density be \( q_i(z_{i,t} | \hypspace_i) = \phi(z_{i,t} | H_i \hypspace_i, V_i) \). 
  Then, the posterior obtained as the product of the likelihood and prior \( \phi(z_{i,t} | H_i \hypspace_i, V_i) 
  \phi(\hypspace_i; \mu, \Omega_i^{-1}) \) is a Gaussian distribution:
  \[
  \calN \left( (H_i^T V_i H_i + \Omega_i)^{-1} (H_i^T V_i z_{i,t} + \Omega_i \mu_i), (H_i^T V_i H_i + \Omega_i)^{-1} \right).
  \]
\end{lemma}
}

For the non-linear log-likelihood
$\qz{i}(z_{i,t+1}|\hypspace_i)$ that does not yield an analytic
update, one can approximate 
the likelihood update using distributed Gaussian variational
inference \cite{PP-NA-SM:23-arxiv} on the mixed pdf
$p^v_{i,t}=\phi(\cdot|\mu^v_{i,t}, \Omega^v_{i,t})$ as, 
\begin{align*}
  \Omega_{i, t+1} & = \Omega^v_{i,t} - \expect_{p_{i,t}^v} [\nabla_{\hypspace_i}^2 \log \qz{i} (z_{i,t+1}|\hypspace_i)], \\
  \mu_{i,t+1} & = \mu_{i,t}^v + (\Omega_{i,t}^v)^{-1} \expect_{p_{i,t}^v} [\nabla_{\hypspace_i} \log \qz{i} (z_{i,t+1}|\hypspace_i)].
\end{align*}

In the partial distributed mapping example explained later, we implement 
this algorithm to estimate Gaussians with diagonal covariance matrices.
Therefore, we present a modified mixing step for the marginal distributed estimation 
algorithm in the following lemma.
\begin{lemma}[Distributed partial diagonalized Gaussian estimation]\label{lemma:dgvis}
  Assume that agent $i$ receives observation $z_{i,t+1}$ with
  likelihood $\qz{i}(z_{i,t+1}|\hypspace_i)$ and neighbor estimates
  $p_{j,t}(\hypspace_j) = \calN(\hypspace_j|\mu_{j,t},
  \Omega_{j,t}^{-1})$ at time $t$.  Upon weighing neighbor opinions
  with elements of matrix $A$, the mean $\mu_{i,t+1}$ and information
  matrix $\Omega_{i,t+1}$ of the pdf $p_{i,t+1}$ is,
\begin{align}
    \label{eqn:dgvi_sample}
    &\tilde{\Omega}_{ji,t} = R_{ij} \Omega_{j,t} + S_{ij} \Omega_{i,t}, 
    \tilde{\mu}_{ji,t} = R_{ij} \mu_{j,t} + S_{ij} \mu_{i,t}\\
    &\Omega_{i,t}^v = \sum_{j \in \nodes} A_{ij} \tilde{\Omega}_{ji,t}, \Omega_{i,t}^v \mu^v_{i,t} = 
    \sum_{j \in \nodes} A_{ij} \tilde{\Omega}_{ji,t}\tilde{\mu}_{ji,t}\nonumber \\
  &\Omega_{i,t+1} = \Omega_{i,t}^v -  \expect_{v_{i,t}} [\nabla_{\hypspace}^2 \log \qz{i} (z_{i,t+1}|\hypspace_i)],  \nonumber \\
  & \mu_{i,t+1} = \mu_{i,t}^v + (\Omega_{i,t}^v)^{-1} \expect_{v_{i,t}} [\nabla_{\hypspace} \log \qz{i} (z_{i,t+1}|\hypspace_i)], \nonumber 
\end{align}
where mixed pdf
$v_{i,t} = \phi(\hypspace_i| \mu_{i,t}^v, \Omega_{i,t}^v)$,
and matrices
$R_{ij} \in \crl{0,1}^{\mathfrak{d}_i \times \mathfrak{d}_j}$ and
$S_{ij} \in \crl{0,1}^{\mathfrak{d}_i \times
  \mathfrak{d}_i}$. Here, $R_{ij}[s_i, s_j] = 1$
where $s_i, s_j$ are indices in agents $i,j$ corresponding to a common
variable. The matrix $S_{ij}$ is a diagonal matrix with $1$ at
variable index distinct from agent $j$.
\end{lemma}

\begin{proof}
  The updates on marginals and distributed consensus follow from
  prior discussion. The matrices $S, R$ match the indices
  between the agents and hypotheses to compute the diagonal information matrices. 
\end{proof}

\subsection*{\revisionAdd{Distributed Relative Localization: An Example}}
\begin{figure*}[t]
  \centering
  \includegraphics[width=0.9\textwidth]{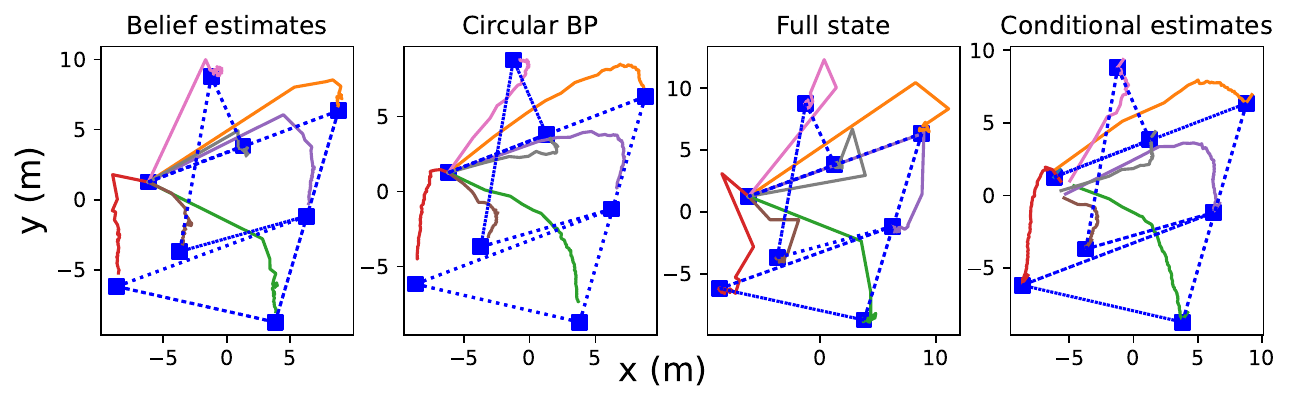}
  \includegraphics[width=0.87\textwidth]{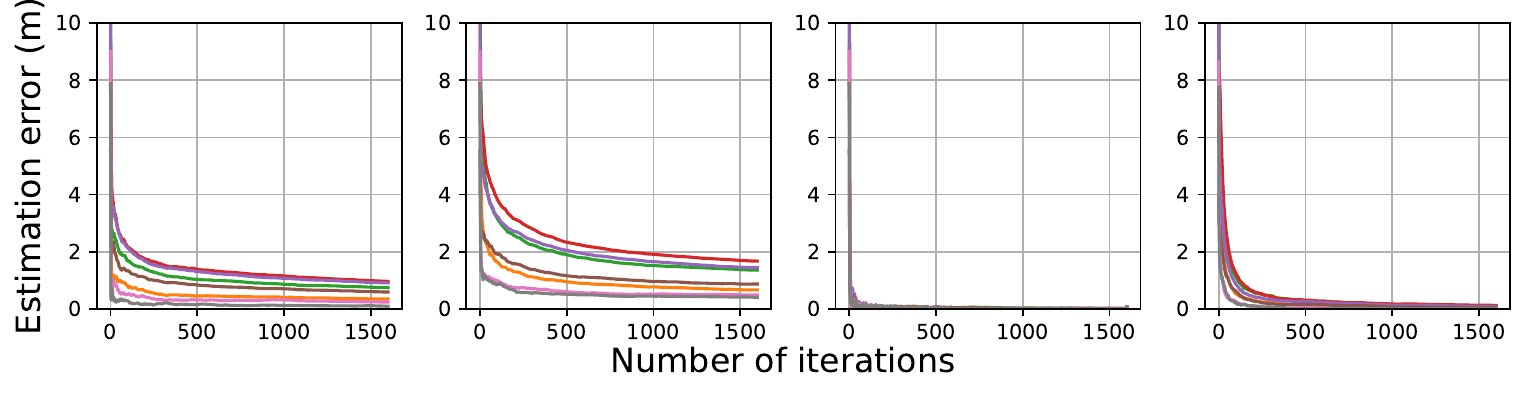}
  \caption{\revisionAdd{Trajectories of estimated node positions 
  $\mu_{i,t}$ in an $8$ agent ring network with true positions shown as blue squares (top). Estimation error $\|\mu_{i,t}-\vect{x}_i\|$ over $1600$ time steps (bottom).}}
  \label{fig:rel_loc_traj_err}
\end{figure*}

\begin{figure*}[t]
  \centering
  \includegraphics[width=0.9\textwidth]{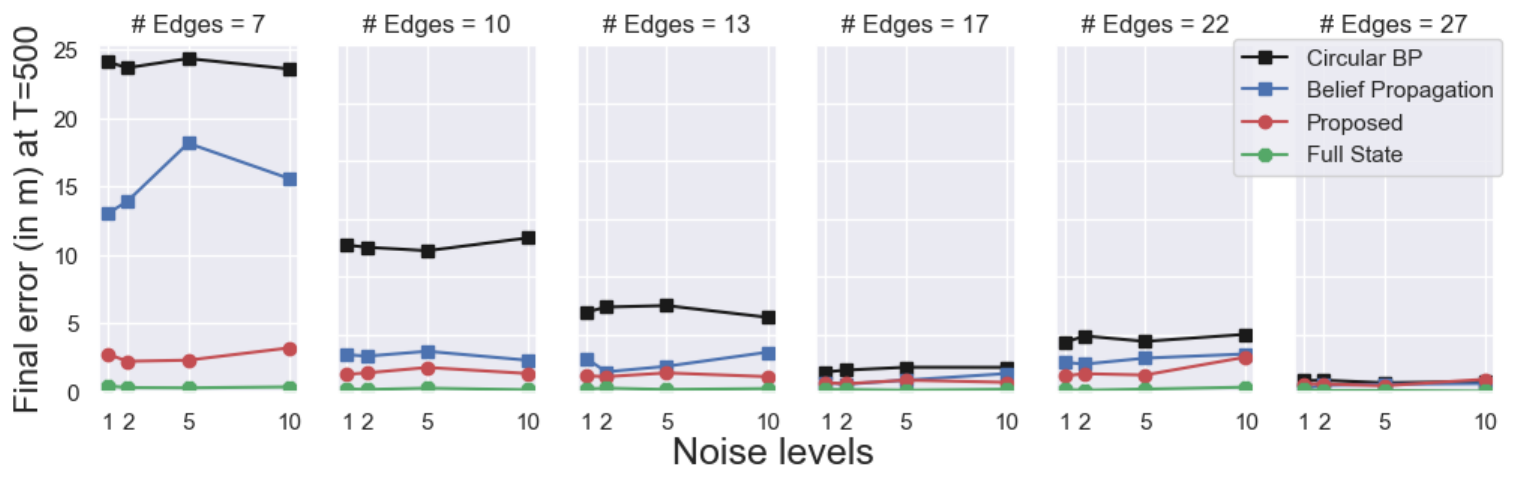}
  \caption{Plots of the $500$-step average localization error, given by $1/n\sum_{i \in \nodes}\Vert \mu_{i,t} - \vect{x}_i\Vert$, using belief propagation, circular belief propagation, the proposed marginal estimation, 
  and full state estimation algorithms in an $8$ node network. The comparisons span measurement noise variances $\Sigma_{ij} = b \In_2$ for $b \in \crl{1,2,5,10}$ and network connectivities ranging from a line graph with $7$-edges to a $27$-edge fully connected one.}
  \label{fig:rel_loc}
\end{figure*}

\revisionAdd{
We consider a network of $n=8$ agents aiming to estimate
their positions $\vect{x}_i \in \real^2$ using noisy relative position measurements.
To ensure a unique solution, we assume the presence of an anchor agent with known 
position at $(0,0)$. 
Each agent $i$ observes the relative position of its neighbor $j$ sampled as $\vect{z}_{ij} \sim \calN(\vect{x}_i - \vect{x}_j, \Omega_{ij}^{-1})$.
The relevant set of variables at agent $i$ is thus given by 
$\hypspace_i = \crl{\vect{x}_j}_{j \in \nodes_i}$.
The combined observation model at agent $i$ for the observations relative to its neighbors 
$\vect{z}_i = \crl{\vect{z}_{ij}}_{j \in \nodes_i}$ is,
\begin{equation}\label{eqn:rel_loc_lik}
  \qz{i}(z_i|\hypspace_i) 
  = \prod_{j\in \nodes_i} \qz{i}(z_{ij}|\vect{x}_i, \vect{x}_j).
\end{equation}
The doubly stochastic matrix $A$ represents agent communication 
 as described in Assumption~\ref{assume:dsmat}.
We first mention the application of our distributed and marginal estimation algorithms, followed by standard and circular BP algorithms.}

\revisionAdd{In the distributed setting, each agent~$i$ maintains a Gaussian distribution $\calN(\mu_{i,t}, 
\Omega_{i,t}^{-1})$ with pdf $p_{i,t}(\hypspace)$ at time step $t$ over all unknown 
variables $\hypspace = [\vect{x}_1^{\top}, \dots, \vect{x}_n^{\top}]^{\top}$.
The corresponding observation model in~\eqref{eqn:rel_loc_lik} is expressed 
in terms of the variable $\hypspace$ as 
$\qz{i}(z_{i,t}|\hypspace) = \calN(z_{i,t}| H_i^{(d)} \hypspace, V_i^{(d)})$
where $H_i^{(d)} \in \real^{d \times nd}$.
Each step in the distributed SMD algorithm in~\eqref{eqn:dSMD_update}  
at agent ~$i$ uses data likelihood $\qz{i}(z_{i,t}|\hypspace)$,
and neighbor pdfs $p_{j,t}(\hypspace)$ and weights $A_{ij}$ for 
neighbors $j \in \neighbor{i}$, to obtain the mixed pdf $v_{i,t}(\hypspace)$ as:
\[
  \calN( (\Omega^g_{i,t+1})^{-1} ( \sum_{j \in \nodes_i} A_{ij} \Omega_{j,t} \mu_{j,t} ),
  (\Omega^g_{i,t+1})^{-1} ),
\]
where $\Omega^g_{i,t+1}\!=\!\sum_{j \in \nodes_i} A_{ij} \Omega_{j,t}$.
This is followed by the Gaussian likelihood update in Lemma~\ref{lemma:gaussian_likelihood_update}
using the mixed pdf $v_{i,t}(\hypspace)$ and the Gaussian likelihood $\calN(z_{i,t}| H_i^{(d)} \hypspace, V_i^{(d)})$.
}

\revisionAdd{
Next, we consider the marginal estimation setting, where each 
agent $i$ estimates a pdf over the set of relevant variables $\hypspace_i$, 
given by the vectorized version of $\crl{\vect{x}_j}_{j\in \nodes_i}$.
For this setting, we express the observation model given in \eqref{eqn:rel_loc_lik}
as $\qz{i}(z_{i,t}|\hypspace_i) = \calN(H_i^{(m)} \hypspace_i, V_i^{(m)})$.
We implement the Gaussian version of the marginal estimation 
using the mixed pdf update in Algorithm~\ref{algo:marginal_gaussian} 
followed by the likelihood update defined via the update 
in Lemma~\ref{lemma:gaussian_likelihood_update}. 
}

\revisionAdd{
Next, we will describe the BP algorithm and a recent 
circular BP version \cite{VB-RJ-SD:24}, with further details in \cite{VB:21}.
The BP algorithm allows the network to estimate a density 
of the form $\prod_{i\in \nodes} p_{i,t}(\vect{x}_i)$, 
such that agent $i$ estimates the pdf $p_{i,t}(\vect{x}_i)$.
In an undirected network, each agent $i$ generates 
a message $m_{ij,t}(\vect{x}_j)$ for its neighbor $j$ at time $t$, and vice-versa. 
Then, agent $i$ merges the neighbor messages to form its own belief, 
and computes their marginal to generate the next set of messages as follows, 
\begin{align}
  m_{ij, t+1}(\vect{x}_j) & = \int_{\vect{x}_i} \qz{i}(\vect{z}_{ij}|\vect{x}_i, \vect{x}_j)
  p_{i,t}(\vect{x}_i) \prod_{k \in \nodes_i \backslash \crl{j}}m_{ki,t}(\vect{x}_i) \nonumber \\
  p_{i,t+1}(\vect{x}_i) & \propto p_{i,t}(\vect{x}_i) \prod_{k \in \neighbor{i}} m_{ki,t}(\vect{x}_i)
\end{align}
A recent version named circular BP \cite{VB-RJ-SD:24} 
  relies on scaling the message $m_{ji,t-1}(\vect{x}_j)$ 
  with a symmetric pair-specific coefficients dependent on $(j,i)$:
\begin{align}
  m_{ij,t+1}(x_j) & \propto \int_{x_i} \qz{i}(\vect{z}_{ij}|\vect{x}_i, \vect{x}_j)^{\beta_{ij}} \\
  &  \Bigl( p_{i,t}(x_i)^{\gamma_i} 
  m_{ji,t}(x_i)^{1- \frac{\alpha_{ij}}{\kappa_i}} \prod_{k \in \neighbor{i} \setminus \{j\}} m_{ki,t}(x_i) \Bigr)^{\kappa_i}.
  \nonumber
\end{align}
With $\alpha_{ij} = \beta_{ij} = \kappa_i = \gamma_i = 1$, this algorithm reduces to the standard BP. There exists a sufficiently small $\alpha_{ij} = \alpha_{ji} = \alpha \in (0,1)$ and the rest of the terms equal to one satisfying the convergence criterion in \cite[Theorem~5.2]{VB-RJ-SD:24}, and further details in \cite{VB:21}. The theoretical fixed-point analysis in this work, however, remains limited to estimating binary probabilities. The Gaussian version of the update rule is derived in the following lemma.}
\revisionAdd{
\begin{lemma}
  Given data~$\vect{z}_{ij}$ sampled by agent $i$ from the likelihood $\phi(\vect{z}_{ij}|\vect{x}_j - \vect{x}_i, 
  \Omega_{ij}^{-1})$, prior self and neighbor messages $\phi(\vect{x}_i;\mu_{ji,t}^{(m)}, (\Omega_{ji,t}^{(m)})^{-1})$
  for $j \in \neighbor{i}$, the circular BP message with $\alpha_{ij} = \alpha \in (0,1)$ and 
  $\beta_{ij} = \gamma_i = \kappa_i = 1$ to agent $j$ is,
  \begin{align*}
    \Omega^{(m)}_{ij,t+1} & = \Omega_{ij} - \Omega_{ij}(\Omega^g_{ij,t}+\Omega_{ij})^{-1}\Omega_{ij} \\
    \mu^{(m)}_{ij,t+1} & = \vect{z}_{ij} + (\Omega^{(m)}_{ij,t+1})^{-1}
                            \Omega_{ij}(\Omega^g_{ij,t}+\Omega_{ij})^{-1}\Omega^g_{ij,t} \mu^g_{ij,t}
  \end{align*}
  where the information matrix is $\Omega^g_{ij,t+1} = \Omega_{i,t}+(1-\alpha)\Omega^{(m)}_{ji,t}+
  \sum_{k \in \neighbor{i}\backslash \crl{j}}\Omega^{(m)}_{ki,t}$ 
  and the mean is $\mu^g_{ij,t+1} = (\Omega^g_{ij,t+1})^{-1} (\Omega_{i,t}\mu_{i,t}+(1-\alpha)\Omega^{(m)}_{ji,t}\mu^{(m)}_{ji,t}
                    +\sum_{k \in \neighbor{i}\backslash \crl{j}}\Omega^{(m)}_{ki,t}\mu^{(m)}_{ki,t})$.
\end{lemma}
\begin{proof}
  We start by noting that for $\alpha_{ij}=\alpha$, the product of the densities $\Bigl( p_{i,t}(x_i)
  m_{ji,t}(x_i)^{1- \alpha} \prod_{k \in \neighbor{i} \setminus \{j\}} m_{ki,t}(x_i) \Bigr)$
  is given by the Gaussian with parameters $p^g_{ij,t}(\vect{x}_i) = \phi(\mu^g_{ij,t+1}, \Omega^g_{ij,t+1})$.
  Next, we define $\bar{\vect{x}}_j= \vect{x}_j - \vect{z}_{ij}$ and start with expressing the 
  integral coefficient in terms of $\vect{x}_i$ as,
  \begin{align*}
    & \int \qz{i}(\vect{z}_{ij}|\vect{x}_i, \vect{x}_j)p^g_{ij,t}(\vect{x}_i)d\vect{x}_i \\
    & \propto \int \exp\left( - \frac{1}{2} [\vect{x}_i^{\top}(\Omega^g_{ij,t}+\Omega_{ij})\vect{x}_i 
    - 2 \vect{x}_i^{\top}(\Omega^g_{ij,t}\mu^g_{ij,t}+\Omega_{ij}\bar{\vect{x}}_j) \right. \\
    & \quad \left. + (\mu^g_{ij,t})^{\top}\Omega^g_{ij,t}\mu^g_{ij,t} + \bar{\vect{x}}_j^{\top}\Omega_{ij}\bar{\vect{x}}_j ]\right)
    d \vect{x}_i.
  \end{align*}
  Next, we recall from \cite[Fact~14.12.1]{DB:18} $\int \exp(-\frac{1}{2} \vect{x}^{\top} A \vect{x} + \vect{c}^{\top}\vect{x} + a)
  = \sqrt{2 \pi A^{-1}} \exp \left[\frac{1}{2} \vect{c}^{\top}A^{-1}\vect{c} + a\right]$ for a symmetric 
  matrix $A \in \real^{d\times d}$, $\vect{c}\in \real^{d}, a \in \real$. 
  We can compute the mean and information matrix of the marginal by setting 
  $A = \Omega^g_{ij,t}+\Omega_{ij}$, $\vect{c} = \Omega^g_{ij,t}\mu^g_{ij,t}+\Omega_{ij}\bar{\vect{x}}_j$
  and $a = (\mu^g_{ij,t})^{\top}\Omega^g_{ij,t}\mu^g_{ij,t} + \bar{\vect{x}}_j^{\top}\Omega_{ij}\bar{\vect{x}}_j$.
  The terms containing $\bar{\vect{x}}_j$ in $\vect{c}^{\top}A^{-1}\vect{c} + a$ are,
  \begin{align*}
    - \bar{\vect{x}}_j^{\top} & (\Omega_{ij}-\Omega_{ij}(\Omega^g_{ij,t}+\Omega_{ij})^{-1}\Omega_{ij})\bar{\vect{x}}_j \\
    & + 2 \bar{\vect{x}}_j^{\top}\Omega_{ij}(\Omega^g_{ij,t}+\Omega_{ij})^{-1}\Omega^g_{ij,t}\mu^g_{ij,t},
  \end{align*}
  which yields the final result.
\end{proof}
}

We compared the distributed, marginal, BP, and circular BP algorithms in estimating the agent positions in an $8$-agent network. Each agent collects data from the model with $\Omega_{ij} = \In_2$ and initializes their mean $\mu_{i,0}$ at $(0,0)$. The evolution of position means $\mu_{i,t}$ and their error with respect to the true positions $\vect{x}_i$ are shown in Fig.~\ref{fig:rel_loc_traj_err}. The BP algorithms converge slower than the proposed distributed and marginal SMD algorithms.

  Fig.~\ref{fig:rel_loc} compares the performance of various
  algorithms as the noise levels and graph connectivity vary. The
  chosen performance metric is the estimation error of each
  algorithm at time step $T=500$, after all algorithms have
  converged. Each of the six subplots represents a different graph
  with $8$ nodes, ranging from a line graph ($7$ edges, leftmost
  subplot) to a fully connected graph ($27$ edges, rightmost
  subplot). In each subplot, estimation error ($y$ axis) 
  is plotted for algorithms implemented using noisy data sampled with information matrix value ($x$ axis) , $b\mathbb{I}_2$, 
  with magnitudes $b = 1,2, 5, 10$. We present the circular BP algorithm results with $\alpha_{ij}=0.8$ for all $i,j \in \nodes$.

  From the plots, we note that the best
  performing algorithm across the board is the full state estimation
  algorithm, showing negligible error for all graphs and error
  levels. This is ascribed to the tracking and sharing of 
  individual agent probabilities defined over all unknown variables.
  Taking this as a baseline, we can observe that the proposed algorithm 
  follows closely to this,
  and provides lower error values over sparser graphs (3 left
  subplots) than other algorithms for all noise levels. The error of
  the proposed algorithm increases as the graph becomes more dense
  and the noise increases (values for $b=10$ on the 3 right
  subplots.) In this case the performance of the belief propagation
  algorithm surpasses the proposed algorithm's; however, this
  performance difference is small and comparable.
  
  Further, we see that circular BP is the least
    accurate on sparse graphs as we increase observation noise
    magnitude, owing to insufficient countering of the loop effects in
    circular BP algorithm. In denser graphs, the errors remain too close
    to compare.

\subsection*{Distributed Mapping: An Example}


\begin{figure*}[t]
  \centering
  \includegraphics[width=0.33\textwidth]{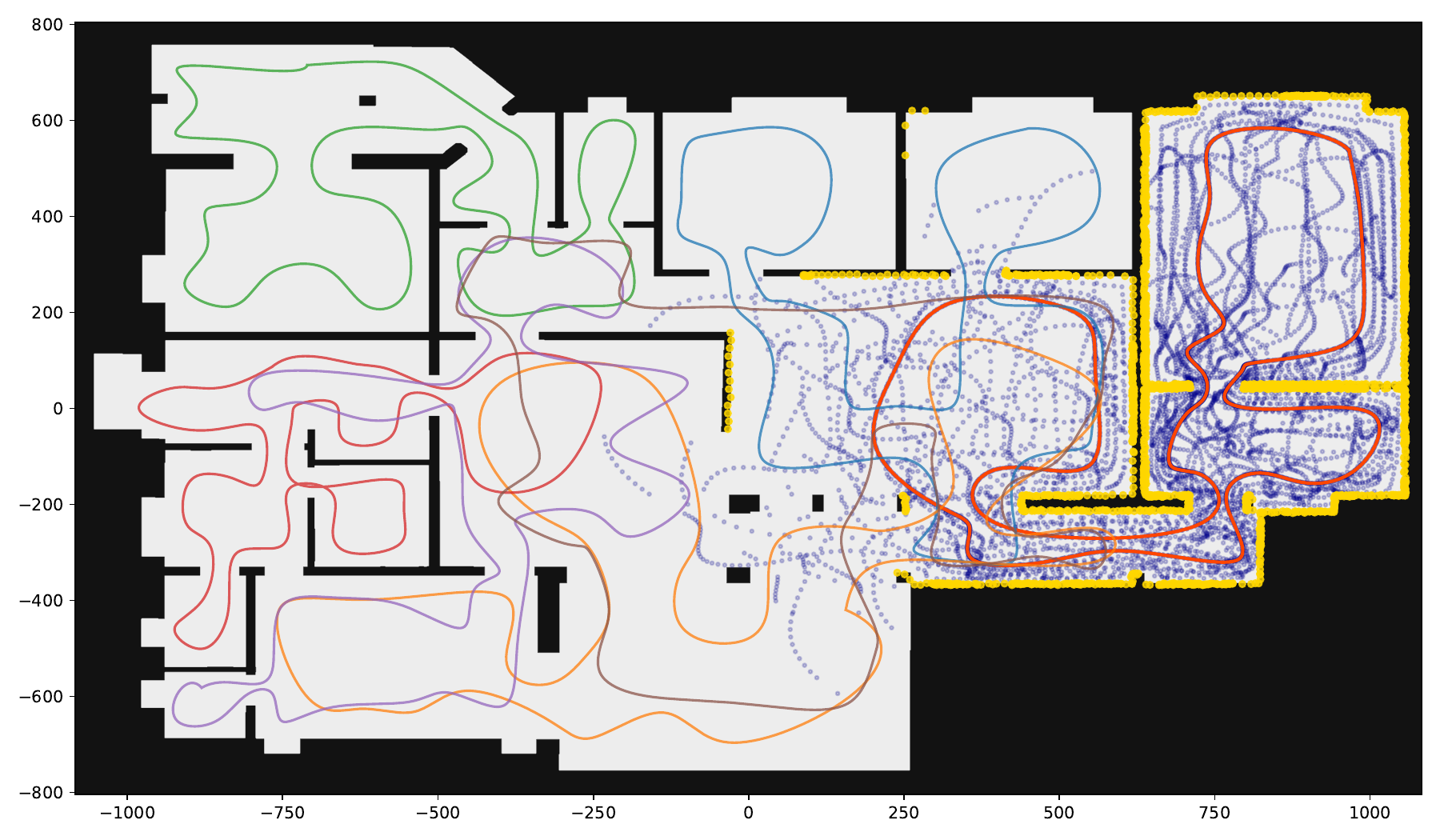}%
  \includegraphics[width=0.32\textwidth]{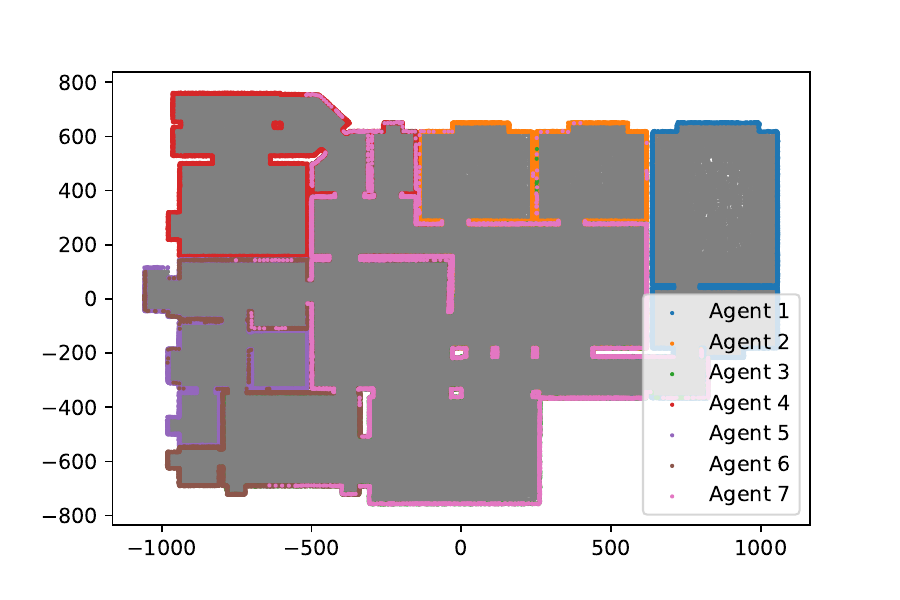}%
  \includegraphics[width=0.31\textwidth]{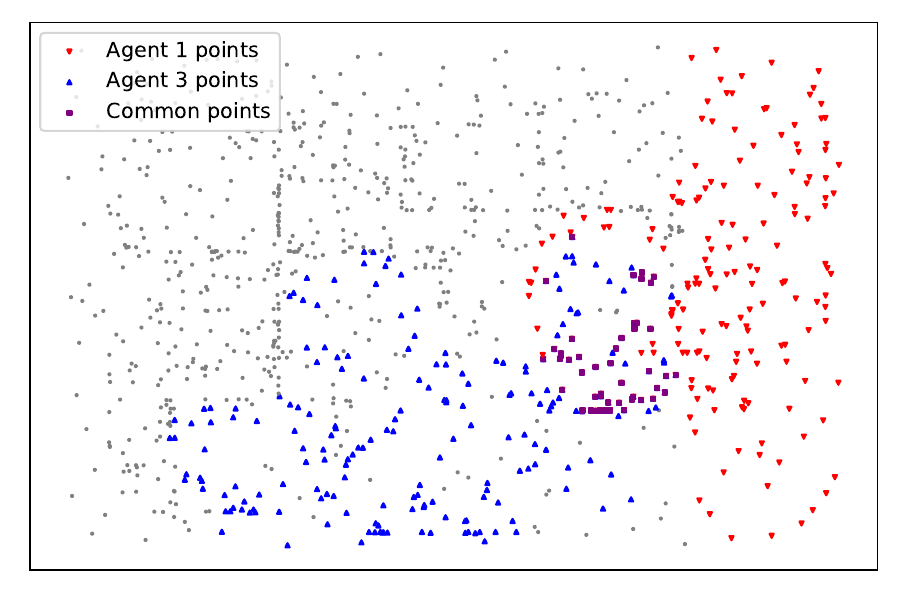}
  \caption{Agent trajectories with training samples collected by agent $1$ with free and occupied points in blue and yellow respectively (left).
  Binary data in all training sets with gray free and labeled occupied points (center).
  Distinct and shared feature points in the likelihoods of agents $1$ and $3$ (right).}
  \label{fig:inputs}
\end{figure*}

\begin{figure*}[t]
  \centering
  \includegraphics[width=0.32\textwidth]{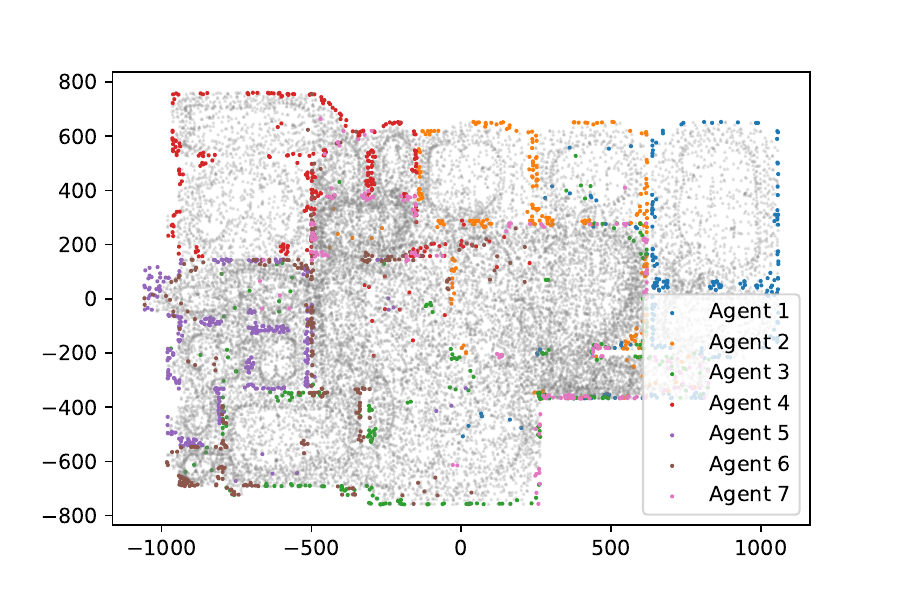}
  \includegraphics[width=0.32\textwidth]{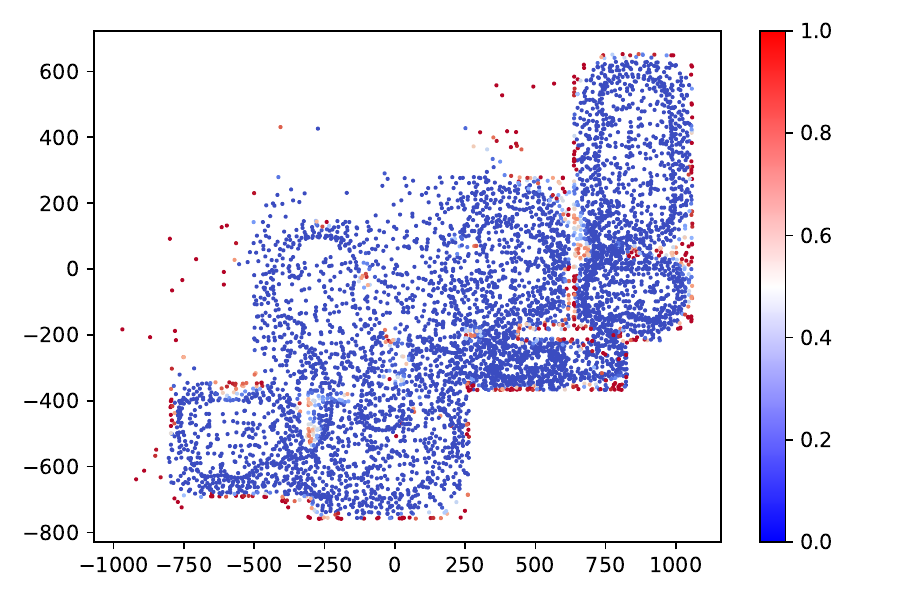}
  \includegraphics[width=0.31\textwidth]{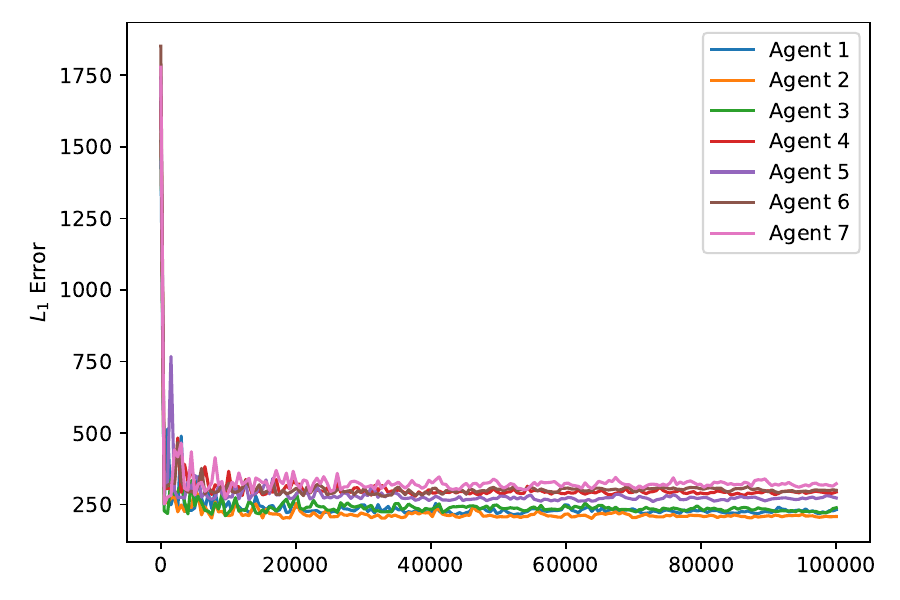}
  \caption{ 
  Predicted classes by the agents over the verification set with gray free points and labeled occupied ones (left).
  Occupancy probability for points in verification sets for agents $1$ and $3$ (center).
  $L^1$ error over the verification set during the $100$k training steps.}
  \label{fig:predictions}
\end{figure*}

In this section, we apply the marginal estimation
algorithm to distributed mapping. Please see
\cite{PP-NA-SM:20} for a simpler example solving relative localization
problem with linear observation model, where both the agent
observation models and their estimates depend on self and neighbor
states. In this multi-robot setting, each robot follows their own
trajectory allowing them to gather data describing a portion of the
map. Here, the challenge arises from the ability to achieve consensus 
over common areas by sharing partial information relevant to another robot's map.
With the knowledge of observation models describing gathered
data, the agents thus share a subset of the model parameters to
collectively create a map of the entire space.  Here, we use LiDAR
post-processed distance data to obstacles for generating
points in the free and occupied spaces.

Consider $n=7$ robots collecting data of the form $z = (x, y)$ where
$x$ is a point in the observed space and $y$ is a binary variable
indicating free or occupied status.  The point $x$ can be embedded
into the feature space using kernel functions
$k_s(x) = \gamma_1 \exp(-\gamma_2 \Vert x - x^{(s)} \Vert^2)$ centered
at $x^{(s)}$ and rescaled with parameters $\gamma_1, \gamma_2 > 0$
chosen to suit the domain and regularity of the model.  In the partial
distributed setting, this vector embedding at agent~$i$ is
$\Phi_i(x) = [1, k_{i_1}(x), \dots, k_{i_f}(x)] \in \real^{m_i+1}$.
Since some of the kernel functions are shared with neighboring agents,
the number of kernels is $m < \sum_{i} m_i$.  The modeled likelihood
of an observation $z = (x,y)$ at agent~$i$ with input $x \in \real^{\ell_i - 1}$, 
feature $\Phi_i(x)$, and label $y \in \{0,1\}$ is,
\begin{equation}\label{eqn:observation_model_sig}
  \qz{}(z | \hypspace_i) = \sigma(\Phi_i(x)^{\top} \hypspace_i)^y (1-\sigma(\Phi_i(x)^{\top} \hypspace_i))^{1-y},
\end{equation}
where $\hypspace_i$ are the agent relevant weights and $\sigma$ is the sigmoid function.
The consensus constraint enforces equality of the weights assigned to common kernel 
functions in the agent models.
To understand the role of any element $i_{\ell}$ in parameter $\hypspace_i$ for constructing a map, 
note that its positivity emphasizes the confidence in occupancy prediction 
around feature point~$x^{i_{\ell}}$ and vice-versa. 

In a marginal distributed setting, agent~$i$ models the spatial
occupancy in terms of kernels centered at relevant feature points
$\crl{x^{(s)}}_{s=i_1}^{i_f}$ out of a fixed set of $1000$ such points
across the entire map.  We construct these subsets by selecting
feature points whose distance to agent~$i$'s trajectory are under a
threshold.  For a distance threshold of $50$-units, the number of
parameters observed by the seven agents is
$(208, 195, 247, 188, 180, 224, 216)$, thus bringing the number of
variables across agents down from $7$K to $1458$ parameters.  Out of
the $216$ parameters at the last agent, the number of parameters
common with others is $(62, 66, 88, 41, 11, 42)$.  The agent training
datasets at each agent contain $80$K-$100$K points and the verification
sets consist of $3$K-$3.7$K points approximately.  If any two agent
likelihood models contain the same feature point $x^{(s)}$, then they
communicate through the network $A$ to consent over common weight
parameters.

In Figure~\ref{fig:inputs}, we present the robot trajectories for data collection,
the training set, and the distinct and shared feature points embedded in the relevant space at two of the robots. 
For generating the map, we use Lemma~\ref{lemma:dgvis} in conjunction with \cite[Lemma~4]{PP-NA-SM:23-arxiv} to simplify the 
expected gradient and Hessian terms. 
The predictions on the verification set is presented in Figure~\ref{fig:predictions}, with maps estimated 
by individual agents in center figure, with error on 
agent-specific verification sets on the right of Figure~\ref{fig:predictions}.

\section{Conclusion}
This work designs and analyzes a novel distributed estimation algorithm for estimating marginal densities over relevant variables at each agent in an inference network. The Bayes-like distributed algorithm is designed from a stochastic mirror descent perspective, with almost sure convergence guarantees.
Based on our analysis, we claim that any consensus rule with a geometric convergence rate can be coupled to stochastic mirror descent to convergence almost surely to the optimal pdf. This insight has far-reaching implications for developing distributed estimation algorithms in several metric spaces. 
The distributed mapping implementation demonstrates the vast storage savings due to the proposed algorithm. 
This algorithm can reduce storage and communication costs in networked estimation problems, based on computation-communication trade-offs.


%



\ifCLASSOPTIONcaptionsoff
  \newpage
\fi




\bibliographystyle{IEEEtran}
\bibliography{main}
%
%

\appendices

\section{Convergence analysis for SMD algorithm}
\label{app:Centralized}
\begin{proof}[Proposition~\ref{prop:variations}]
  The Gateaux derivative of $\Lambda[p]$ follows from the
  Definition~\ref{def:gateaux}.
  The derivative of $\Psi[p]$ along $\eta \in \calF_d$ is, 
\begin{equation*}
  \scaleMathLine{\begin{aligned}
      & \Psi'[p,\eta] =
      \lim_{\epsilon \rightarrow 0^+}\frac{1}{\epsilon} \left( \int (p + \epsilon \eta) \log(p + \epsilon
        \eta) d\mu - \int p \log p d\mu \right) \nonumber \\
      & =  \int \lim_{\epsilon \rightarrow 0^+}\frac{1}{\epsilon} ((p + \epsilon \eta) \log(p + \epsilon
      \eta) - p \log p)d\mu = \int \prl{1 +\log(p)}\eta d\mu,
\end{aligned}}
\end{equation*}
where we use the dominated convergence theorem~\cite{WR:06} to
exchange the limit with the integral (as $\epsilon$ can be taken to be
$ 0 \le \epsilon \le 1$, we have that $(p +\epsilon \eta) \log(p +
\epsilon \eta) \le (p + \eta) \log (p + \eta)$, which is an integrable
function). Since the KL-divergence is a linear combination of 
differential entropy and a linear functional, so is its derivative.
\end{proof}

\begin{proof}[Lemma~\ref{lemma:tvholder}]
  We start by recalling an alternative definition of total variation
  distance in~\cite[Lemma~6]{EM-MR:18},
\begin{align*}
\| p-g \|_{TV} = \frac{1}{2} \sup_{\| \Psi \|_{\infty} \leq 1} \left|\int \Psi (p-g) d\vect{x} \right|. 
\end{align*}
For a function $\Psi_0$ with $\|\Psi_0\|_\infty \neq 0$, we have,
\begin{align*}
  \langle \Psi_0, p-g\rangle  = \| \Psi_0 \|_{\infty} \int
  \frac{\Psi_0}{\| \Psi_0 \|_{\infty}} (p-g) d\vect{x}. 
\end{align*}
Upon upper bounding with the supremum,
\begin{align*}
  \langle \Psi_0, p-g\rangle & \leq \| \Psi_0 \|_{\infty} \sup_{\| \Psi \|_{\infty} \leq 1} \left|\int \Psi (p-g) d\vect{x} \right| .\\
  \implies \langle \Psi_0, p-g\rangle & \leq 2 \| \Psi_0 \|_{\infty} \| p-g \|_{TV}.
\end{align*}
The result follows trivially for $\Psi_0$ with $\|\Psi_0\|_{\infty}
= 0$.
\end{proof}

\begin{proof}[Proposition~\ref{prop:cSMD}]
  Eqn.~\eqref{eqn:cSMD} defines an equality-constrained optimization
  problem over the pdf space~$\calF_{d}$. To take the constraint
  $\int p = \langle 1, p \rangle = 1$ into account, we consider the
  Lagrangian,
\begin{equation*}
  \calL(p, \lambda) = J_t[p, p_t]
  + \lambda \prl{\langle 1, p \rangle - 1},
\end{equation*}
where $\lambda$ is a multiplier. Following Definition~\ref{def:var1},
the first variation of $\calL$ w.r.t.~$p$ is,
\begin{align*}
  \frac{\delta \calL}{\delta p} = \alpha_t  \frac{\delta F_t}{\delta
  p}[p_t] + \prl{ 1 + \log p - \log
  p_t} + \lambda .
\end{align*}
Setting the variation to zero and solving for $p$ leads to,
\begin{align*}
  p = e^{-1 - \alpha_t\lambda - \frac{\delta F_t}{\delta p}[p_t]}p_t .
\end{align*}
 
The value of $\lambda$ can be obtained from the constraint,
\begin{align*}
  1 = \int p d\mu = e^{-1 - \alpha_t\lambda} \underbrace{\int  e^{- \alpha_t \frac{\delta F_t}{\delta p}[p_t]}p_t d\mu}_{Z} ,
\end{align*}
showing that $p(\hypspace) = \frac{1}{Z} \exp\prl{- \alpha_t \frac{\delta
    F}{\delta p}[p_t ,z_t]}p_t $.
\end{proof}

\begin{proof}[Proposition~\ref{prop:optim_KL}]
  From Proposition \ref{prop:variations}, the Gateaux derivative of $J_t[\bar{p}, p_t]$ 
  evaluated at $\bar{p}$ along any direction $\eta \in \calF_{d}$ is,

  \scalebox{0.9}{\parbox{1.1\linewidth}{%
\begin{align}
  & \left\langle \frac{\delta J_t[\bar{p}, p_t]}{\delta p}, \eta
  \right\rangle
  = \left\langle \alpha_t \frac{\delta F_t[p_t]}{\delta p} + (1 + \log(\bar{p})) - \log(p_{t}), \eta \right\rangle \nonumber \\
\label{eqn:gateaux_dir}
& = \alpha_t \left\langle \frac{\delta F_t[p_t]}{\delta p}, \eta \right\rangle 
  + \left\langle \frac{\delta \Psi}{\delta p} [\bar{p}] - \frac{\delta \Psi}{\delta p} [p_{t}], \eta \right\rangle 
  +\langle 1, \eta \rangle, 
\end{align}
}}
where the entropy functional~$\Psi$ follows from Definition~\ref{prop:variations}. 
Since $p_{t+1}$ minimizes the convex functional~$J_t[\bar{p}, p_t]$, 
it follows that the first variation~$\frac{\delta J_t}{\delta p} [p_{t+1}, p_{t}] = 0$.
 Further, upon choosing~$\eta = p - p_{t+1}$ in \eqref{eqn:gateaux_dir} 
 with $\int \eta = \int p - p_{t+1} = 0$, 
 the following holds for any~$p \in \calF_{d}$,
%
\begin{align}
\label{eqn:f_ineq}
\alpha_t \left\langle \frac{\delta F_t[p_t]}{\delta p},
  p - p_{t+1}\right\rangle & + \left\langle \frac{\delta \Psi}{\delta p} [p_{t+1}], p-p_{t+1} \right\rangle \nonumber \\
- & \left\langle \frac{\delta \Psi}{\delta p}[p_{t}], p-p_{t+1}
\right\rangle = 0 .
\end{align}
These terms are simplified using the generalized
Pythagorean inequality in~\eqref{eqn:gpi} over the pdfs~$p,
p_{t}, p_{t+1}\in \calF$,
\begin{align}
  & \left\langle \frac{\delta \Psi}{\delta p} [p_{t+1}], p-p_{t+1}
  \right\rangle - \left\langle
    \frac{\delta \Psi}{\delta p}[p_{t}], p-p_{t+1} \right\rangle \nonumber \\
  & \quad = \KL [p, p_{t}] - \KL [p, p_{t+1}] - \KL [p_{t+1}, p_{t}] \nonumber \\
\label{eqn:pythagorean}
& \quad \leq \KL [p, p_{t}] - \KL[p, p_{t+1}] - 2 \|
p_{t+1}- p_{t} \|_{TV}^2.
\end{align}
Pinsker's inequality in Lemma~\ref{lemma:pinsker} leads to the inequality 
in~\eqref{eqn:pythagorean}. 
Combining this \eqref{eqn:pythagorean} with \eqref{eqn:f_ineq},
\begin{align}
\label{eqn:ipdiv}
\alpha_t & \left\langle \frac{\delta F_t[p_t]}{\delta p}, p -
  p_t +
  p_t - p_{t+1}\right\rangle \nonumber\\
& + \KL [p , p_{t}] - \KL [p , p_{t+1}] \geq 2 \| p_{t+1}- p_{t}
\|_{TV}^2.
\end{align}
We next use a total variation inequality from
Lemma~\ref{lemma:tvholder}, further upper bound using the
arithmetic-geometric mean inequality and multiply by $2$ to obtain the
following,
\scalebox{0.97}{\parbox{1.0\linewidth}{%
\begin{align}
&  \left\langle \alpha_t\frac{\delta F_t[p_t]}{\delta p}, p_t
    - p_{t+1}\right\rangle \leq 2 \left\| \alpha_t\frac{\delta F_t[p_t]}{\delta p}
    \right\|_{\infty}
  \left\|  p_{t} - p_{t+1} \right\|_{TV} \nonumber \\
\label{eqn:holder}
& \leq 2 \alpha_t^2 \left\| \frac{\delta F_t[p_t]}{\delta p}\right\|_{\infty}^2 + 2 \left\| p_{t} - p_{t+1} \right\|_{TV}^2 \\
& \leq 2 \alpha_t^2 L^2 + \KL [p , p_{t}] - \KL [p , p_{t+1}] \nonumber \\
& + \alpha_t \left\langle \frac{\delta F_t[p_t]
        }{\delta p}, p_t - p_{t+1}\right\rangle 
+ \alpha_t \left\langle \frac{\delta F_t[p_t]}{\delta p}, p - p_t\right\rangle, \nonumber
\end{align}
}} where the last inequality follows from~\eqref{eqn:ipdiv} and
bounded gradients in Assumption~\ref{assume:gradbound}.  Canceling out
the inner product term yields the desired upper bound on the
divergence,

\scalebox{0.92}{\parbox{1.0\linewidth}{%
\begin{align*}
  \alpha_t & \left\langle \frac{\delta F_t[p_t]}{\delta p},
             p - p_t\right\rangle + 2 \alpha_t^2 L^2 \geq \KL [p, p_{t+1}] - \KL [p, p_{t}] .
\end{align*}
}}

\end{proof}

\begin{proof}[Lemma~\ref{lemma:grad_expect}]
    Computing $\frac{\delta f}{\delta p}[p_t]$ via an arbitrary pdf
    $\eta \in \Lp$,
  \begin{align*}
    & \lim_{\epsilon \rightarrow 0^+} \frac{1}{\epsilon} \left( \underset{z_t \sim \qz{}^\star}{\expect} F_t[p_t+ \epsilon \eta] - \underset{z_t \sim \qz{}^\star}{\expect} F_t[p_t] \right) \\
    = & \lim_{\epsilon \rightarrow 0^+} \underset{z_t \sim \qz{}^\star}{\expect} \frac{1}{\epsilon} \left( F_t[p_t+ \epsilon \eta] - F_t[p_t] \right) \\
    = & \lim_{\epsilon \rightarrow 0^+} \underset{z_t \sim
        \qz{}^\star}{\expect} \frac{1}{\epsilon} \left( - \langle \epsilon
        \eta, \log(\qz{}(z_t|\hypspace)\rangle \right).
  \end{align*}
  The upper bound on the likelihood term due to
  Assumption~\ref{assume:gradbound} bounds the functional argument in
  the expectation:
  \begin{align*}
  \langle \eta, \log(\qz{}(z_t|\hypspace)\rangle & \leq \| \eta \|_1 \| \log(\qz{}(z_t|\hypspace)\rangle \|_{\infty} 
  \leq L.
  \end{align*} 
  Therefore, we can apply the Dominated Convergence theorem~\cite{RD:19}
  to swap the limit and expectation above 
  \begin{align*}
    \frac{\delta f}{\delta p}[p_t] &
                                     = \underset{z_t \sim \qz{}^\star}{\expect} \lim_{\epsilon \rightarrow 0^+} \frac{1}{\epsilon} \left( F_t[p_t+ \epsilon \eta] - F_t[p_t] \right)
                                     = \underset{z_t \sim \qz{}^\star}{\expect} \frac{\delta F_t}{\delta p}[p_t],
  \end{align*}
  thus proving the statement in the lemma.
\end{proof}

\begin{proof}[Theorem~\ref{thm:centralized_distance}]
  In~\eqref{eqn:c_objective0}, we introduced the set of minimizers $\calF^\star = \arg\min_{p \in \calF_d} f[p] $
  over the expected data functional as~$f[p] = \expect_{z_t \sim \qz{}^\star} \, F_t[p]$. 
  Any minimizer~$p^\star \in \calF^\star$ of the linear functional $f[p]$ satisfies the convexity property~$\langle \frac{\delta f}{\delta p}[p_t],
  p^\star-p_{t}\rangle \leq 0$ for derivative $\frac{\delta f}{\delta p}$ evaluated at an arbitrary pdf~$p_t \in \calF_d$.
  Stating Proposition~\ref{prop:optim_KL} for a minimizer~$p^\star$ and simplifying with $\frac{\delta f}{\delta p}$, 
\begin{align}
\label{eqn:breg_diff}
\KL [p^\star, p_{t+1}] & \leq \KL [p^\star, p_{t}] + 2 \alpha_t^2 L^2 \\
& \; + \alpha_t \langle \frac{\delta F_t[p_t]}{\delta p} +
\frac{\delta f[p_t]}{\delta p} - \frac{\delta f[p_t]}{\delta p}, p^\star - p_t\rangle \nonumber \\
\leq \KL [p^\star, p_{t}] & + \alpha_t \langle \frac{\delta F_t[p_t]}{\delta p} 
- \frac{\delta f[p_t]}{\delta p}, p^\star -
p_t\rangle + 2 \alpha_t^2 L^2.  \nonumber
\end{align}
From the expected gradient definition in Lemma~\ref{lemma:grad_expect}, 
\begin{align*}
  \frac{\delta f[p_t]}{\delta p} = \underset{z_t \sim
    \qz{}^\star}{\expect} \frac{\delta F_t[p_t]}{\delta p}.
\end{align*}
As a result, the expected gradient difference 
terms~$G_t[p_t] = \frac{\delta F_t[p_t]}{\delta p} - \frac{\delta f[p_t]}{\delta p}$ 
form a martingale difference sequence with a zero expectation w.r.t. 
the observation model as, 
\begin{align*}
\expect_{z_t \sim \qz{}^\star} \left[G_t[p_t] | \calZ_{t-1} \right] = 0.
\end{align*}
By definition, the gradient difference~$G_t[p_t]$ is
independent of the natural filtration of the previous
samples~$\calZ_{t-1} = \sigma(\vect{z}_{1}, \cdots, \vect{z}_{t-1})$.
The value of~$\KL [p^\star, p_{t}]$ is precisely known for any sequence
of~$z_{1:t-1}$, therefore $\expect [ \KL [p^\star, p_{t}]
|\calZ_{t-1}]= \KL [p^\star, p_{t}]$. Defining the inner product of 
the gradient difference~$g_t = \langle G_t[p_t], p^\star - p_t\rangle$, 
and computing the conditional expectation w.r.t.~$\calZ_{t-1}$, we get,
\begin{align*}
  \expect [ \KL [p^\star, p_{t+1}] |\calZ_{t-1}] 
  & \leq \KL [p^\star, p_{t}] + \alpha_t \expect [ g_t|\calZ_{t-1}] + 2 \alpha_t^2 L^2  \nonumber \\
  & = \KL [p^\star, p_{t}] + 2 \alpha_t^2 L^2.
\end{align*}

Using Gladyshev's result stated in Lemma~\ref{lemma:gladyshev}, the
property~$\sum_{t=1}^{\infty} \alpha_t^2 < \infty$ implies that
$\KL [p^\star, p_{t}]$ converges almost surely to some finite
non-negative value $d^\star$.  Since the KL-divergence functional is
continuous w.r.t. argument pdf $p$, and its range is $(0, \infty)$,
there exist pdfs $p_{\infty} \in \calF_{d}$ such that
$\KL[p^\star, p^\infty] = d^\star$.
\end{proof}

\begin{proof}[Theorem~\ref{thm:centralized_pdf}]
  For this proof, we assume that the expectation terms $\expect$
  are defined w.r.t. samples $z_t \sim \qz{}^\star$.
  We upper bound the expected gradient difference $G_t[p_t] =
  \frac{\delta F_t[p_t]}{\delta p} - \frac{\delta
    f[p_t]}{\delta p}$ using the gradient bound in Assumption~\ref{assume:gradbound} as, 
\begin{align}
  & \expect[G_t[p_t] \vert \calZ_{t-1}] = 0, \|G_t[p_t]\|_{\infty} \leq 2L, \\
  & \expect[\|G_t[p_t]\|_{\infty}^2 \vert \calZ_{t-1}] \leq 4L^2.\nonumber 
\end{align}
To prove almost sure convergence to the set of
minimizers~$\calF^{\star}$, we follow a contradiction argument, and we
assume that the pdfs in the sequence $\{p_t\}$ enter the set
neighborhood $\mathcal{B}(\calF^\star, \epsilon)$ a `finite' number of
times.  This implies the existence of an iteration~$t_0$ such
that~$\KL[p^\star, p_t] \geq \epsilon, \forall t \geq t_0$.  Given
that $\langle \frac{\delta f}{\delta p}[p], p - p^\star\rangle$ is
continuous, it attains a (possibly $-\infty$) minimum over the
set~$\overline{\calF \backslash \mathcal{B}(\calF^\star, \epsilon)}$.
Since the inner product
$\langle \frac{\delta f}{\delta p}[p], p - p^\star\rangle \le 0$ and
is equal to $0$ only if $p \in \calF^\star$, there exists
a~$c \in \real_{>0}$ s.t.
\begin{align}
\label{eqn:error}
\langle \frac{\delta f}{\delta p}[p], p^\star - p \rangle \leq -c < 0,
\forall p \in \calF \backslash \mathcal{B}(\calF^\star, \epsilon).
\end{align}
Substituting this upper bound in~\eqref{eqn:error} to the inequality~\eqref{eqn:breg_diff}
in Theorem~\ref{thm:centralized_distance}'s proof, 
the following holds for iterations $t \geq t_0$,
\begin{align*}
  \KL [p^\star, p_{t+1}] \leq & \KL [p^\star, p_{t}] + \alpha_t
  \langle
  \frac{\delta f[p]}{\delta p}, p^\star - p_t\rangle \\
  & + \alpha_t \langle G_t[p_t], p^\star - p_t\rangle + 2 \alpha_t^2 L^2  \\
  \leq \KL [p^\star, p_{t}] - & \alpha_t c + \alpha_t \langle
  G_t[p_t], p^\star - p_t\rangle + 2 \alpha_t^2 L^2.
\end{align*}
To relate the estimate to initial priors, 
we substitute the sum of step sizes~$\beta_T = \sum_{t=0}^T \alpha_t$ and
the inner product of the gradient and pdf differences as~$g_t = \langle
G_t[p_t], p^\star - p_t\rangle$ as,
\begin{align*}
  \KL & [p^\star, p_{T+1}] \leq \KL [p^\star, p_{0}] - c \beta_T +
  \sum_{t=0}^T \alpha_t g_t + 2 L^2 \sum_{t=0}^T \alpha_t^2, \\
  & = \KL [p^\star, p_{0}] - \beta_T \left[ c - \frac{\sum_{t=0}^T
      \alpha_t g_t}{\beta_T} \right] + 2 L^2 \sum_{t=0}^T \alpha_t^2.
\end{align*} 
Now, we will evaluate the limit on the upper bound.
The expected value of the gradient difference at iteration~$t$ is,
\begin{align*}
  \expect [g_t|\calZ_{t-1}] & =
  \expect [\langle G_t[p_t], p^\star - p_t \rangle|\calZ_{t-1} ]  \\
  & = \langle \expect[ G_t[p_t] |\calZ_{t-1}], p^\star - p_t
  \rangle = 0.
\end{align*}
Using H{\"o}lder's inequality on~$g_t$ and bounded gradients in
Assumption~\ref{assume:gradbound},
\begin{align*}
  \langle G_t[p_t], p^\star - p_t\rangle & \leq \left\|
    G_t[p_t] \right\|_{\infty} \left\| p^\star - p_{t}
  \right\|_{1} \leq 4 L.
\end{align*}
Thus, the expected value of~$g_t^2$ is bounded as,
\begin{align*}
  \expect [g_t^2 |z_0, \hypspace, z_t]
  & \leq 4 \expect [\| G_t[p_t]\|_{\infty}^2 |\calZ_{t-1}] \leq 16 L^2 \\
  \implies \sum_{t=0}^{\infty} & \frac{\expect[|\alpha_t g_t|^2 |
    \calZ_{t-1}]}{\beta_t^2} \leq 16 L^2 \sum_{t=0}^{\infty}
  \frac{\alpha_t^2}{\beta_t^2} < \infty.
\end{align*} Since~$\lim_{t \rightarrow \infty}\beta_T= \infty$, we
can use the strong law of large numbers for martingale difference
sequences in Lemma~\ref{lemma:stronglaw} (for $X_t =
\alpha_t g_t$ and $p = 2$) to conclude that
\begin{align*}
  & \frac{\sum_{t=0}^T \alpha_t g_t}{\beta_T}
  \rightarrow 0 \quad as \quad T \rightarrow \infty \;\; (\text{a.s.}) \\
  \implies & \beta_T \left[ c - \frac{\sum_{t=0}^T \alpha_t
      g_t}{\beta_T} \right] \rightarrow \infty \;\;(\text{a.s.})
\end{align*}
With a bounded~$\sum_{t=0}^{\infty} \alpha_t^2$, we thus have,
\begin{align*}
  \lim \sup_{T \rightarrow \infty} \KL [p^\star,p_T] = - \infty.
\end{align*}
contradicting the assumption
$\KL[p^\star, p_t] \geq \epsilon, \forall t \geq t_0$.  Therefore, the
sequence~$\{p_t\}$ enters the set $\mathcal{B}(\calF^\star, \epsilon)$
infinitely often.  From Theorem~\ref{thm:centralized_distance}, the
KL-divergence between the pdf sequence and an optimal pdf
$p^\star \in \calF^\star$ converges to a fixed value,
i.e.~$\KL[p^\star,p_t] \rightarrow d^\star$. Therefore, the
sequence~$\{p_t\}$
satisfies~$\KL[p^\star,p_t] \rightarrow d^\star \leq \epsilon$ for any
$\epsilon > 0$.
\end{proof}

\begin{proof}[Theorem~\ref{thm:centralized_rate}]
  We begin by taking the conditional expectation of the statement in
  Proposition~\ref*{prop:optim_KL},
  \begin{align*}
    & \expect[\KL [p, p_{t+1}]|\calZ_{t-1}] \nonumber \\
    & \leq \KL [p, p_{t}]  + \alpha_t \left\langle \frac{\delta f[p_t]}{\delta p},
        p - p_t\right\rangle + 2 \alpha_t^2 L^2, \\
    & = \KL [p, p_{t}] - \alpha_t (f[p_t]- f[p^\star] - 2 \alpha_t L^2),
  \end{align*}
  where the equality follows from the linearity of functional~$f$.
  Upon choosing $\alpha_t = (f[p_t]- f[p^\star])/2aL^2, a > 1$,
  \begin{align*}
    & \expect[\KL [p, p_{t+1}]|\calZ_{t-1}] 
    \leq \KL [p, p_{t}] - \frac{(f[p_t]- f[p^\star])^2}{2 a^2 L^2/(a-1)}, \\
    & \leq \KL [p, p_{0}] - \sum_{k=1}^t \frac{(f[p_t]- f[p^\star])^2}{2 a^2 L^2/(a-1)}.
  \end{align*}
  By non-negativity of the divergence term, 
  \begin{align*}
    \KL [p, p_{0}] & \geq \KL [p, p_{0}] - \expect[\KL [p, p_{t+1}]|\calZ_{t-1}] \\
    & \geq (a-1) \sum_{k=1}^t (f[p_t]- f[p^\star])^2/2 a^2 L^2.
  \end{align*}
  Using linearity of the objective function,
  \begin{align*}
    & f[\bar{p}_t] - f[p^\star] = \frac{1}{t} \sum_{k=1}^t (f[p_t] - f[p^\star]), \nonumber \\
    & \leq \sqrt{\frac{1}{t} \sum_{k=1}^t (f[p_t] - f[p^\star])^2}
    \leq \sqrt{\frac{2 a^2 L^2 \KL[p^\star, p_0]}{(a-1)t}}.
  \end{align*}
\end{proof}



\section{Analyzing DSMD algorithm}
\label{app:distributed}

\begin{proof}[Proposition~\ref{prop:d_dist_dec}]
    We start with substituting the mixed pdfs in the divergences and employ the column stochasticity of the
    communication matrix~$A$ as follows,
  \begin{align*}
    & \sum_{i=1}^n \KL [p, v_{i, t}]
      = \sum_{i=1}^n \KL [p, \prod_{j = 1}^n p_{j, t}^{A_{ij}}/Z^v_{i,t}] \\ 
    & = n \langle p, \log(p) \rangle
      - \sum_{i=1}^n \langle p,
      \sum_{j=1}^n A_{ij}\log(p_{j, t}) - \log(Z^v_{i,t})\rangle \\
    & = n \langle p, \log(p) \rangle
      -  \langle p, \sum_{j=1}^n \log(p_{j, t})\rangle
      + \langle p, \sum_{i=1}^n \log(Z^v_{i,t}) \rangle \\
    & = \sum_{j=1}^n \KL [p, p_{j, t}]
      + \log(Z^v_{j,t}).
  \end{align*}
  Here, the log-normalization factor simplifies because the pdf
  $p$ integrates to one.
  Since the geometric mean of positive numbers lower bounds their
  arithmetic mean, we have,
  \begin{align*}
    Z^v_{i,t}
    = \int \prod_{j=1}^n p_{j,t}^{A_{ij}} d\hypspace
    \leq \int \sum_{j=1}^n A_{ij} p_{j,t} d\hypspace = 1.
  \end{align*}
  and, thus, $\log(Z^v_{i,t}) \leq 0$. 
  
  Now, we will establish that the equality between mixed and original divergences 
  occurs iff all estimated pdfs are equal. 
  If, for some pdf $p\in \calF$, agent estimates $p_{i,t} = p,\forall i \in \nodes$, 
  then the previous equality holds trivially. 
  In the other direction, we have $Z^v_{i,t}=1$ implying
  $\int \prod_{j=1}^n p_{j,t}^{A_{ij}} d\hypspace - \int \sum_{j=1}^n
  A_{ij} p_{j,t} d\hypspace = 0$.  This holds iff
  $\prod_{j=1}^n p_{j,t}^{A_{ij}} = \sum_{j=1}^n A_{ij} p_{j,t}$ a.e..
  Since these are the weighted geometric and arithmetic means of non-negative terms,
  this equality holds iff the components are equal a.e.
  In a.e. sense, $p_{i,t} = p_{j,t}$ for all agents $i\in \nodes$ and their
  neighbors $j \in \nodes$.
  Further, since the graph is connected, the estimates $p_{i,t} = p_{j,t} = p$
  for all agents $i,j \in \nodes$.
\end{proof}

\begin{proof}[Proposition~\ref{prop:dist_bound}]
    The first variation of the objective
    $\frac{\partial}{\partial p} J_{i,t}[p, v_{i,t}]$ evaluated at the
    minimizer $p_{i, t+1}$ satisfies,
  
  \scalebox{0.96}{\parbox{1.0\linewidth}{%
  \begin{align}
  \label{eqn:gub_marg}
    & \alpha_t \frac{\delta F_{i,t}}{\delta p}[p_{i,t}] + \log (p_{i,t+1}) - \log(v_{i,t}) = 0. \\
    & \left\langle \alpha_t \frac{\delta F_{i,t}}{\delta p}[p_{i,t}] + \log(p_{i,t+1}) - \log(v_{i,t}), v_{i,t} - p_{i,t+1} \right\rangle = 0. \nonumber
  \end{align}
  }} Using H{\"o}lder's inequality and the gradient upper bound,
  \begin{align*}
  & \langle \log(v_{i,t}) - \log(p_{i,t+1}), v_{i,t} - p_{i,t+1} \rangle \\
  & = \left\langle \alpha_t \frac{\delta F_{i,t}}{\delta p}[p_{i,t}], v_{i,t} - p_{i,t+1} \right\rangle \leq 2 \alpha_t L \| v_{i,t} - p_{i,t+1} \|_{TV}.
  \end{align*}
  From the definition of KL-divergence and the Pinsker's inequality in Lemma~\ref{lemma:pinsker},
  \begin{align*}
    & \langle \log(v_{i,t}) - \log(p_{i,t+1}), v_{i,t} - p_{i,t+1} \rangle \\
    &   = \KL[p_{i,t+1}, v_{i,t}] + \KL[v_{i,t}, p_{i,t+1}]
  \geq 4 \| v_{i,t} - p_{i,t+1} \|_{TV}^2 .
  \end{align*}
  Therefore, $\| v_{i,t} - p_{i,t+1} \|_{TV} \leq \alpha_t L/2$ proves our claim.
  With the non-negativity of KL-divergence,
  \begin{align}
  \KL [p_{i,t+1}, v_{i,t}] & \leq \KL[p_{i,t+1}, v_{i,t}] + \KL[v_{i,t}, p_{i,t+1}] \nonumber \\
  \label{eqn:gub_res}
  & \leq 2 \alpha_t L \| v_{i,t} - p_{i,t+1} \|_{TV} \leq \alpha_t^2 L^2.
  \end{align}
  \end{proof}

\begin{proof}[Proposition~\ref{prop:dist_bound_f}]
    For the geometric average estimate~$p_t$, we have, 
    \begin{align*}
    & p_{t+1} = \prod_{i=1}^n p_{i,t+1}^{1/n}/ Z_{t+1}  \\
    & =  \exp \left(\frac{\alpha_t}{n} \sum_{i=1}^n \frac{\delta F_{i,t}}{\delta p}[p_{i,t}]\right)  \prod_{i=1}^n \left( \prod_{j=1}^n p_{j,t}^{A_{ij}/n}\right)/ Z_{t+1}  \\ 
     & = \exp \left(\frac{\alpha_t}{n} \sum_{i=1}^n\frac{\delta F_{i,t}}{\delta p}\right) p_{t} \left/ \int \exp\left(\frac{\alpha_t}{n} \sum_{i=1}^n\frac{\delta F_{i,t}}{\delta p}\right) p_{t} d\hypspace \right.  \\ 
    & = \arg\min_{p \in \calF_{m}} \left\{ \frac{\alpha_t}{n} \left\langle \sum_{i=1}^n \frac{\delta F_{i,t}}{\delta p}[p_{i,t}], p \right\rangle + \KL [p, p_{t}] \right\} .
    \end{align*}
    Here, the third step follows from the column stochasticity of matrix $A$
    and the last step relies on Proposition~\ref{prop:cSMD}.
    
    Please note that this optimization has the same structure as the functional $J_{i,t}[p, v_{i,t}]$ in Proposition~\ref{prop:dist_bound}.
    Therefore, we can recreate the proof steps in~\eqref{eqn:gub_marg} to \eqref{eqn:gub_res} to obtain 
    $\alpha_t L\| p_{t} - p_{t+1} \|_{TV} \leq \alpha_t^2 L^2/2$.
    \end{proof}

\begin{proof}[Proposition~\ref{prop:summable_c}]
    Since $p^\star$ minimizes centralized objective $f$, $f[p^\star] - f[p_t] \leq 0$ for any $p_t$. 
    Using linearity of the objective function and mean estimate $p_t = \prod_{i=1}^n p_{i,t}^{1/n}$, 
    \begin{align}
    & f[p^\star] - \sum_{i=1}^n f_i[v_{i,t}] = f[p^\star] - f[p_t] + f[p_t]- \sum_{i=1}^n f_i[v_{i,t}] \nonumber \\
    & \leq \sum_{i=1}^n (f_i[p_t - v_{i,t}])
    = \sum_{i=1}^n \underset{z_{i,t}}{\expect} \langle \frac{\delta F_{i,t}}{\delta p}[p_{i,t}], p_t - v_{i,t} \rangle \nonumber 
    \end{align}
    We substitute the expected objective gradient and employ
    Lemma~\ref{lemma:tvholder} in conjunction with the bound in
    Assumption~\ref{assume:gradbound},
    \begin{align}
    & f[p^\star] - \sum_{i=1}^n f_i[v_{i,t}] \leq \sum_{i=1}^n \langle \frac{\delta f_i}{\delta p} [p_{i,t}], p_t - v_{i,t} \rangle , \nonumber \\
    &  \leq \sum_{i=1}^n 2 L \| p_t - v_{i,t} \|_{TV} \leq 2 \sigma L \sum_{i=1}^n \| p_t - p_{i,t} \|_{TV}, \nonumber
    \end{align}
    where the last inequality follows from Proposition~\ref{prop:attract}.
  \end{proof}


  \begin{proof}[Proposition~\ref{prop:summable_cTV}]
    This proof uses the triangle inequality property of the TV distance
    in $\| p_{t+1} - p_{i,t+1}\|_{TV}$ to show its convergence as
    a sequence.  To begin, we upper bound the norm by introducing mixed
    pdfs~$v_{i,t}$ and their average~$p_t$ as follows,
  \begin{align}
    \label{eqn:sum_cTV_bound}
    \| p_{t+1} & - p_{i,t+1}\|_{TV} \leq \| p_t -
                    v_{i,t}\|_{TV} \\
                  &  + \| p_{t+1} - p_t\|_{TV} +
                    \| v_{i,t} - p_{i,t+1}\|_{TV}. \nonumber
  \end{align}
  Based on the contraction property of TV in
  Proposition~\ref{prop:attract} due to the mixing step, we have,
  \begin{align}
  \label{eqn:attract}
    \| v_{i,t} - p_t\|_{TV} \leq \sigma(A) \| p_{i,t}
    - p_t\|_{TV}.
  \end{align}
  Additionally, Propositions~\ref{prop:dist_bound}
  and~\ref{prop:dist_bound_f} upper bound the second and third terms by
  the gradient bound $L$ as,
  \begin{align*}
  \| p_{t+1} - p_t\|_{TV} \leq \alpha_t L/2, 
   \| v_{i,t} - p_{i,t+1}\|_{TV}  \leq \alpha_t L/2.
  \end{align*}
  In conjunction with~\eqref{eqn:attract}, we upper bound
  \eqref{eqn:sum_cTV_bound} as,
  \begin{align}
  \label{eqn:TV_update}
    \| p_{t+1} - p_{i,t+1} \|_{TV} \leq \sigma(A)
    \| p_t - p_{i,t}\|_{TV} +  \alpha_t L.
  \end{align}
  Multiplying~\eqref{eqn:TV_update} with~$\alpha_{t+1} L$ 
  and bounding~$\alpha_{t+1}<\alpha_{t}$,
  \begin{align*}
  \alpha_{t+1} & L \| p_{i,t+1} - p_{t+1}\|_{TV} \leq \alpha_{t} L \| p_{i,t+1} - p_{t+1}\|_{TV} \\
  & \leq \sigma(A) \alpha_t L \| p_t - p_{i,t}\|_{TV} +  \alpha_t^2 L^2 .\\
  \implies & \alpha_{t+1} L \sum_{i=1}^n \| p_{i,t+1} - p_{t+1}\|_{TV} \\
  & \leq \sigma(A) \alpha_t L \sum_{i=1}^n \| p_t - p_{i,t}\|_{TV} + n \alpha_t^2 L^2.
  \end{align*}
  Defining $d_t = 2 \alpha_t L \sum_{i=1}^n \| p_t - p_{i,t}\|_{TV}$ leads to,
  \begin{align*}
  d_{t+1} & \leq \sigma(A) d_t + 2 n \alpha_t^2 L^2 \\
  & \leq \sigma(A)^t d_0 + 2 n L^2  \left( \sum_{k=0}^t \sigma(A)^{t-k} \alpha_k^2 \right). \\
  \sum_{t=0}^T d_t & \leq \sum_{t=0}^T \sigma(A)^t d_0 + 2 n L^2 \left( \sum_{t=0}^T \sum_{k=0}^t \sigma(A)^{t-k} \alpha_k^2 \right).
  \end{align*}
  From convergence of geometric series and~$\sum_{t=0}^{\infty} \alpha_t^2 = M$,
  \begin{align*}
  \sum_{t=0}^{\infty} \sigma(A)^t d_0 & = \frac{\sigma(A)}{1- \sigma(A)} d_0. \\
  \sum_{t=0}^{\infty} \sum_{k=0}^t \sigma(A)^{t-k} \alpha_k^2 & = \sum_{t=0}^{\infty} \sum_{l=0}^t \sigma(A)^{l} \alpha_{t-l}^2 \nonumber \\
  & = \sum_{l=0}^{\infty} \sigma(A)^l \sum_{t=l}^{\infty} \alpha_{t-l}^2 \leq \frac{M \sigma(A)}{1 - \sigma(A)}. \\
  \implies \sum_{t=0}^{\infty} d_t & \leq \frac{d_0 \sigma(A)}{1 - \sigma(A)} + \frac{2 n L^2 M \sigma(A)}{1 - \sigma(A)}.
  \end{align*}
  Thus, the
  sequence~$\alpha_t L \sum_{i=1}^n \| p_t - p_{i,t} \|_{TV}$ is
  summable.
  \end{proof}
  
\begin{proof}[Theorem~\ref{thm:distributed_distance}]
    From the upper bound in Proposition~\ref{prop:d_dist_dec},  we have,
    \begin{align*}
    \sum_{i=1}^n \KL [p^\star, v_{i, t+1}] & \leq \sum_{i=1}^n \KL [p^\star, p_{i, t+1}], \qquad \forall \, t \ge 0.
    \end{align*}
    Let us now recall that the pdf~$p_{i,t+1}$ minimizes the objective functional~$J_{i,t}[p_i, v_{i,t}]$ at iteration $t$. 
    This objective functional is structured the same as centralized setting in \eqref{eqn:cSMD} with prior density $v_{i,t}$ instead of $p_t$, 
    and optimal argument $p_{i,t+1}$ instead of $p_{t+1}$.
    Thus, applying Proposition~\ref{prop:optim_KL} to any agent~$i$'s objective~$J_{i,t}[p_i, v_{i,t}]$ with $\eta = p^\star - p_{i,t+1}$, we have, 
    \begin{align}
    \label{eqn:d1_f_ineq}
    & \KL [p^\star, p_{i, t+1}] - \KL [p^\star, v_{i, t}] \\
    & \leq \alpha_t \langle \frac{\delta F_{i,t} [p_{i,t}]}{\delta p}, p^\star - v_{i,t} \rangle + 2 \alpha_t^2 L^2. \nonumber
    \end{align}
    Since the expected stochastic gradient~$\frac{\delta f_i}{\delta p} [p_i]$ is a  linear functional, 
    it satisfies the property~$\langle \frac{\delta f_i}{\delta p}[p_{i,t}], p^\star-v_{i,t}\rangle \leq f_i[p^\star] - f_i[v_{i,t}]$. 
    We use this property by introducing the expected gradient and the stochastic gradient sample $\frac{\delta F_{i,t} [p_{i,t}]}{\delta p}$,
    \begin{align}
    \label{eqn:d_breg_diff}
    & \KL [p^\star, v_{i, t+1}] \leq  \KL [p^\star, p_{i, t+1}] \leq \KL [p^\star, v_{i, t}] + 2 \alpha_t^2 L^2 \nonumber\\
    & + \alpha_t \langle \frac{\delta F_{i,t} [p_{i,t}]}{\delta p} + \frac{\delta f_i[p_{i,t}]}{\delta p} - \frac{\delta f_i[p_{i,t}]}{\delta p}, p^\star - v_{i,t}\rangle  \\
    & \leq \KL [p^\star, v_{i,t}] + \alpha_t(f_i[p^\star] - f_i[v_{i,t}]) \\
    & + 2 \alpha_t^2 L^2 \nonumber  + \alpha_t \langle \frac{\delta F_{i,t} [p_{i,t}]}{\delta p} - \frac{\delta f_i[p_{i,t}]}{\delta p}, p^\star - v_{i,t}\rangle .\nonumber 
    \end{align}
    
    the divergence sum at iteration $t+1$ satisfies,
    \begin{align}
      \sum_{i=1}^n \KL [p^\star,  v_{i, t+1}]  & \leq \sum_{i=1}^n \KL [p^\star, v_{i, t}] + \alpha_t \sum_{i=1}^n (f_i[p^\star] -  f_i[v_{i,t}]) \nonumber \\
      \label{eqn:as_dsmd}
       + 2 n \alpha_t^2 L^2 + \sum_{i=1}^n & \alpha_t \langle \frac{\delta F_{i,t} [p_{i,t}]}{\delta p} - \frac{\delta f_i[p_{i,t}]}{\delta p}, p^\star - v_{i,t}\rangle . 
    \end{align}
    The terms~$G_{i,t}[p_{i,t}] = \frac{\delta F_{i,t} [p_{i,t}]}{\delta p} - \frac{\delta f_i[p_{i,t}]}{\delta p}$ form a martingale difference sequence as,
    \begin{align*}
    \expect_{z_{i,t}} \left[G_{i,t}[p_{i,t}]| \calZ_{t-1} \right] = 0,
    \end{align*}
    implying that the expected gradient difference~$G_{i,t}[p_{i,t}]$ 
    is independent of the natural filtration of previous samples~$\calZ_{t-1} = \sigma(\vect{z}_{1:n,0}, \cdots, \vect{z}_{1:n,t-1})$. Thus, \eqref{eqn:as_dsmd} becomes,
    \begin{align*}
      & \expect \left[\sum_{i=1}^n \KL [p^\star,  v_{i, t+1}] \right] \\
      & \leq \sum_{i=1}^n \KL [p^\star, v_{i, t}] + 2n  \alpha_t^2 L^2 
          + \alpha_t \sum_{i=1}^n (f_i[p^\star] - f_i[v_{i,t}]) \\
      & \leq \sum_{i=1}^n \KL [p^\star, v_{i, t}] + 2n  \alpha_t^2 L^2 + \sigma \alpha_t L \sum_{i=1}^n \| p_t - p_{i,t}\|_{TV},
    \end{align*}
    where the second inequality follows from Proposition~\ref{prop:summable_c}.
    The Robbins-Monro sequence $\alpha_t$ is square-summable and the sequence $\sigma \alpha_t L \sum_{i=1}^n \| p_t - p_{i,t}\|_{TV}$
    is summable due to Proposition~\ref{prop:summable_cTV}.
    Thus, we claim that $\sum_{i=1}^n \KL [p^\star, v_{i, t+1}]$ almost surely converges to a finite non-negative value using Gladyshev's result in Lemma~\ref{lemma:gladyshev}.
    \end{proof}
    

\begin{proof}[Theorem~\ref{thm:distributed_pdf}]
    Following the proof method for Theorem~\ref{thm:centralized_pdf}, we assume that for at least one agent~$l \in \nodes$, the sequence~$\{v_{l,t}\}$ enters~$\mathcal{B}(\calF^\star, \epsilon)$ a finite number of times to get to a contradiction. This implies the existence of~$t_0$ such that~$\KL[p^\star, v_{l,t}] \geq \epsilon , \forall t \geq t_0$. 
    We also have~$\langle \frac{\delta f}{\delta p}[p_{l,t}], v_{l,t} - p^\star\rangle = 0$ only if~$v_{l,t} \in \calF^\star$.  Therefore, there exists a~$c \in \real_{>0}$,
    \begin{align}
      \label{eqn:mc}
    \langle \frac{\delta f}{\delta p}[p_{l,t}], p^\star - v_{l,t} \rangle \leq -c < 0, \forall v_{l,t} \in \calF \backslash \mathcal{B}(\calF^\star, \epsilon).
    \end{align}
    As before, we note that the agent objective $J_{i,t}[p_i, v_{i,t}]$ mirrors the central objective in \eqref{eqn:cSMD} with 
    prior $v_{i,t}$ and optimal pdf $p_{i,t+1}$ instead of pdfs $p_t, p_{t+1}$ respectively.
    Thus, we can apply Proposition~\ref{prop:optim_KL} to any agent~$i$'s objective functional~$J_{i,t}[p_i, z_{i,t}]$ with $\eta = p^\star - p_{i,t+1}$ and include $\frac{\delta f_i[p_{i,t}]}{\delta p}$ for simplification as, 
    \begin{align*}
      & \KL [p^\star, v_{i, t+1}] \leq \KL [p^\star, v_{i, t}] + 2 \alpha_t^2 L^2 \\
      & + \alpha_t \langle \frac{\delta F_{i,t} [p_{i,t}]}{\delta p} + \frac{\delta f_i[p_{i,t}]}{\delta p} - \frac{\delta f_i[p_{i,t}]}{\delta p}, p^\star - v_{i,t}. \rangle \nonumber
    \end{align*}
    Summing the terms for all agents and including the gradient difference term~$G_{i,t}[p_{i,t}] = \frac{\delta F_{i,t} [p_{i,t}]}{\delta p} - \frac{\delta f_i[p_i]}{\delta p}$,
    \begin{align}
      \label{eqn:en_dsmd}
    \sum_{i=1}^n \KL [p^\star, v_{i,t+1}] & \leq \sum_{i=1}^n \left( \KL [p^\star, v_{i,t}] + \alpha_t \langle \frac{\delta f_i[p_{i,t}]}{\delta p}, p^\star - v_{i,t}\rangle \right. \nonumber \\
    & \left. + \alpha_t \langle G_{i,t}[p_{i,t}], p^\star - v_{i,t}\rangle + 2 \alpha_t^2 L^2 \right). 
    \end{align}
    To  simplify the previous expression, we first examine $f[v_{l,t}] = \sum_{i=1}^n f_i[v_{l,t}]$ for an arbitrary $l \in \nodes$ as follows in order to admit the sharper bound in~\eqref{eqn:mc}, 
    \begin{align*}
    & \alpha_t \sum_{i=1}^n \langle \frac{\delta f_i[p_{i,t}]}{\delta p}, p^\star - v_{i,t}\rangle = \alpha_t(f[p^\star] - \sum_{i=1}^n f_i[v_{i,t}]) \\
    & = \alpha_t \left(f[p^\star] - \sum_{i=1}^n f_i[v_{l,t}] + f[p_t] - \sum_{i=1}^n f_i[v_{i,t}]\right. \nonumber \\
    & \left. \qquad + \sum_{i=1}^n f_i[v_{l,t}] -  f[p_t] \right) \nonumber \\
    & \leq - \alpha_t c + 2 \sigma \alpha_t L (\sum_{i=1}^n \| p_t - p_{i,t}\|_{TV}+\| p_t - p_{l,t}\|_{TV}),
    \end{align*}
    where the TV upper bounds follow from the application of Proposition~\ref{prop:summable_c} to single and multi-agent settings.
    Now, define a sequence with terms~$d_t = 2 \sigma \alpha_t L \left( \sum_{i=1}^n(\| p_t - p_{i,t}\|_{TV}) + n \| p_t - p_{l,t}\|_{TV} \right)$. 
    Following Proposition~\ref{prop:summable_cTV}, the sequence~$d_t$ is summable. 
    Adding the prior upper bound with definition $d_t$ to \eqref{eqn:en_dsmd}, we have,
    \begin{align*}
    \sum_{i=1}^n \KL [p^\star, v_{i,t+1}] & \leq \sum_{i=1}^n \KL [p^\star, v_{i,t}] 
                                          - \alpha_t c + \left(2 n \alpha_t^2 L^2 + d_t \right) \nonumber \\
    & \, + \alpha_t \sum_{i=1}^n \langle G_{i,t}[p_{i,t}], p^\star - v_{i,t}\rangle .
    \end{align*}
    To represent the sum of divergence terms in terms of priors at time $t=0$,
    we introduce a temporal sum~$\beta_T = \sum_{t=0}^T \alpha_t$, and 
    the agent inner product $g_{i,t} = \langle G_{i,t}[p_{i,t}], p^\star - v_{i,t}\rangle$ as,
    \begin{align*}
    \sum_{i=1}^n \KL [p^\star, v_{i,T+1}] \leq & \sum_{i=1}^n \KL [p^\star, v_{i,0}] - c \beta_T 
    + \sum_{t=0}^T \alpha_t \sum_{i=1}^n g_{i,t} \\
    & + \sum_{t=0}^T \left(2 n \alpha_t^2 L^2 + d_t \right) .
    \end{align*}
    \begin{align}
      \label{eqn:dsmd_psubseq}
      \implies & \sum_{i=1}^n \KL [p^\star, v_{i,T+1}] \leq \sum_{i=1}^n \KL [p^\star, v_{i, 0}] \\
      & - \beta_T \left[ c  - \frac{\sum_{t=0}^T \alpha_t \sum_{i=1}^n g_{i,t}}{\beta_T} \right] 
      + \sum_{t=0}^T \left(2 n \alpha_t^2 L^2 + d_t \right). \nonumber
    \end{align}
    Using the gradient bound in Assumption~\ref{assume:gradbound} for any observation~$z_{i,t}$, we compute similar upper bounds on the expected gradient difference $G_{i,t}[p_{i,t}] = \frac{\delta F_{i,t} [p_{i,t}]}{\delta p} - \frac{\delta f[p_{i,t}]}{\delta p}$ as,
    \begin{align*}
    \expect[G_{i,t}[p_{i,t}] \vert \calZ_{t-1}] = 0, \quad
    \|G_{i,t}[p_{i,t}]\|_\infty \leq 2L, \\
    \expect[\|G_{i,t}[p_{i,t}]\|_{\infty}^2 \vert \calZ_{t-1}] \leq 4L^2.
    \end{align*}
    Using this, we can show that the terms~$g_{i,t}$ also form a martingale difference sequence,
    \begin{align*}
    \expect [g_{i,t}|\calZ_{t-1}] & = \expect [\langle G_{i,t}[p_{i,t}], p^\star - v_{i,t} \rangle|\calZ_{t-1} ]  \\
    & =  \langle \expect[ G_{i,t}[p_{i,t}] |\calZ_{t-1}], p^\star - v_{i,t} \rangle = 0.
    \end{align*}
    
    Using H{\"o}lder's inequality, the following holds for every agent~$i \in \nodes$ and~$p, v_{i, t} \in \calF$,
    \begin{align*}
    \langle G_{i,t}[p_{i,t}], p - p_{i,t}\rangle & \leq \left\| G_{i,t}[p_{i,t}] \right\|_{\infty} \left\|  p - v_{i,t} \right\|_{1} \leq 4L. \nonumber \\
    \expect [g_{i,t}^2 |\vect{z}_0, \cdots, \vect{z}_t] & \leq 4 \expect [G_{i,t}[p_{i,t}]^2 |\calZ_{t-1}] \leq 16 L^2. \nonumber\\
    \expect [g_{i,t}g_{j,t} |\vect{z}_0, \cdots, \vect{z}_t] & \leq 4 \expect [G_{i,t}[p_{i,t}]G_{j,t}[p_{j,t}] |\calZ_{t-1}] \leq 16 L^2.
    \end{align*}
    Thus, the expected value of~$g_t^2$ is bounded as,
    \begin{align*}
    \sum_{t=0}^{\infty} \frac{\expect[|\alpha_t \sum_{i=1}^n g_{i,t}|^2 | \calZ_{t-1}]}{\beta_t^2} \leq 16 n^2 L^2 \sum_{t=0}^{\infty} \frac{\alpha_t^2}{\beta_t^2}.
    \end{align*}
    Since~$\lim_{t \rightarrow \infty}\beta_T= \infty$, we
    can use the strong law of large numbers for martingale difference
    sequences in Lemma~\ref{lemma:stronglaw} (for $X_t =
      \sum_{i=1}^n g_{i,t}$ and $p = 2$) to conclude that,
    \begin{align*}
    & \frac{\sum_{t=0}^T \alpha_t (\sum_{i=1}^n g_{i,t})}{\beta_T} \rightarrow 0 \quad as \quad T \rightarrow \infty \quad (\text{a.s.}) \\
    \implies & \beta_T \left[ c  - \frac{\sum_{t=0}^T \alpha_t \sum_{i=1}^n g_{i,t}}{\beta_T} \right] \rightarrow \infty\quad (\text{a.s.})
    \end{align*}
    With the summable~$\alpha_t^2$ and $d_t$ in \eqref{eqn:dsmd_psubseq}, we thus have, 
    \begin{align*}
    \lim \sup_{T \rightarrow \infty} \sum_{i=1}^n \KL [p^\star, v_{i,T}] = - \infty.
    \end{align*}
    Along with the non-negativity of the other divergence terms for agents $i \neq l$, 
    this contradicts our assumption that the remaining term $\KL[p^\star, v_{l,t}] \geq \epsilon, \forall t \geq t_0$. 
    Therefore, every agent's sequence~$\{v_{i,t}\}$ enters the set $\mathcal{B}(\calF^\star, \epsilon)$ infinitely many times for all agents~$i \in \nodes$.
    By Theorem~\ref{thm:distributed_distance}, the KL-divergence between the pdfs in sequence~$v_{i,t}, \forall i \in \nodes$ to an optimal pdf $p^\star$ converges to a constant value $d^\star$.
    
    Now, since estimates $v_{l,t}$ do not satisfy $\KL[p^\star,v_{l,t}] \ge \epsilon$ for an infinite number of $t$, for all $\epsilon >0$, it follows that $d^\star < n \epsilon $. Thus, $d^\star \equiv 0$ and the conclusion on convergence follows.
    \end{proof}



\section{Convergence analysis in centralized setting}
\label{app:marginal}

\begin{proof}[Proposition~\ref{prop:marg_equality}]
    $(\leftarrow)$ Suppose that $\sum_{i=1}^n \KL [\bar{p}_i, p_{i, t}] = 0$ for some $\bar{p}$. Since the KL divergences are non-negative, each term satisfies $\KL [\bar{p}_i, p_{i, t}] = 0$.
    Therefore, for any $i, j \in \nodes$, the pdf $p_{i, t} = \bar{p}_i$
    and $p_{j, t} = \bar{p}_j$ almost everywhere. Additionally, their marginals on the common space are equal, $p_{ij, t} = p_{ji, t} = \bar{p}_{ij}$.
    For equal marginals, $\tilde{p}_{ji,t} = p_{i,t}$ for any neighbors $(i,j) \in \edges$ with normalization factor,
    \[
        Z^v_{i,t} = \int \prod_{j=1}^n \left( \tilde{p}_{ji,t} \right)^{A_{ij}} = \int \prod_{j=1}^n \left( p_{i,t} \right)^{A_{ij}} = \int p_{i,t} = 1.
    \]

    $(\rightarrow)$ Now, suppose that the product of normalization factors is $\prod_{i=1}^n Z^v_{i,t} = 1$. We begin by noting that the product of conditional-marginal density is normalized, i.e. 
    \begin{align*}
      &\int \frac{p_{i,t}}{p_{ij,t}} p_{ji,t} d \hypspace_i 
      = \int_{\hypspace_{ij}} \int_{\hypspace_i \backslash \hypspace_{ij}} \frac{p_{i,t}}{p_{ij,t}} p_{ji,t}  \\
      & = \int_{\hypspace_{ij}} \frac{p_{ji,t}}{p_{ij,t}} \left(\int_{\hypspace_i \backslash \hypspace_{ij}} p_{i,t}\right) 
      = \int_{\hypspace_{ij}} \frac{p_{ji,t}}{p_{ij,t}} p_{ij,t}  = 1.  
    \end{align*}
    Arithmetic means upper bound geometric means for non-negative numbers, $\prod_{j=1}^n \left( \tilde{p}_{ji,t} \right)^{A_{ij}} \leq \sum_{j=1}^n A_{ij} \tilde{p}_{ji,t} $, implying that $Z^v_{i,t} \leq 1$. To satisfy the product equality $\prod_{i=1}^n Z^v_{i,t} = 1$, each normalization factor 
    $Z^v_{i,t} = 1$.
    Since the two means are non-negative functions satisfying 
    $\int \left(\sum_{j=1}^n A_{ij} \tilde{p}_{ji,t} - \prod_{j=1}^n \left( \tilde{p}_{ji,t} \right)^{A_{ij}}\right) = 0$, we have, 
    \begin{align*}
        \sum_{j=1}^n A_{ij} \tilde{p}_{ji,t} = \prod_{j=1}^n \left( \tilde{p}_{ji,t} \right)^{A_{ij}} \text{ a.e.}.
    \end{align*}
    The elements of the weighted adjacency matrix $A$ satisfy $A_{ij} \in (0,1)$ for all $(i, j) \in \edges$ because the graph is connected.
    Thus, with $A_{ij} < 1$, the means are equal if and only if the arguments are equal to each other. 
    In a.e. sense, for all agents $i\in \nodes$ and their neighbors $j \in \nodes$,
    \begin{align*}
        & p_{i,t} = \tilde{p}_{ji,t} = p_{i,t}\frac{p_{ji,t}}{p_{ij,t}}, \\
        & p_{ij,t} = p_{ji,t} , \quad \forall i,j \in \edges.
    \end{align*}
    Per Assumption~\ref{assume:marginal_network}, the set of agents observing any particular variable are connected. 
    By extension, for any two agents~$i, j$ in the network, estimated pdfs are equal on the common variables~$\hypspace_{ij}$ 
    leading to $\sum_{i=1}^n \KL [\bar{p}_i, p_{i, t}] = 0$ for some $\bar{p}$.
\end{proof}

\begin{proof}[Proposition~\ref{prop:m_dist_dec}]
    We define marginal pdfs $p_{i} = \int_{\hypspace \backslash \hypspace_{i}} p$, 
    $p_{ij} = p_{ji} = \int_{\hypspace \backslash \hypspace_{ij}} p$.
    The marginal pdfs $p_{ji,t}, p_{ij,t}$ are similarly derived from $p_{j,t}, p_{i,t}$ respectively. 
    Then, the KL-divergence between the relevant marginals 
    of pdf $p$ and any mixed pdf $v_{i,t}$ is,
    \begin{align*}
    \KL \left( p_{i}, v_{i,t} \right)  = \KL \left[ p_{i}, \prod_{j=1}^n \left( \tilde{p}_{ij,t} \right)^{A_{ij}} \right] + \log(Z^v_{i,t}) ,
    \end{align*}
    with normalization factor $Z^v_{i,t} = \int \prod_{j=1}^n \left( \tilde{p}_{ij,t} \right)^{A_{ij}} d\hypspace_i$ and 
    \begin{align*}
    & \KL \left[ p_{i}, \prod_{j=1}^n \left( \tilde{p}_{ij,t} \right)^{A_{ij}} \right] 
    = \KL \left[ p_{i}, \prod_{j=1}^n \left( \frac{p_{i,t}}{p_{ij,t}} p_{ji,t} \right)^{A_{ij}} \right] \\
    & = \langle p_{i} , \log(p_{i}) \rangle - \Bigg\langle p_{i} , \log\prod_{j=1}^n \left( \frac{p_{i,t}}{p_{ij,t}} p_{ji,t} \right)^{A_{ij}}  \Bigg\rangle  \\
    & = \KL [p_i, p_{i, t}] - \sum_{j=1}^n R(i,j),
    \end{align*}
    where $R(i,j) =  A_{ij} \left( \left\langle p_{i}, \log(p_{ji,t})\right\rangle - \left\langle p_{i},  \log(p_{ij,t}) \right\rangle \right)$.
    Due to the symmetry of the communication matrix $A_{ij} = A_{ji}$, 
    and $\left\langle p_{i}, \log(p_{ji,t})\right\rangle = \left\langle p_{ij}, \log(p_{ji,t})\right\rangle$,
    the complementary residual terms cancel out, i.e., $R(i, j) + R(j, i) = 0$.
    Further, since the graph is undirected, 
    \begin{align*}
    \sum_{i=1}^n \sum_{j=1}^n R(i, j) = 0.
    \end{align*}
    
    As described in the Proposition~\ref{prop:m_dist_dec}, 
    the product of conditional-marginal densities~$\tilde{p}_{ji,t}$ is normalized. Thus,
    \begin{align}
      \label{eqn:}
    & \sum_{i=1}^n \KL [p_i, v_{i, t}] = \sum_{i=1}^n \left( \KL [p_i, p_{i, t}] + \log(Z^v_{i,t}) \right)\\
    & = \sum_{i=1}^n \left( \KL [p_i, p_{i, t}] + \log \int \prod_{j \in \nodes_i} \tilde{p}_{ji,t}^{A_{ij}} \right) \tag{GM$\leq$AM} \\
    & \leq \sum_{i=1}^n \left( \KL [p_i, p_{i, t}] + \log \int \sum_{j \in \nodes_i} A_{ij} \tilde{p}_{ji,t} \right) \\
    & \leq \sum_{i=1}^n \KL [p_i, p_{i, t}].
    \end{align}
    The equality follows from Proposition~\ref{prop:marg_equality}.
    \end{proof}


\begin{proof}[Proposition~\ref{prop:m_mix_eqm}]
    Consider a pdf $\bar{p}_t \in \calF_d$ with marginals $\crl{\bar{p}_{i,t}}$. 
    Define a sequence with terms $V_{t}[\crl{p_{i,t}}] = \sum_{i=1}^n \KL [\bar{p}_{i,t}, p_{i,t}]$ 
    with a shorthand $V_{t,k} = V_{t}[\crl{p_{i,t}^{(k)}}]$. 
    Due to its non-negativity, the divergence sum is lower bounded by zero.
    If the pdfs $\crl{p_{i,t}^{(k)}}$ obtained from $k$-step application of 
    the marginal consensus steps on pdfs $\crl{p_{i,t}}$ do not lie in the set $E_t$, then from Proposition~\ref{prop:m_dist_dec},  
    \begin{align*}
      V_{t,k+1} - V_{t,k}
      & < \sum_{i=1}^n \log Z^{v,k}_{i,t} \leq \sum_{i=1}^n (Z^{v,k}_{i,t} - 1) < 0,
    \end{align*}
    where the second inequality follows from concavity and normalization factor $Z^{v,k}_{i,t}<1$.
    By monotone convergence theorem, the sequence $V_{t,k}$ is convergent. 
    
    To proceed with contradiction, assume that this sequence converges to some $\alpha > 0$. 
    Define a set $\Delta = \crl{\crl{p_{i,t}}| r_1 \leq V_{t}[\crl{p_{i,t}}] \leq r_2}$, 
    such that $\alpha \in (r_1, r_2)$.
    By the strict inequality over the pdfs not in equilibrium set, there exists $\gamma >0$ such that,
    \begin{align*}
      \gamma = \min_{\crl{p_{i,t}^{(k)}} \in \Delta} V_{t}\left[\crl{p_{i,t}^{(k)}}\right] - V_{t}\left[\crl{p_{i,t}^{(k+1)}}\right]
    \end{align*}
    Since $\lim_{k \rightarrow \infty} V_{t,k} = \alpha$, there exists $k_1$ such that for all $k > k_1$, $V_{t,k} \leq \alpha + \gamma'$,
    with $\gamma' < \gamma$. 
    Since $\crl{p_{i,t}^{(k)}}$ belongs to $\Delta$, we have $V_{t,k} - V_{t,k+1} \geq \gamma$,
    upper bounding $V_{t,k+1} \leq -\gamma + V_{t,k} \leq \alpha + \gamma' - \gamma < \alpha$, which is a contradiction.
  \end{proof}


\begin{proof}[Lemma~\ref{lemma:marginal_likelihood}]
    From Assumption \ref{assume:gradbound}, the agent likelihood at agent~$i$ satisfies $\qz{i,t}(z_{i,t}|\hypspace_i)^{\alpha_t} \in [e^{-\alpha_t L}, e^{\alpha_t L}]$. Since we are integrating the agent likelihood with unit measure pdfs $v_{i,t}(\vect{y})$, the resulting agent-variable likelihood satisfies $\qz{i,t}(z_{i,t}|\vect{x})^{\alpha_t} \in [e^{-\alpha_t L}, e^{\alpha_t L}]$ as well.
  \end{proof}

  \begin{proof}[Lemma~\ref{lemma:attract}]
    Let us define the log-ratio of the product form of the mixed pdf at two arbitrary values $\hypspace^{(1)}, \hypspace^{(2)} \in \real^d$ 
    composed of $m$-vector components corresponding to~$\crl{\vect{x}_{\ell}^{(1)}, \vect{x}_{\ell}^{(2)}}_{\ell=1}^m$ as,
    \begin{align*}
        H_{i,t+1} & = \log \left[\frac{\prod_{\vect{x}^{(1)} \in \hypspace_i^{(1)}} v_{i,t+1}(\vect{x}^{(1)})}{\prod_{\vect{x}^{(2)} \in \hypspace_i^{(2)}} v_{i,t+1}(\vect{x}^{(2)})}\right], \\
        & = \sum_{j=1}^n A_{ij} \log \left[\frac{p_{i,t}(\hypspace_i^{(1)}\backslash \hypspace_{ij}^{(1)}|\hypspace_{ij}^{(1)}) p_{ji,t}(\hypspace_{ij}^{(1)})}
        {p_{i,t}(\hypspace_i^{(2)} \backslash \hypspace_{ij}^{(2)}|\hypspace_{ij}^{(2)}) p_{ji,t}(\hypspace_{ij}^{(2)})}\right].
    \end{align*}
    Due to the independence assumption on the variables, we define the conditional density as $p_{i,t}(\hypspace_i^{(1)}\backslash \hypspace_{ij}^{(1)}|\hypspace_{ij}^{(1)}) = \prod_{\vect{x} \in \hypspace_i^{(1)} \backslash \hypspace_{ij}^{(1)}} p_{i,t} (\vect{x})$ and the shared marginal as $p_{ji,t}(\hypspace_{ij}^{(1)}) = \prod_{\vect{x} \in \hypspace_{ij}^{(1)}} p_{j,t} (\vect{x})$.
    Thus, we can write the log-probability ratios $H_{i,t}^{(\vect{x})}$ at any variable $\vect{x} \in \hypspace$ as linear weighted sum for all agents $i \in \nodes$,
    \begin{align}
      \label{eqn:ind_H}
        H_{i,t+1}^{(\vect{x})} & = \sum_{j \in \nodes \backslash \nodes(\vect{x})} A_{ij} H_{i,t}^{(\vect{x})}
        + \sum_{j \in \nodes(\vect{x})} A_{ij} H_{j,t}^{(\vect{x})} \\
        & = (1-\sum_{j \in \nodes(\vect{x})} A_{ij}) H_{i,t}^{(\vect{x})}
        + \sum_{j \in \nodes(\vect{x}) \backslash \{i\}} A_{ij} H_{j,t}^{(\vect{x})} \nonumber
    \end{align}
    The agents observing the variable $\vect{x}$ form the set $\nodes(\vect{x})$.
    Per Assumption~\ref{assume:marginal_network}, the agents in $\nodes(\vect{x})$ induce a 
    connected subgraph from graph $\graph$.
    For each $\vect{x}$, this representation leads to row-stochastic linear updates in log-proability ratio $H^{(\vect{x})}_{t}$.
    Let us define the communication matrix associated to $\vect{x}$ as $A(\vect{x}) \in [0,1]^{|\nodes(\vect{x})|\times |\nodes(\vect{x})|}$
    with the terms in \eqref{eqn:ind_H}. 
    Because of the base symmetry of the matrix $A$, the matrices $A(\vect{x})$ are column stochastic as well. 
    With the stochasticity, symmetry and underlying connectivity of $A(\vect{x})$,
    we employ the Theorem~$5$ in \cite{SB-SJC:14} to claim with $\sigma(A(\vect{x})) \in (0,1)$ that,
    \begin{align*}
      \| v_{i,t}(\vect{x}) - \bar{p}_{t}(\vect{x})\|_{TV} \leq \sigma(A(\vect{x}))\| p_{i,t}(\vect{x}) - \bar{p}_{t}(\vect{x}) \|_{TV}.
    \end{align*} 
    Thus, we can prove the claim with $\sigma(A) = \max_{\vect{x} \in \hypspace} \sigma(A(\vect{x})) < 1$,
    where the rate at any variable $\vect{x}$ is $\sigma(A(\vect{x})) = \lambda_{|\nodes(\vect{x})|-1}(A(\vect{x})^{\top}A(\vect{x}))$,
    written in terms of the second largest eigenvalue.
    It follows from the proof for \cite[Thm.~$5$]{SB-SJC:14} when expressed for 
    the sum of independent components that individually converge at an exponential rate $\sigma(A(\vect{x}))$.
    
    The doubly stochastic nature of the communication matrix $A(\vect{x})$ allows us to create an update rule for 
    the independent marginals defined in \eqref{eqn:independent_average} over the variable~$\vect{x}$.
    The update following Lemma~\ref{lemma:marginal_likelihood} is,
    \begin{align*}
      \bar{p}_{t+1}(\vect{x}) & \propto \prod_{i \in \nodes(\vect{x})}  \left( \qz{i}(z_{i,t}|\vect{x})^{\alpha_t} v_{i,t}(\vect{x}) \right)^{\frac{1}{|\nodes(\vect{x})|}} \\
      & \propto \prod_{i \in \nodes(\vect{x})}  \Bigl( \qz{i}(z_{i,t}|\vect{x})^{\alpha_t} \prod_{j \in \nodes(\vect{x})} p_{j,t}(\vect{x})^{A(\vect{x})_{ij}} \Bigr)^{\frac{1}{|\nodes(\vect{x})|}} \\
      & = \prod_{i \in \nodes(\vect{x})} \qz{i}(z_{i,t}|\vect{x})^{\frac{\alpha_t}{|\nodes(\vect{x})|}} \bar{p}_t(\vect{x}).
    \end{align*} 
    Define log-marginal likelihood $D_{i,t} = - \log(\qz{i}(z_{i,t} | \vect{x}))$ that inherits the bound $|D_{i,t}| \leq L$ from Lemma~\ref{lemma:marginal_likelihood}.
    We rewrite the update over all variables $\hypspace$ using the normalization factor 
    $\bar{Z}_{t+1}(\hypspace) = \int \prod_{\vect{x} \in \hypspace} \exp(\frac{- \alpha_t}{|\nodes(\vect{x})|} 
    \sum_{i \in \nodes(\vect{x})} D_{i,t}) \bar{p}_{t}(\hypspace) d\hypspace$ as, 
    \begin{align}
         & \bar{p}_{t+1}(\hypspace) = \exp \Bigl(\sum_{\vect{x} \in \hypspace}\frac{- \alpha_t}{|\nodes(\vect{x})|} \sum_{i \in \nodes(\vect{x})} D_{i,t} \Bigr) \bar{p}_{t}(\hypspace) \left/ \bar{Z}_{t+1} \right. \nonumber \\ 
        & = \arg\min_{p \in \calF_{d}} \left\{ \left\langle  \sum_{\vect{x} \in \hypspace}\frac{-\alpha_t}{|\nodes(\vect{x})|} \sum_{i \in \nodes(\vect{x})} D_{i,t}, p \right\rangle + \KL[p, \bar{p}_{t}] \right\}. \nonumber
    \end{align}
    Using the bound $|D_{i,t}| \leq L$, the inner product term satisfies, 
    $$\left\vert\sum_{\vect{x} \in \hypspace}\frac{-\alpha_t}{|\nodes(\vect{x})|} \sum_{i \in \nodes(\vect{x})} D_{i,t} \right\vert \leq m \alpha_t L. $$
    One can now follow the proof steps for Proposition~\ref{prop:dist_bound} since
    the pdf $\bar{p}_{t+1}$ optimizes an objective similar to $J_{i,t}[p, \bar{p}_t]$.
    \end{proof}

\begin{proof}[Proposition~\ref{prop:m_dist_bound}]
    The proof follows from Proposition~\ref{prop:dist_bound}, with the same bounds on the gradient term in the agent objective $J_{i,t}$ as defined in \eqref{eqn:dmsmd}, and the mixed pdf $v_{i,t}$ replacing the one in distributed setting.
  \end{proof}

\begin{proof}[Proposition~\ref{prop:m_summable_c}]
    Recall that $p^\star$ minimizes $f[p] = \sum_i f_i[p_i]$ where $p_i$ is the marginal density of pdf $p$. 
    For any pdf $\bar{p}_t$, we thus have~$f[p^\star] - f[\bar{p}_t] \leq 0$.  
    Using linearity of function~$f_i$ and simplifying with the average pdf $\bar{p}_t$ in \eqref{eqn:marginal_def}, 
    \begin{align*}
     & f[p^\star] - \sum_{i=1}^n f_i[v_{i,t}] = f[p^\star] - f[\bar{p}_t] + f[\bar{p}_t]- \sum_{i=1}^n f_i[v_{i,t}] \nonumber \\
    & \leq \sum_{i=1}^n (f_i[\bar{p}_t - v_{i,t}]) 
    =  \sum_{i=1}^n \expect_{z_{i,t}} \langle \frac{\delta F_{i,t}}{\delta p} [p_{i,t}], \bar{p}_t - v_{i,t} \rangle\\
    & = \sum_{i=1}^n \langle \frac{\delta f_i}{\delta p} [p_{i,t}], \bar{p}_t - v_{i,t} \rangle \leq \sum_{i=1}^n L \| \bar{p}_{i,t} - v_{i,t} \|_{TV} \\
    & \leq \sigma(A) L \sum_{i=1}^n \| \bar{p}_{i,t} - p_{i,t} \|_{TV},
    \end{align*}
    where the last two inequalities follow from Lemma~\ref{lemma:tvholder} and Conjecture~\ref{conj:attract}.
  \end{proof}

\begin{proof}[Proposition~\ref{prop:m_summable_cTV}]
    Applying the triangle inequality to the total variation distance between $p_{i,t+1}$ and $\bar{p}_{i,t+1}$ leads to:
    \begin{align*}
        & \| p_{i,t+1} (\hypspace_i) - \bar{p}_{i,t+1}(\hypspace_i)\|_{TV} \leq \| \bar{p}_{i,t}(\hypspace_i) - v_{i,t}(\hypspace_i)\|_{TV} \\
        &  + \| \bar{p}_{i,t+1}(\hypspace_i) - \bar{p}_{i,t}(\hypspace_i)\|_{TV} + \| v_{i,t}(\hypspace_i) - p_{i,t+1}(\hypspace_i)\|_{TV}.
    \end{align*}
    Based on the contraction in total variation in Conjecture~\ref{conj:attract} due to the mixing step, we have,
    \begin{align*}
    \sum_{i=1}^n \| v_{i,t}(\hypspace_i) - \bar{p}_{i,t}(\hypspace_i)\|_{TV} 
    \leq \sigma(A) \sum_{i=1}^n \| p_{i,t}(\hypspace_i) - \bar{p}_{i,t}(\hypspace_i)\|_{TV}
    \end{align*}
    The upper bounds on second and third term in Conjecture~\ref{conj:attract} and Proposition~\ref{prop:m_dist_bound} respectively lead to,
    \begin{align}
    & \sum_{i=1}^n \| p_{i,t+1}(\hypspace_i) - \bar{p}_{i,t+1}(\hypspace_i)\|_{TV} \nonumber \\
    \label{eqn:dm_summable}
    & \leq \sigma(A) \sum_{i=1}^n \| \bar{p}_{i,t}(\hypspace_i) - p_{i,t}(\hypspace_i)\|_{TV} + c\alpha_t L/2
    \end{align}
    Multiplying $\sigma(A) \alpha_t L$ to \eqref{eqn:dm_summable}, and substituting $a_t$ from this proposition's statement, we have, 
    $$a_{t+1} \leq \sigma(A) a_t + \sigma(A)c \alpha_t^2 L^2/2.$$
    This summability follows from the fact that this equation has the same form as~\eqref{eqn:TV_update} in the proof to Proposition~\ref{prop:summable_c}. 
\end{proof}

\begin{proof}[Proposition~\ref{prop:m_summable_cl}]
    Using Scheff\'{e}'s theorem in \cite[Lemma 2.1]{AT:08}, the total-variation distance is equivalent to the $L^1$ norm with the relation $\| \bar{p} - p \|_{TV} = 1/2 \int |\bar{p} - p |d \hypspace$ for any pdfs $\bar{p}, p \in \calF$.
    We construct a set $\Omega \in \mathcal{B}(\real^d)$ with points where the mixed pdf is greater than the agent estimate, $\Omega= \{\hypspace| \bar{p}_{i,t}(\hypspace_i) \geq v_{i,t}(\hypspace_i)\}$. 
    Note that the set is defined over a larger set of variables $\hypspace$ than the ones defining the probabilities $\hypspace_i$. 
    Then, Fubini-Tonelli theorem allows integrating out the common conditional as,
    \begin{align*}
      & \| \bar{p}_t - \bar{p}_t(\hypspace|\hypspace_i)v_{i,t} \|_{TV}
      = \| \bar{p}_t - \bar{p}_t(\hypspace|\hypspace_i)v_{i,t} \|_{1} \\
      & = \int_{\Omega} \bar{p}_t(\hypspace|\hypspace_i)(\bar{p}_{i,t} - v_{i,t}) d\hypspace - \int_{\Omega^c} \bar{p}(\hypspace|\hypspace_i)(\bar{p}_{i,t} - v_{i,t}) d\hypspace \\
      & = \int_{\Omega} (\bar{p}_{i,t} - v_{i,t}) d\hypspace_i - \int_{\Omega^c} (\bar{p}_{i,t} - v_{i,t}) d\hypspace_i \\
      & = \| \bar{p}_{i,t} - v_{i,t} \|_{1} = \frac{1}{2}\| \bar{p}_{i,t} - v_{i,t} \|_{TV} \leq \frac{\sigma(A)}{2}\| \bar{p}_{i,t} - p_{i,t} \|_{TV}.
    \end{align*}
    The summability now follows from Proposition~\ref{prop:m_summable_cTV}.
    \end{proof}
    

\begin{proof}[Theorem~\ref{thm:m_distributed_distance}]
    From the upper bound in Proposition~\ref{prop:m_dist_dec}, we have
      \begin{align*}
      \sum_{i=1}^n \KL [p^\star_i, v_{i, t+1}] & \leq \sum_{i=1}^n \KL [p^\star_i, p_{i, t+1}].
      \end{align*}
    This proof follows the arguments in Theorem~\ref{thm:distributed_distance} over the divergences 
    with pdfs~$p^\star_i(\hypspace_i)$ instead of~$p^\star(\hypspace)$.
    We note that the agent objective $J_{i,t}[p_i, v_{i,t}]$ mirrors the central objective in \eqref{eqn:cSMD} with 
    prior $v_{i,t}(\hypspace_i)$ and optimal pdf $p_{i,t+1}(\hypspace_i)$ instead of pdfs $p_t, p_{t+1}$ respectively. 
    Assume that the pdf $p_{i,t+1}$ minimizes functional~$J_{i,t}[p, v_{i,t}]$ defined in~\eqref{eqn:dmsmd}. 
    Following Proposition~\ref{prop:optim_KL} with~$\eta_i = p_i^\star - p_{i, t+1}$, 
    \begin{align}
    \label{eqn:m_kl_step}
    & \KL [p^\star_i, v_{i, t+1}] \leq \KL [p^\star_i, p_{i, t+1}]  \nonumber \\
    & \leq  \KL [p^\star_i, v_{i, t}] + \alpha_t \langle \frac{\delta F_{i,t}[p_{i,t}]}{\delta p}, p^\star_i - v_{i,t} \rangle + 2 \alpha_t^2 L^2.
    \end{align}
    
    The gradient of the linear functional $f_i[p_i]$ satisfies $\langle \frac{\delta f_i}{\delta p}[p_{i,t}], p^\star_i-v_{i,t}\rangle \leq f_i[p^\star_i] - f_i[v_{i,t}]$. 
    To simplify this, we add and subtract the expected gradient followed by expectation with respect to the natural filtration of the previous samples $\calZ_{t-1} = \sigma(z_{1:n,0}, \cdots, z_{1:n,t-1})$,
    \begin{align}
      \underset{\calZ_{t-1}}{\expect}[\KL [p^\star_i, v_{i, t+1}]] & \leq \KL [p^\star_i, v_{i,t}] + \alpha_t(f_i[p^\star_i] - f_i[v_{i,t}]) \nonumber \\
      + 2\alpha_t^2 L^2 \nonumber + \alpha_t & \underset{\calZ_{t-1}}{\expect}\left[\langle \frac{\delta F_{i,t}[p_{i,t}]}{\delta p} - \frac{\delta f_i[p_{i,t}]}{\delta p}, p^\star_i - v_{i,t}\rangle \right].
    \end{align}
    The term $G_{i,t}[p_{i,t}] = \frac{\delta F_{i,t}[p_{i,t}]}{\delta p} - \frac{\delta f[p_{i,t}]}{\delta p}$ forms a martingale difference sequence as~$\expect_{z_{i,t}} \left[G_{i,t}[p_{i,t}]| \calZ_{t-1} \right] = 0$ that is independent with respect to $\calZ_{t-1}$. Summing across agents and using Proposition~\ref{prop:m_dist_dec},
    \begin{align*}
    & \underset{\calZ_{t-1}}{\expect} \left[\sum_{i=1}^n \KL [p_i^\star,  v_{i, t+1}]\right] \\
    & \leq \sum_{i=1}^n \KL [p^\star, v_{i, t}] + 2 n \alpha_t^2 L^2 
    + \alpha_t \sum_{i=1}^n (f_i[p_i^\star] - f_i[v_{i,t}]) \\
    & \leq \sum_{i=1}^n \KL [p^\star, v_{i, t}] + 2 n \alpha_t^2 L^2
    + \sigma(A) \alpha_t L \sum_{i=1}^n \| \bar{p}_{i,t} - p_{i,t} \|_{TV},
    \end{align*}
    where the last inequality follows from Proposition~\ref{prop:m_summable_c}.
    The sequences $\alpha_t^2$ and~$\alpha_t L \sum_{i=1}^n \| \bar{p}_{i,t} - p_{i,t} \|_{TV}$ are summable following Assumption~\ref{assume:summable} and Proposition~\ref{prop:m_summable_cTV}. Therefore, by Gladyshev's lemma, $\sum_{i=1}^n \KL [p^\star, v_{i, t+1}]$ almost surely converges to a positive value.
    \end{proof}


\begin{proof}[Theorem~\ref{thm:m_distributed_pdf}]
    Analogous to the contradiction statement in the proof to Theorem~\ref{thm:distributed_pdf}, we assume at least one agent's mixed estimates~$\{v_{l,t}\}$ enter the partial neighborhood~$\mathcal{B}_l(\calF^\star, \epsilon)$ a finite number of times. 
    This implies existence of a time step~$t_0$ such that~$\KL[p^\star_l, v_{l,t}] \geq \epsilon, \forall t \geq t_0$. 
    Now, by definition of the set $\mathcal{B}_l(\calF^\star, \epsilon)$, the pdf $v_{l,t}(\hypspace_i) \notin \calF^\star_{\mathfrak{l}}$. 
    Therefore, the product $p(\hypspace|\hypspace_i)v_{i,t}(\hypspace_i) \notin \calF^\star$ for any pdf $p(\hypspace)$.
    
    Define $\pi_{t}(\hypspace) = \bar{p}_t(\hypspace | \hypspace_i) v_{l,t}(\hypspace_i)$.
    Since $\frac{\delta f}{\delta p}[p_{i,t}]$ is continuous in~$p_{i,t}$ with~$\langle \frac{\delta f}{\delta p}[p_{i,t}], \pi_{t} - p^\star\rangle = 0$ only if~$v_{l,t} \in \calF^\star$. Therefore, there exists~$c > 0$ for all $v_{l,t} \in \calF_{\mathfrak{d}_l} \backslash \mathcal{B}_i(\calF^\star, \epsilon)$,
    \begin{align}
      \label{eqn:c_marg_upper}
    & \langle \frac{\delta f}{\delta p}[\pi_{t}], p^\star - \pi_{t} \rangle \leq -c < 0, \\
    \implies & f[p^\star] - f[\pi_{t}] = f[p^\star] - \sum_{i=1}^n f_i[\pi_{i,t}] \leq -c, \nonumber 
    \end{align}
    where $\pi_{i,t}$ is the marginal of the pdf $\pi_t$ over variables $\hypspace_i$.
    Following the proof of Theorem~\ref{thm:m_distributed_distance}, we apply Proposition~\ref{prop:optim_KL} to the sum of objectives $J_{i,t}[p_{i,t}, v_{i,t}]$ defined in \eqref{eqn:dmsmd}, then use Proposition~\ref{prop:m_dist_dec}, and substitute gradient difference term $G_{i,t}[p_{i,t}] = \frac{\delta F_{i,t} [p_{i,t}]}{\delta p} - \frac{\delta f_i[p_{i,t}]}{\delta p}$ to obtain,
    \begin{align*}
      & \sum_{i=1}^n \KL [p_i^\star,  v_{i, t+1}] \leq \sum_{i=1}^n \KL [p^\star_i, v_{i, t}] + 2 n \alpha_t^2 L^2 \\
      & + \alpha_t \langle G_{i,t}[p_{i,t}], p^\star_i - v_{i,t}\rangle + \alpha_t \sum_{i=1}^n (f_i[p_i^\star] - f_i[v_{i,t}]) .
    \end{align*}
    To introduce the sharper upper bound described in~\eqref{eqn:c_marg_upper}, we simplify the last term by including objectives $f[\pi_{t}], f[\bar{p}_t]$,
    \begin{align*}
      & \alpha_t \sum_{i=1}^n (f_i[p_i^\star] - f_i[v_{i,t}]) = \alpha_t (f[p^\star] - \sum_{i=1}^n f_i[v_{i,t}]) \\
      & = \alpha_t (f[p^\star] - f[\pi_{t}] + f[\pi_{t}] - f[\bar{p}_t] + f[\bar{p}_t] - \sum_{i=1}^n f_i[v_{i,t}]) \\
      & \leq -\alpha_t c + 2\alpha_t L \sum_{i=1}^n (\| \pi_{t} - \bar{p}_{t} \|_{TV} + \sigma(A) \| \bar{p}_{i,t} - p_{i,t} \|_{TV}),
    \end{align*}
    where the total variation bounds follow from the linearity and upper bounding arguments in Proposition~\ref{prop:m_summable_c}.
    Now, define a sequence with terms~$d_t = 2 \alpha_t L \left( \sum_{i=1}^n \sigma (\| \bar{p}_{i,t} - p_{i,t} \|_{TV}) + n \| \pi_{t} - \bar{p}_{t} \|_{TV} \right)$. Based on Propositions~\ref{prop:m_summable_cTV} and~\ref{prop:m_summable_cl}, the sequence~$d_t$ is summable.
    Adding the definition $d_t$ and Proposition~\ref{prop:m_dist_dec} on mixed pdfs,
    \begin{align*}
    \sum_{i=1}^n \KL [p^\star_i, v_{i,t+1}] & \leq \sum_{i=1}^n \KL [p^\star_i, v_{i,t}] 
                                          - \alpha_t c + \left(2 n \alpha_t^2 L^2 + d_t \right) \nonumber \\
    & \, + \alpha_t \sum_{i=1}^n \langle G_{i,t}[p_{i,t}], p^\star_i - v_{i,t}\rangle 
    \end{align*}
    The remainder of the proof follows the proof to Theorem~\ref{thm:distributed_pdf}, and we have presented the key arguments for establishing this theorem. 
    As before, representing the divergence sum in terms of priors at time $t=0$ 
    with~$\beta_T = \sum_{t=0}^T \alpha_t$, and the inner product $g_{i,t} = \langle G_{i,t}[p_{i,t}], p^\star - v_{i,t}\rangle$,
    \begin{align*}
      \implies & \sum_{i=1}^n \KL [p^\star_i, v_{i,T+1}] \leq \sum_{i=1}^n \KL [p^\star_i, v_{i, 0}] \\
      & - \beta_T \left[ c  - \frac{\sum_{t=0}^T \alpha_t \sum_{i=1}^n g_{i,t}}{\beta_T} \right] 
      + \sum_{t=0}^T \left(2 n \alpha_t^2 L^2 + d_t \right)
    \end{align*}
    Since the agent objective functions $F_i$ are the same as Theorem~\ref{thm:distributed_pdf}, same upper bounds hold for expected gradient difference $G_{i,t}[p_{i,t}]$ and its inner product $g_{i,t}$ satisfies the martingale difference sequence condition,
    \begin{align*}
    \expect[G_{i,t}[p_{i,t}] \vert \calZ_{t-1}] = 0,
    \expect[||G_{i,t}[p_{i,t}]||_{\infty}^2 \vert \calZ_{t-1}] \leq 4L^2,\\
    \expect [g_{i,t}|\calZ_{t-1}] = 0, 
    \expect [g_{i,t}^2 |\calZ_{t-1}], \expect [g_{i,t}g_{j,t} |\calZ_{t-1}] \leq 16 L^2.
    \end{align*}
    Thus, the expected value of~$|\alpha_t \sum_{i=1}^n g_{i,t}|^2/\beta_t^2$ is bounded as,
    \begin{align*}
    \implies \sum_{t=0}^{\infty} \frac{\expect[|\alpha_t \sum_{i=1}^n g_{i,t}|^2 | \calZ_{t-1}]}{\beta_t^2} \leq 16 n^2 L^2 \sum_{t=0}^{\infty} \frac{\alpha_t^2}{\beta_t^2}.
    \end{align*}
    Since~$\lim_{t \rightarrow \infty}\beta_T= \infty$, we
    can use the strong law of large numbers for martingale difference
    sequences in Lemma~\ref{lemma:stronglaw} (for $X_t =
      \sum_{i=1}^n g_{i,t}$ and $p = 2$) to conclude that,
    \begin{align*}
    & \frac{\sum_{t=0}^T \alpha_t (\sum_{i=1}^n g_{i,t})}{\beta_T} \rightarrow 0 \quad as \quad T \rightarrow \infty \, (a.s.) \\
    \implies & \beta_T \left[ c  - \frac{\sum_{t=0}^T \alpha_t \sum_{i=1}^n g_{i,t}}{\beta_T} \right] \rightarrow \infty (a.s.)
    \end{align*}
    With the bounded~$\sum_{t=0}^{\infty} \alpha_t^2$, we thus have, 
    \begin{align*}
    \lim \sup_{T \rightarrow \infty} \sum_{i=1}^n \KL [p^\star_i, v_{i,T}] = - \infty.
    \end{align*}
    Along with non-negativity of divergence terms, this contradicts our assumption that $\KL[p^\star_l, v_{l,t}] \geq \epsilon, \forall t \geq t_0$.
    Therefore, every agent's sequence~$\{v_{i,t}\}$ enters the set $\mathcal{B}_i(\calF^\star, \epsilon)$ infinitely many times for all agents~$i \in \nodes$.
    From Theorem~\ref{thm:m_distributed_distance}, the KL-divergence sum of the sequences~$v_{i,t}, \forall i \in \nodes$ to optimal marginals converges to a constant value, 
    i.e., $\sum_{i=1}^n \KL[p^\star_i, v_{i,t}] \rightarrow d^\star$. Now, since estimates $v_{l,t}$ do not satisfy $\KL[p^\star_i,v_{l,t}] \ge \epsilon$ for an infinite number of $t$, for all $\epsilon >0$, it follows that $d^\star < n \epsilon $. Thus, $d^* \equiv 0$ and the conclusion on convergence follows.
    \end{proof}

%

\begin{IEEEbiography}
[{\includegraphics[width=1in,height=1.25in,clip,keepaspectratio]{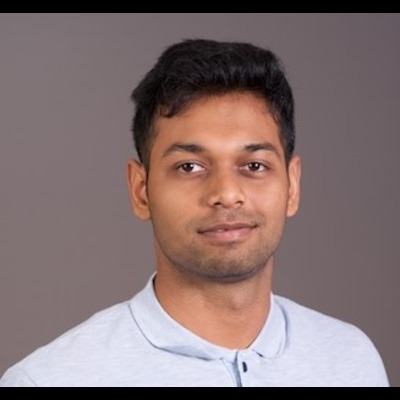}}]
{Parth Paritosh}
is currently a Postdoctoral Fellow with Research Associateship Program at the U.S. Army Combat Capabilities DEVCOM Army Research Laboratory.
He earned his Ph.D. from the Mechanical and Aerospace Engineering Department at the University of California, San Diego (UCSD). 
He obtained his M.S. in Mechanical Engineering from Purdue University in May 2017 and his 
B.Tech. in Mechanical Engineering with a minor in Computer Science and Engineering in May 2015. 
His research focuses on enhancing robotic localization and inference capabilities, particularly in multi-agent autonomous systems.
\end{IEEEbiography}

\begin{IEEEbiography}[{\includegraphics[width=1in,height=1.25in,clip,keepaspectratio]{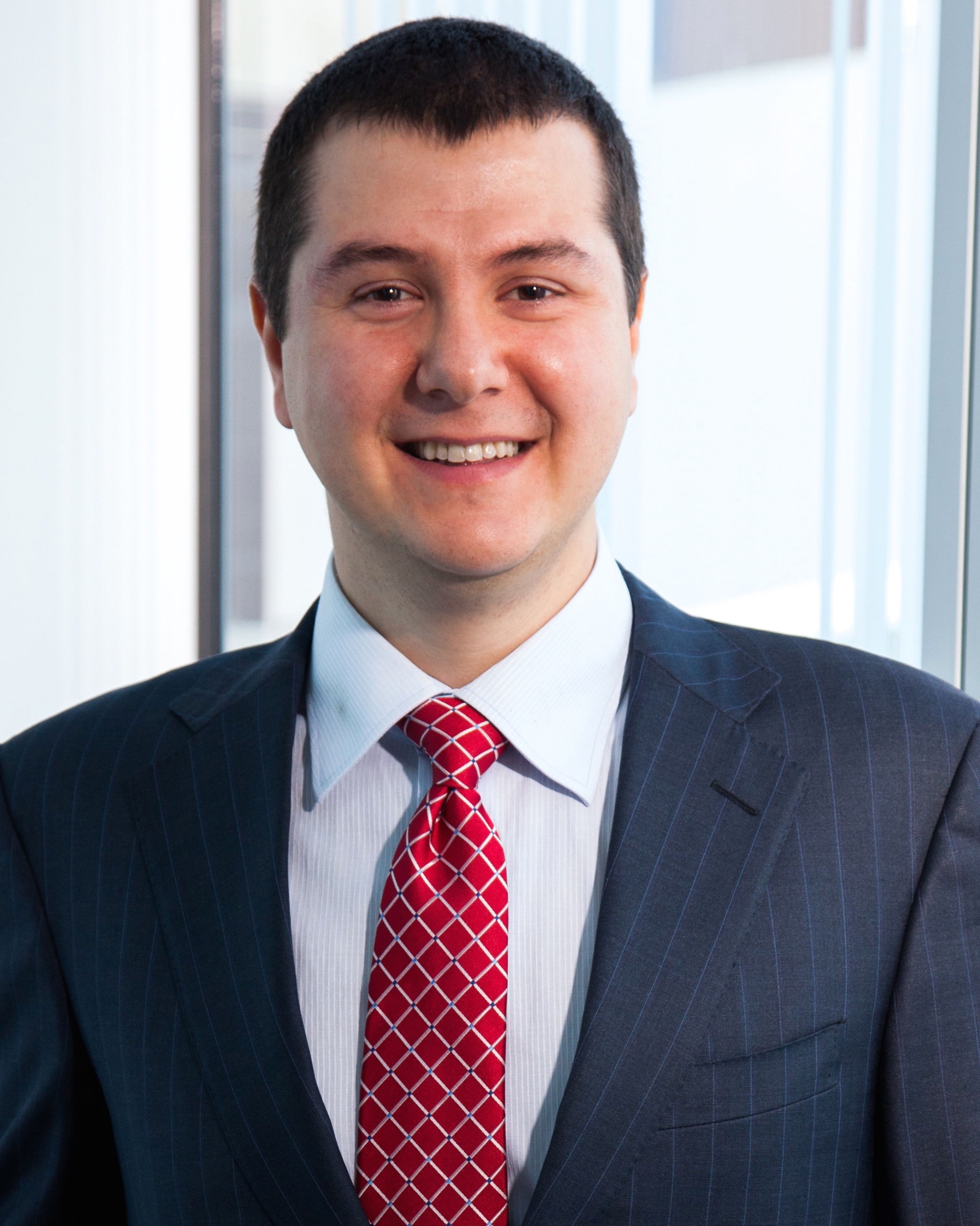}}]{Nikolay Atanasov}
(S'07-M'16-SM'23) is an Assistant Professor of Electrical and Computer Engineering at the University of California San Diego, La Jolla, CA, USA. He obtained a B.S. degree in Electrical Engineering from Trinity College, Hartford, CT, USA in 2008 and M.S. and Ph.D. degrees in Electrical and Systems Engineering from the University of Pennsylvania, Philadelphia, PA, USA in 2012 and 2015, respectively. Dr. Atanasov's research focuses on robotics, control theory, and machine learning, applied to active perception problems for autonomous mobile robots. He works on probabilistic models that unify geometric and semantic information in simultaneous localization and mapping (SLAM) and on optimal control and reinforcement learning algorithms for minimizing probabilistic model uncertainty. Dr. Atanasov's work has been recognized by the Joseph and Rosaline Wolf award for the best Ph.D. dissertation in Electrical and Systems Engineering at the University of Pennsylvania in 2015, the Best Conference Paper Award at the IEEE International Conference on Robotics and Automation (ICRA) in 2017, the NSF CAREER Award in 2021, and the IEEE RAS Early Academic Career Award in Robotics and Automation in 2023.
\end{IEEEbiography}

\begin{IEEEbiography}[{\includegraphics[width=1in,height=1.25in,clip,keepaspectratio]{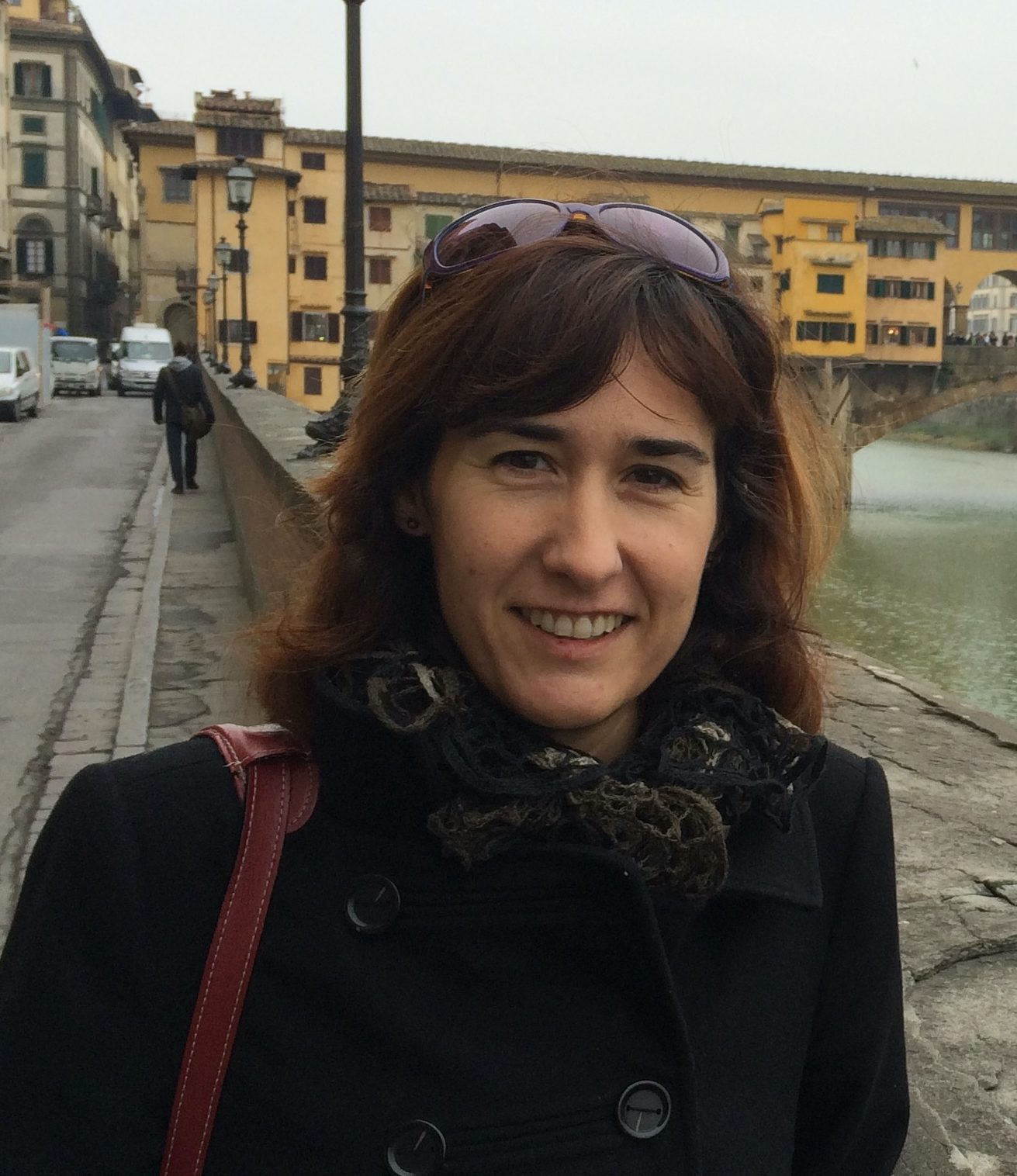}}]{Sonia Mart{\'\i}nez}
  (M'02-SM'07-F'18) is a Professor of Mechanical
  and Aerospace Engineering at the University of California, San
  Diego, CA, USA. She received her Ph.D. degree in Engineering
  Mathematics from the Universidad Carlos III de Madrid, Spain, in May
  2002. She was a Visiting Assistant Professor of Applied Mathematics
  at the Technical University of Catalonia, Spain (2002-2003), a
  Postdoctoral Fulbright Fellow at the Coordinated Science Laboratory
  of the University of Illinois, Urbana-Champaign (2003-2004) and the
  Center for Control, Dynamical systems and Computation of the
  University of California, Santa Barbara (2004-2005).  Her research
  interests include the control of networked systems, multi-agent
  systems, nonlinear control theory, and planning algorithms in
  robotics. She is a Fellow of IEEE. She is a co-author (together with
  F. Bullo and J. Cort\'es) of ``Distributed Control of Robotic
  Networks'' (Princeton University Press, 2009). She is a co-author
  (together with M. Zhu) of ``Distributed Optimization-based Control
  of Multi-agent Networks in Complex Environments'' (Springer, 2015).
  She is the Editor in Chief of the recently launched \textit{CSS IEEE
  Open Journal of Control Systems.}
\end{IEEEbiography}




\end{document}

%% file: packages.tex
\usepackage{amssymb}
\usepackage{amsthm}
\usepackage{amsmath}
\usepackage{epsfig}
\usepackage[noend]{algorithm2e}
\usepackage{amsfonts}
\usepackage{amsmath}
\usepackage{multirow}
\usepackage{pdflscape}
\usepackage{epstopdf}
\usepackage{listings}  
\usepackage{dsfont}
\usepackage[breaklinks=true, colorlinks, bookmarks=true, citecolor=Black, urlcolor=Violet,linkcolor=Black]{hyperref}
\usepackage[normalem]{ulem}
\usepackage{cleveref}
\usepackage[table,usenames,dvipsnames]{xcolor}      
\usepackage{tikz}

\newcommand{\Be}{\mathbb{B}}
\newcommand{\In}{\mathbb{I}}
\newcommand{\Lp}{\mathrm{L}^{1}}
\newcommand{\expect}{\mathbb{E}}

\newcommand{\revisionAdd}[1]{\textcolor{Black}{#1}}

\newcommand{\calN}{\mathcal{N}}
\newcommand{\calM}{\mathcal{M}}
\newcommand{\calF}{\mathcal{F}}
\newcommand{\calG}{\mathcal{G}}
\newcommand{\calL}{\mathcal{L}}

\newcommand{\calW}{\mathcal{W}}
\newcommand{\calX}{\mathcal{X}}

\newcommand{\calZ}{\mathcal{Z}}

\newtheorem{theorem}{Theorem}

\newtheorem{lemma}[theorem]{Lemma}
\newtheorem{proposition}[theorem]{Proposition}
\theoremstyle{definition}
\newtheorem{definition}{Definition}
\newtheorem{assumption}{Assumption}
\newtheorem{conjecture}{Conjecture}
\newtheorem*{assumption*}{Assumption}
\newtheorem*{problem*}{Problem}
\newtheorem{problem}{Problem}
\theoremstyle{remark}

\newcommand{\scaleMathLine}[2][1]{\resizebox{#1\linewidth}{!}{$\displaystyle{#2}$}}

\newcommand{\real}{\ensuremath{\mathbb{R}}}

\newcommand{\nodes}{\mathcal{V}}
\newcommand{\edges}{\mathcal{E}}
\newcommand{\graph}{\mathcal{G}}

\newcommand{\hypspace}{\mathcal{X}}

\newcommand{\until}[1]{\{1,\dots, #1\}}

\newcommand{\qz}[1]{\operatorname{q}_{#1}}
\newcommand{\vect}[1]{\boldsymbol{#1}}
\newcommand{\ones}[1][]{\vect{1}_{#1}}

\newcommand{\neighbor}[1]{\mathcal{V}_{#1}}

\newcommand{\KL}{\operatorname{KL}} 

\newcommand{\squaresym}{\hbox{$\blacksquare$}}
\newcommand{\proofend}{\relax\ifmmode\else\unskip\hfill\fi\squaresym \\}
\renewenvironment{proof}{\textit{Proof.}}{\proofend}


%% file: main.bbl
\begin{thebibliography}{10}
\providecommand{\url}[1]{#1}
\csname url@samestyle\endcsname
\providecommand{\newblock}{\relax}
\providecommand{\bibinfo}[2]{#2}
\providecommand{\BIBentrySTDinterwordspacing}{\spaceskip=0pt\relax}
\providecommand{\BIBentryALTinterwordstretchfactor}{4}
\providecommand{\BIBentryALTinterwordspacing}{\spaceskip=\fontdimen2\font plus
\BIBentryALTinterwordstretchfactor\fontdimen3\font minus \fontdimen4\font\relax}
\providecommand{\BIBforeignlanguage}[2]{{%
\expandafter\ifx\csname l@#1\endcsname\relax
\typeout{** WARNING: IEEEtran.bst: No hyphenation pattern has been}%
\typeout{** loaded for the language `#1'. Using the pattern for}%
\typeout{** the default language instead.}%
\else
\language=\csname l@#1\endcsname
\fi
#2}}
\providecommand{\BIBdecl}{\relax}
\BIBdecl

\bibitem{FZ-AG-KKL:19}
F.~Zafari, A.~Gkelias, and K.~K. Leung, ``A survey of indoor localization systems and technologies,'' \emph{IEEE Commun. Surv. Tutor.}, vol.~21, no.~3, pp. 2568--2599, 2019.

\bibitem{SK-AD-SSH-JSK-SES:16}
S.~Kumar, A.~Deshpande, S.~S. Ho, J.~S. Ku, and S.~E. Sarma, ``Urban street lighting infrastructure monitoring using a mobile sensor platform,'' \emph{IEEE Sens. J.}, vol.~16, no.~12, pp. 4981--4994, 2016.

\bibitem{FB-JC-SM:09}
F.~Bullo, J.~Cort\'es, and S.~Mart{\'\i}nez, \emph{Distributed control of robotic networks: a mathematical approach to motion coordination algorithms}.\hskip 1em plus 0.5em minus 0.4em\relax Princeton University Press, 2009.

\bibitem{AJ-PM-AS-ATS:12}
A.~Jadbabaie, P.~Molavi, A.~Sandroni, and A.~Tahbaz-Salehi, ``\BIBforeignlanguage{en}{Non-{Bayesian} social learning},'' \emph{\BIBforeignlanguage{en}{Games. Econ. Behav.}}, vol.~76, no.~1, pp. 210--225, Sep. 2012.

\bibitem{RTC-RLW:99}
R.~T. Clemen and R.~L. Winkler, ``Combining probability distributions from experts in risk analysis,'' \emph{Risk Anal.}, vol.~19, no.~2, pp. 187--203, 1999.

\bibitem{TM:05}
T.~Minka, ``Divergence measures and message passing,'' Tech. Rep., 2005, {M}icrosoft {R}esearch MSR-TR-2005-173.

\bibitem{FM-OH-FH:13}
F.~Meyer, O.~Hlinka, and F.~Hlawatsch, ``Sigma point belief propagation,'' \emph{IEEE Trans. Signal Process.}, vol.~21, no.~2, pp. 145--149, 2013.

\bibitem{MK-YI-ET-AS:23}
M.~Kayaalp, Y.~Inan, E.~Telatar, and A.~H. Sayed, ``On the arithmetic and geometric fusion of beliefs for distributed inference,'' \emph{IEEE Transactions on Automatic Control}, 2023.

\bibitem{MEK-HR:23}
M.~E. Khan and H.~Rue, ``The bayesian learning rule,'' \emph{J. Mach. Learn. Res.}, vol.~24, no. 281, pp. 1--46, 2023.

\bibitem{AG-TSJ-SV-HZ:04}
A.~Garg, T.~S. Jayram, S.~Vaithyanathan, and H.~Zhu, ``Generalized opinion pooling,'' \emph{Ann. Math. Artif. Intell.}, 2004.

\bibitem{GK-YE-PD-FH:22}
G.~Koliander, Y.~El-Laham, P.~M. Djurić, and F.~Hlawatsch, ``Fusion of probability density functions,'' \emph{Proceedings of the IEEE}, vol. 110, no.~4, pp. 404--453, 2022.

\bibitem{AN:12}
A.~Nemirovski, ``Tutorial: Mirror descent algorithms for large-scale deterministic and stochastic convex optimization,'' in \emph{Conference on Learning Theory}, 2012.

\bibitem{ZZ-PM-NB-SPB-PWG:20}
Z.~Zhou, P.~Mertikopoulos, N.~Bambos, S.~P. Boyd, and P.~W. Glynn, ``On the convergence of mirror descent beyond stochastic convex programming,'' \emph{SIAM J. optim.}, vol.~30, no.~1, pp. 687--716, 2020.

\bibitem{KRR-ATS:10}
K.~R. Rad and A.~Tahbaz-Salehi, ``Distributed parameter estimation in networks,'' in \emph{{IEEE} Conf. on Decision and Control}.\hskip 1em plus 0.5em minus 0.4em\relax IEEE, 2010, pp. 5050--5055.

\bibitem{AN-AO-CAA:17}
A.~Nedi{\'c}, A.~Olshevsky, and C.~A. Uribe, ``Fast convergence rates for distributed non-{B}ayesian learning,'' \emph{IEEE Trans. Autom. Contr.}, vol.~62, no.~11, pp. 5538--5553, 2017.

\bibitem{AL-AS-TJ:14}
A.~Lalitha, A.~Sarwate, and T.~Javidi, ``Social learning and distributed hypothesis testing,'' in \emph{IEEE Int. Symp. on Info. Theory}.\hskip 1em plus 0.5em minus 0.4em\relax IEEE, 2014, pp. 551--555.

\bibitem{TTD-SB-HDN-CLB:18}
T.~T. Doan, S.~Bose, D.~H. Nguyen, and C.~L. Beck, ``Convergence of the iterates in mirror descent methods,'' \emph{IEEE Control Syst. Lett.}, vol.~3, no.~1, pp. 114--119, 2018.

\bibitem{CAA-AO-AN:22}
C.~A. Uribe, A.~Olshevsky, and A.~Nedich, ``Non-asymptotic concentration rates in cooperative learning part {I}: Variational non-{B}ayesian social learning,'' \emph{IEEE Trans. Control. Netw. Syst.}, 2022.

\bibitem{NA-RT-VP:14}
N.~Atanasov, R.~Tron, V.~M. Preciado, and G.~J. Pappas, ``Joint estimation and localization in sensor networks,'' in \emph{IEEE Conf. on Decision and Control}, 2014, pp. 6875--6882.

\bibitem{GP-IS-BF-FB-BDA:13}
G.~Piovan, I.~Shames, B.~Fidan, F.~Bullo, and B.~D. Anderson, ``On frame and orientation localization for relative sensing networks,'' \emph{Automatica}, vol.~49, no.~1, pp. 206--213, 2013.

\bibitem{JP:82}
J.~Pearl, ``Reverend {B}ayes on inference engines: A distributed hierarchical approach,'' in \emph{Proceedings of the Second AAAI Conference on Artificial Intelligence}, 1982, pp. 133--136.

\bibitem{JSY-WTF-YW:02}
J.~S. Yedidia, W.~T. Freeman, and Y.~Weiss, ``Understanding belief propagation and its generalizations.''

\bibitem{YL-PD:19}
Y.-H. Liu and D.~Poulin, ``Neural belief-propagation decoders for quantum error-correcting codes,'' \emph{Physical review letters}, vol. 122, no.~20, p. 200501, 2019.

\bibitem{DK-SL-AO-OCJ:19}
K.~Desingh, S.~Lu, A.~Opipari, and O.~C. Jenkins, ``Efficient nonparametric belief propagation for pose estimation and manipulation of articulated objects,'' \emph{Science Robotics}, vol.~4, no.~30, p. eaaw4523, 2019.

\bibitem{TH:04}
T.~Heskes, ``On the uniqueness of loopy belief propagation fixed points,'' \emph{Neural Computation}, vol.~16, no.~11, pp. 2379--2413, 2004.

\bibitem{LD:20}
D.~Liu, ``Perspectives on probabilistic graphical models,'' Ph.D. dissertation, KTH Royal Institute of Technology, 2020.

\bibitem{VB:21}
\BIBentryALTinterwordspacing
V.~Bouttier, ``{Circular belief propagation as a model for optimal and suboptimal inference in the brain : extending the algorithm and proposing a neural implementation},'' Theses, {Universit{\'e} Paris Cit{\'e}}, Dec. 2021. [Online]. Available: \url{https://theses.hal.science/tel-04530051}
\BIBentrySTDinterwordspacing

\bibitem{BV:08}
B.~Vantaggi, ``Statistical matching of multiple sources: A look through coherence,'' \emph{Int. J. Approx. Reason.}, vol.~49, no.~3, pp. 701--711, 2008.

\bibitem{YYS-AKY:23}
Y.~Y. Shkel and A.~K. Yadav, ``Information spectrum converse for minimum entropy couplings and functional representations,'' in \emph{2023 IEEE International Symposium on Information Theory (ISIT)}.\hskip 1em plus 0.5em minus 0.4em\relax IEEE, 2023, pp. 66--71.

\bibitem{JK:11}
J.~Krac{\'\i}k, ``Combining marginal probability distributions via minimization of weighted sum of kullback--leibler divergences,'' \emph{Int. J. Approx. Reason.}, vol.~52, no.~6, pp. 659--671, 2011.

\bibitem{FC-LG-UV:19}
F.~Cicalese, L.~Gargano, and U.~Vaccaro, ``Minimum-entropy couplings and their applications,'' \emph{IEEE Transactions on Information Theory}, vol.~65, no.~6, pp. 3436--3451, 2019.

\bibitem{PP-NA-SM:19}
P.~Paritosh, N.~Atanasov, and S.~Martinez, ``Hypothesis assignment and partial likelihood averaging for cooperative estimation,'' in \emph{{IEEE} Int. Conf. on Decision and Control}.\hskip 1em plus 0.5em minus 0.4em\relax IEEE, 2019, pp. 7850--7856.

\bibitem{RP-MF-BT:22}
R.~Parasnis, M.~Franceschetti, and B.~Touri, ``Non-bayesian social learning on random digraphs with aperiodically varying network connectivity,'' \emph{IEEE Transactions on Control of Network Systems}, vol.~9, no.~3, pp. 1202--1214, 2022.

\bibitem{DB:08}
D.~Bickson, ``Gaussian belief propagation: Theory and application,'' \emph{arXiv preprint arXiv:0811.2518}, 2008.

\bibitem{JP:22}
J.~Pearl, ``Fusion, propagation, and structuring in belief networks,'' in \emph{Probabilistic and Causal Inference: The Works of Judea Pearl}, 2022, pp. 139--188.

\bibitem{PP-NA-SM:20}
P.~Paritosh, N.~Atanasov, and S.~Mart{\'\i}nez, ``Marginal density averaging for distributed node localization from local edge measurements,'' in \emph{{IEEE} Int. Conf. on Decision and Control}.\hskip 1em plus 0.5em minus 0.4em\relax IEEE, 2020, pp. 2404--2410.

\bibitem{ATI-ASW:05}
A.~T. Ihler, A.~S. Willsky \emph{et~al.}, ``Loopy belief propagation: Convergence and effects of message errors,'' \emph{J. Mach. Learn. Res.}, vol.~6, no. May, pp. 905--936, 2005.

\bibitem{VB-RJ-SD:24}
V.~Bouttier, R.~Jardri, and S.~Deneve, ``Circular belief propagation for approximate probabilistic inference,'' \emph{arXiv preprint arXiv:2403.12106}, 2024.

\bibitem{RS-PK:67}
R.~Sinkhorn and P.~Knopp, ``Concerning nonnegative matrices and doubly stochastic matrices,'' \emph{Pac. J. Math.}, vol.~21, no.~2, pp. 343--348, 1967.

\bibitem{AS-DD-AR:14}
A.~Shapiro, D.~Dentcheva, and A.~Ruszczyński, \emph{Lectures on Stochastic Programming: Modeling and Theory}.\hskip 1em plus 0.5em minus 0.4em\relax SIAM, 2014.

\bibitem{SB:15}
S.~Bubeck, ``Convex optimization: Algorithms and complexity,'' \emph{Found. Trends Mach. Learn.}, vol.~8, no. 3-4, pp. 231--357, 2015.

\bibitem{ASN-DBY:83}
A.~Nemirovsky and D.~Yudin, \emph{Problem complexity and method efficiency in optimization.}\hskip 1em plus 0.5em minus 0.4em\relax Wiley, 1983.

\bibitem{BAF-SS-MRG:08}
B.~A. Frigyik, S.~Srivastava, and M.~R. Gupta, ``{Functional Bregman divergence and {B}ayesian estimation of distributions},'' \emph{IEEE Trans. Inf. Theory}, vol.~54, no.~11, pp. 5130--5139, 2008.

\bibitem{DL:11}
D.~Liberzon, \emph{Calculus of variations and optimal control theory: a concise introduction}.\hskip 1em plus 0.5em minus 0.4em\relax Princeton University Press, 2011.

\bibitem{WC:01}
W.~Cheney, \emph{Analysis for applied mathematics}.\hskip 1em plus 0.5em minus 0.4em\relax Springer Science \& Business Media, 2001, vol. 208.

\bibitem{AT:08}
A.~B. Tsybakov, \emph{Introduction to Nonparametric Estimation}, 1st~ed.\hskip 1em plus 0.5em minus 0.4em\relax Springer Publishing Company, Incorporated, 2008.

\bibitem{MSP:60}
M.~Pinsker, ``Information and information stability of random variables and processes (in russian),'' 1960.

\bibitem{BTP:87}
B.~T. Polyak, ``Introduction to optimization,'' \emph{Inc., Publications Division, New York}, vol.~1, p.~49, 1987.

\bibitem{PH-CHH:14}
P.~Hall and C.~C. Heyde, \emph{Martingale limit theory and its application}.\hskip 1em plus 0.5em minus 0.4em\relax Academic press, 2014.

\bibitem{PP-NA-SM:22}
P.~Paritosh, N.~Atanasov, and S.~Martinez, ``Distributed {B}ayesian estimation of continuous variables over time-varying directed networks,'' \emph{IEEE Control Syst. Lett.}, vol.~6, pp. 2545--2550, 2022.

\bibitem{SB-SJC:14}
S.~Bandyopadhyay and S.-J. Chung, ``Distributed estimation using {B}ayesian consensus filtering,'' in \emph{2014 American control conference}.\hskip 1em plus 0.5em minus 0.4em\relax IEEE, 2014, pp. 634--641.

\bibitem{BF-SG:22}
B.~Franci and S.~Grammatico, ``Convergence of sequences: A survey,'' \emph{Annu. Rev. Control}, vol.~53, pp. 161--186, 2022.

\bibitem{PP-NA-SM:23-arxiv}
P.~Paritosh, N.~Atanasov, and S.~Martinez, ``Distributed variational inference for online supervised learning,'' \emph{arXiv preprint arXiv:2309.02606}, 2023.

\bibitem{DB:18}
D.~S. Bernstein, \emph{Scalar, Vector, and Matrix Mathematics: Theory, Facts, and Formulas - Revised and Expanded Edition}.\hskip 1em plus 0.5em minus 0.4em\relax Princeton University Press, 2018.

\bibitem{WR:06}
W.~Rudin, \emph{Real and complex analysis}.\hskip 1em plus 0.5em minus 0.4em\relax McGraw-Hill, 2006.

\bibitem{EM-MR:18}
E.~Mariucci and M.~Reiß, ``Wasserstein and total variation distance between marginals of lévy processes,'' \emph{Electron. J. Statist.}, vol.~12, no.~2, pp. 2482--2514, 2018.

\bibitem{RD:19}
R.~Durrett, \emph{Probability: theory and examples}.\hskip 1em plus 0.5em minus 0.4em\relax Cambridge university press, 2019, vol.~49.

\end{thebibliography}
